\documentclass[twoside,11pt]{article}

\usepackage{jmlr2e}
\hypersetup{ hidelinks }
\usepackage{xspace}

\usepackage[nosumlimits,nonamelimits]{amsmath}
\usepackage[utf8]{inputenc} %
\usepackage[T1]{fontenc}    %
\usepackage{hyperref}       %
\usepackage{url}            %
\usepackage{booktabs}       %
\usepackage{amsfonts}       %
\usepackage{nicefrac}       %
\usepackage{microtype}      %
\usepackage{xcolor}         %
\usepackage{adjustbox}
\usepackage{enumitem}
\usepackage{multirow}
\usepackage{diagbox}
\usepackage{amsmath}
\usepackage{mathtools}
\usepackage{float}

\usepackage{subfigure}

\usepackage{caption}

\usepackage{tikz}
\usetikzlibrary{fit,positioning}

\usepackage{dsfont}

\usepackage{graphicx} %
\usepackage{amssymb}
\usepackage{times}
\usepackage{epsfig}
\usepackage{graphicx}
\usepackage{color}
\usepackage{bbm}
\usepackage{multicol}
\usepackage{natbib}
\usepackage{wrapfig}

\providecommand{\dif}{\mathop{}\!\mathrm d}
\providecommand{\Ex}{\mathbb E}
\providecommand{\Var}{\mathbb V}
\providecommand{\hide}[1]{}

\usepackage{rotating}

\let\proof\relax

\usepackage{amsthm}
\usepackage{amsopn,amssymb,thmtools,thm-restate}

\newtheorem{thm}{Theorem}%
\newtheorem{lem}[thm]{Lemma}
\newtheorem{prop}[thm]{Proposition}
\newtheorem{cor}[thm]{Corollary}
\newtheorem{assumption}[thm]{Assumption}

\newtheorem{exmp}{Example}[section]

\usepackage{bm} %
\usepackage{algorithm}%
\usepackage{algpseudocode}%
\newcommand{\vct}[1]{\boldsymbol{#1}} %
\newcommand{\mat}[1]{\boldsymbol{#1}} %

\newcommand{\field}[1]{\mathbb{#1}}
\newcommand{\R}{\field{R}} %
\newcommand{\Z}{\field{Z}} %
\newcommand{\T}{^{\top}} %

\newcommand{\ProbOpr}[1]{\mathbb{#1}}

\newcommand{\expect}[2]{%
\ifthenelse{\equal{#2}{}}{\ProbOpr{E}_{#1}}
{\ifthenelse{\equal{#1}{}}{\ProbOpr{E}\left[#2\right]}{\ProbOpr{E}_{#1}\left[#2\right]}}} %
\newcommand{\var}[2]{%
\ifthenelse{\equal{#2}{}}{\ProbOpr{VAR}_{#1}}
{\ifthenelse{\equal{#1}{}}{\ProbOpr{VAR}\left[#2\right]}{\ProbOpr{VAR}_{#1}\left[#2\right]}}} %

\DeclareMathOperator{\argmax}{arg\,max}
\DeclareMathOperator{\argmin}{arg\,min}

\newcommand{\vmu}{\vct{\mu}}

\newcommand{\vx}{{\vct{x}}}
\newcommand{\vy}{\vct{y}}

\newcommand{\vu}{\vct{u}}

\newcommand{\vv}{\vct{v}}

\newcommand{\mI}{\mat{I}}

\newcommand{\fx}{\mathfrak{X}}

\def\N{{\mathcal{N}}}
\def\GP{\mathcal{GP}}

\def\L{{\mathcal{L}}}

\def\KL{\textrm{KL}}

\def\tr{\textrm{tr}}

\def\card{\textrm{card}}

\newcommand{\pgp}{$\GP(\hat \mu, \hat k\circ\hat \sigma^2)$\xspace}
\newcommand{\gtgp}{$\GP(\mu^*, k^*\circ\sigma_*^2)$\xspace}
\newcommand{\ppgp}{$\GP(\hat \mu, \hat k\circ\hat \sigma^2 \mid D_f)$\xspace}
\newcommand{\pgtgp}{$\GP(\mu^*, k^*\circ\sigma_*^2 \mid D_f )$\xspace}
\newcommand{\iid}{\textit{i.i.d.}\xspace}

\newcommand{\matern}{Mat\'ern}
\newcommand{\srange}[2] {\lbrack #1, #2 \rbrack}

\newcommand{\hyperbo}{HyperBO\xspace}
\newcommand{\bo}{BO\xspace}

\usepackage{lastpage}
\jmlrheading{25}{2024}{1-\pageref{LastPage}}{3/23; Revised
6/24}{7/24}{23-0269}{Zi Wang, George E. Dahl, Kevin Swersky, Chansoo Lee, Zachary Nado, Justin Gilmer, Jasper Snoek and Zoubin Ghahramani}
\ShortHeadings{Pre-trained Gaussian processes for Bayesian optimization}{Wang, Dahl, Swersky, Lee, Nado, Gilmer, Snoek and Ghahramani}

\firstpageno{1}

\begin{document}
\title{Pre-trained Gaussian Processes for Bayesian Optimization}
\author{\name Zi Wang \email wangzi@google.com 
       \AND
       \name George E. Dahl \email gdahl@google.com 
       \AND 
       \name Kevin Swersky \email kswersky@google.com
       \AND 
       \name Chansoo Lee \email chansoo@google.com
       \AND 
       \name Zachary Nado \email znado@google.com
       \AND 
       \name Justin Gilmer \email gilmer@google.com
       \AND 
       \name Jasper Snoek \email jsnoek@google.com
       \AND 
       \name Zoubin Ghahramani \email zoubin@google.com\\
       \addr Google DeepMind
       }
\editor{Marc Peter Deisenroth}

\maketitle

\begin{abstract}%
Bayesian optimization (BO) has become a popular strategy for global optimization of expensive real-world functions. Contrary to a common expectation that BO is suited to optimizing black-box functions, it actually requires domain knowledge about those functions to deploy BO successfully. Such domain knowledge often manifests in Gaussian process (GP) priors that specify initial beliefs on functions. However, even with expert knowledge, it is non-trivial to quantitatively define a prior. This is especially true for hyperparameter tuning problems on complex machine learning models, where landscapes of tuning objectives are often difficult to comprehend. We seek an alternative practice for setting these functional priors. In particular, we consider the scenario where we have data from similar functions that allow us to pre-train a tighter distribution a priori. 
We detail what pre-training entails for GPs using a KL divergence based loss function, and propose a new pre-training based BO framework named \hyperbo. Theoretically, we show bounded posterior predictions and near-zero regrets for \hyperbo without assuming the ``ground truth'' GP prior is known. To verify our approach in realistic setups, we collect a large multi-task hyperparameter tuning dataset by training tens of thousands of configurations of near-state-of-the-art deep learning models on popular image and text datasets, as well as a protein sequence dataset. Our results show that on average, \hyperbo is able to locate good hyperparameters at least 3 times more efficiently than the best competing methods on both our new tuning dataset and existing multi-task BO benchmarks.

\end{abstract}
\vspace{.5em}

\newpage
\tableofcontents
\newpage
\section{Introduction}
\label{sec:intro}
Bayesian optimization (BO)~\citep{garnett2023bayesian} has been successfully applied in numerous real-world global optimization problems, ranging broadly from hyperparameter tuning~\citep{snoek2012practical, kotthoff2019auto} to chemical synthesis~\citep{griffiths2020constrained, shields2021bayesian}, drug discovery~\citep{pyzer2018bayesian}, aerospace engineering~\citep{lam2018advances}, robotics~\citep{driess2017constrained,wang17icra} and more. However, in some scenarios, BO has been reported to under-perform naive strategies including random search~\citep{li2017hyperband}. While recent collective efforts have shown that "Bayesian optimization is superior to random search"~\citep{turner2021bayesian}, we seek more understanding on why BO works in some hands but not others.

Many successful BO applications benefit from expert knowledge on characteristics of the function to be optimized and hands-on experience with BO on similar tasks in the past. Such knowledge or experience can give intuitions about a functional form of the problem and thus specifications of a functional prior, e.g., a Gaussian process (GP) with squared exponential kernels for smoothness~\citep{griffiths2008modeling, wilson2015human}. Sometimes people may be uncertain about their own understanding, and as a result they might choose to use a hierarchical model~\citep{malkomes2018automating,cowen2020empirical,snoek2012practical} or Bayesian neural nets~\citep{springenberg2016bayesian}, such that observed data can play a more important role in modeling. Despite having almost no information about a function, we can guess a generic prior from past experience with BO on other functions~\citep{turner2021bayesian}. 

In the absence of any knowledge or experience, it is reasonable to consider all mathematical functions mapping inputs to outputs as equally likely. In this state of complete uncertainty, there is no single optimization algorithm that is guaranteed to outperform others~\citep{schaffer1994conservation}. This limitation is commonly known as ``no free lunch for optimization''~\citep{wolpert1997no}. In the context of BO, it means no guarantee on performance if no informative prior is available. It is thus not surprising that we might encounter poor empirical performance~\citep{schulz2016quantifying} when using BO without a well-specified prior. 
Theoretically, existing no-regret results only hold if model misspecification is well under control~\citep{bogunovic2021misspecified, berkenkamp2019no}.


\begin{figure}
    \centering
    \includegraphics[width=1.\textwidth]{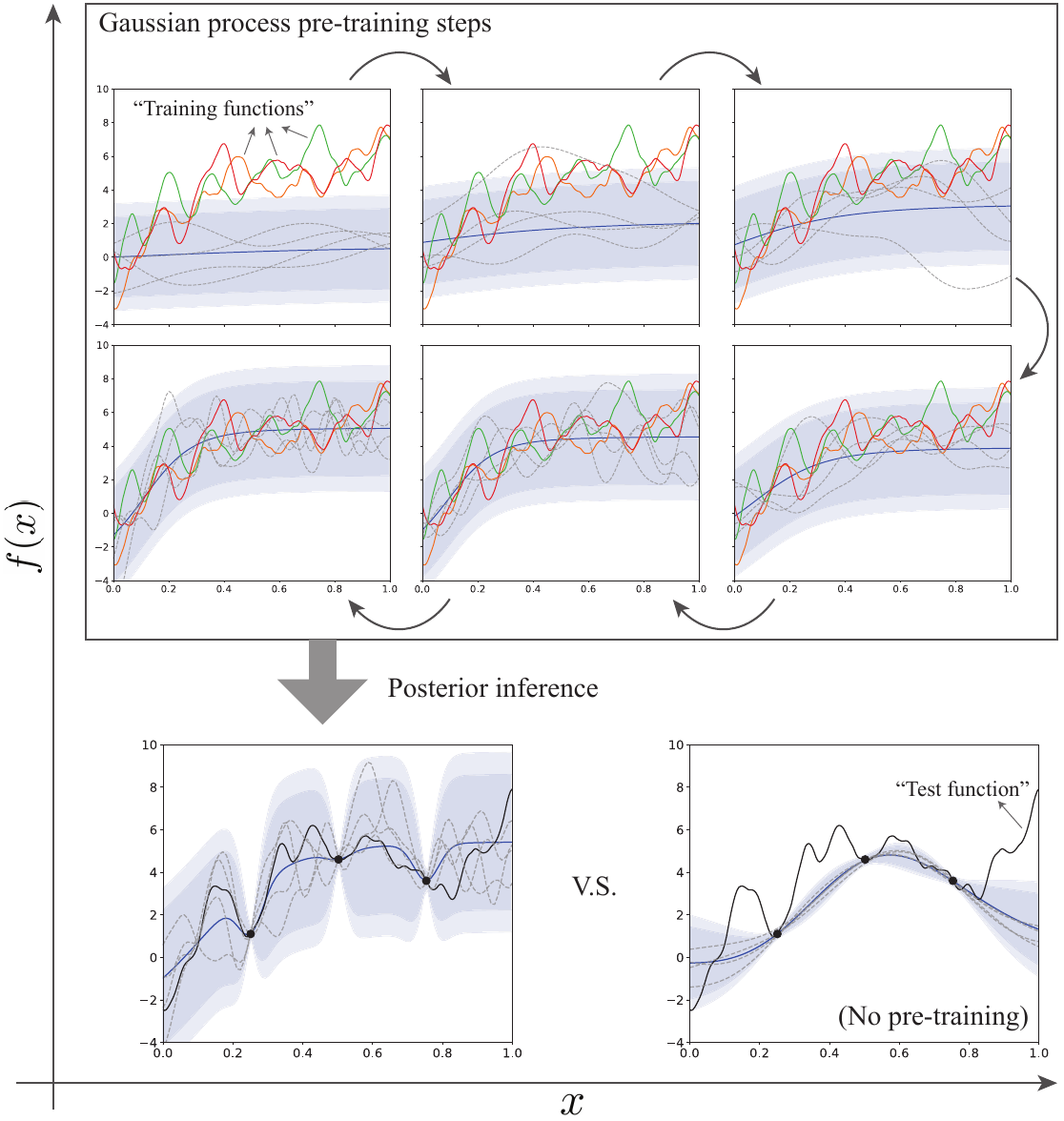}
    \caption{During pre-training, we optimize a Gaussian process (GP) such that it can gradually generate functions (illustrated as grey dotted lines) that are similar to the training functions. The similarity manifests in individual function values and correlations between function values indicated by smoothness and wiggliness. The blue line illustrates the mean function of the GP and the shaded areas are the $99\%$ and $95\%$ confidence intervals. For an unknown test function, we can derive a posterior conditioned on observed datapoints (illustrated as black dots) and the pre-trained GP prior. Compared to a GP fit to observations without pre-training, the pre-trained GP posterior captures the test function much better, which is a critical prerequisite for Bayesian optimization.}
    \label{fig:intro_summary}
\end{figure}

 Barriers of understanding on priors from a target domain and enough experience with BO can often turn away potential practitioners even within the machine learning (ML) community~\citep{bouthillier:hal-02447823}. For example, one of the most challenging domains for quantifying priors in BO is real-world hyperparameter tuning problems for modern deep learning models \citep[e.g., ResNet50 from][]{he2016deep} and large-scale datasets \citep[e.g., ImageNet from][]{imagenet}. For those large models~\citep{he2016deep,raffel2019exploring, brown2020language}, it is especially difficult to understand the landscapes of tuning objectives. %
 Even for experts with relevant experience, it is hard to pin down what exactly this prior looks like, and researchers often have to study effects of hyperparameters one at a time~\citep{sutskever2013importance, zhang2019algorithmic,choi2019empirical} given the complex structures of these problems. Since expert interventions on priors are almost unobtainable, use of BO has been hindered on these challenging but impactful tasks.

We would like to make BO methods more accessible by freeing practitioners from manually translating their own abstract beliefs into a quantitative Bayesian prior. We seek to automate the prior determination process by \emph{pre-training GP priors} on data that are available on different but related tasks. Note that pre-training is also known as prior learning and can be considered a version of meta learning~\citep{UrgenSchmidhuber1995, Baxter1996, minka1997}. While the term is often used in deep learning, we contextualize \emph{pre-training} for GPs in this work with a KL divergence based loss function, and use pre-trained GPs to bypass manual quantification of priors.

We hereby propose \hyperbo: a BO framework with pre-trained GP priors. \hyperbo is an enhancement of traditional BO methods without the requirement for practitioners to quantify their beliefs on functions. Instead, it is sufficient to specify which existing tasks are relevant and point to past evaluations on each of the functions corresponding to the existing tasks. These functions construct our training dataset for pre-training GPs. Figure~\ref{fig:intro_summary} illustrates the iterative training steps to obtain a pre-trained GP prior and use it to derive posteriors for BO. 

In essence, \hyperbo empowers BO to overcome ``no free lunch'' and unlock the full potential of Bayesianism by creating informative priors. One of the key advantages of \hyperbo is that we can guarantee success in terms of regret bounds under mild conditions, as long as the training functions can be viewed as samples from the unknown ground truth GP.  
 Moreover, by replacing manual prior quantification with abstract identification of training functions, \hyperbo streamlines the BO interface and allows easier use of BO for complex functions.

Empirically, we studied \hyperbo on challenging modern ML tuning problems. To fill the vacancy of relevant ``experience'' for such problems, we collected PD1, a large multi-task hyperparameter tuning dataset, by training tens of thousands of configurations of near-state-of-the-art models~\citep{he2016deep, raffel2019exploring, brown2020language} on popular image and text datasets, as well as on a protein sequence dataset. %
With the PD1 tuning dataset for pre-training, we evaluated \hyperbo on deep learning optimizer tuning problems, and on average, \hyperbo achieved at least 3 times speedup than the best alternative method in terms of BO iterations needed to obtain best validation accuracy. Besides PD1, we also benchmarked the performance of \hyperbo on HPO-B~\citep{pineda2021hpob}, a collection of 16 multi-task BO benchmarks for classic machine learning models, and on average, \hyperbo obtained at least 10 times speedup than competitive baselines.

\hyperbo is related to our prior work~\citep{wang2018regret, kim2019learning}. Both~\citet{kim2019learning} and ~\citet{wang2018regret} were motivated by robot manipulation tasks where finite domains are sufficient. \citet{wang2018regret} considered compact domains but the only possible modeling choice is Bayesian linear regression due to the requirement of defining a GP with finite parameters. The key differences and unique contributions of this work are: 
\begin{enumerate}
    \item Significant new insights on a program based view of BO and how \emph{pre-training} is consistent with the system of Bayesian belief reasoning~(\S\ref{sec:explain}). Accordingly, we define a principled loss function and provide a unified view of two simple yet effective approximations for pre-training GPs as learned functional priors~(\S\ref{ssec:objective_overview}).
    \item Substantially relaxed assumptions on data availability and modeling choices. Our new framework is now compatible with any type of GP on both discrete and continuous input domains. %
    \item Entirely new application domains on challenging hyperparameter tuning tasks. Aside from tuning hyperparameters on classic machine learning models~\citep{pineda2021hpob}, we applied our methods to tuning optimizer hyperparameters of modern deep learning models on popular image, text and protein sequence datasets. 
    \item Comprehensive analyses on the practicality of \hyperbo. We provide new insights on how theoretical understandings carry to real-world experiments through case studies. Our empirical results show the notable advantage of \hyperbo over strong baseline methods.
    \item We open-sourced the first large multi-task hyperparameter tuning dataset for modern deep learning models. We spent roughly 12,000 machine-days to collect hyperparameters evaluations by training tens of thousands of configurations of near-state-of-the-art models on various scales of data, ranging from millions of images to billions of words. Together with our open-sourced code for \hyperbo, the released dataset ensures the reproducibility of our work\footnote{Both open-sourced code and dataset are available at \url{https://github.com/google-research/hyperbo}.}. More importantly, the dataset provides a realistic benchmark for multi-task BO, with open opportunities to explore detailed metrics for each training step and other auxiliary information.
\end{enumerate}

Next, we discuss related work in \S\ref{sec:related}, explain foundational concepts of BO in \S\ref{sec:explain}, formulate our problem in \S\ref{sec:pf} and introduce the core GP pre-training method in \S\ref{ssec:objective_overview}. %
In \S\ref{sec:bo}, we present our \hyperbo framework for the black-box function optimization. To understand the implications of substituting the ground truth with a pre-trained GP model, we provide theoretical insights on the asymptotic properties of the pre-trained model for posterior inference, and show regret bounds to explain when it is a good idea to use the pre-trained model for BO. In \S\ref{sec:exp}, we provide empirical evidence showing promising results of \hyperbo for real-world black-box function optimization tasks. Finally, we discuss fully Bayesian interpretations and open problems of \hyperbo in \S\ref{sec:discuss}, and conclude in \S\ref{sec:conclu}.

\section{Literature Review}
\label{sec:related}
There is a rich literature of innovative methodologies to improve the efficiency of BO given related tasks or additional context.  Here we discuss the most closely related work and explain why these do not solve the specific scenario which we envision.  Specifically, our goal is a methodology that is scalable enough to share information across thousands of tasks, each with potentially hundreds of observations, such as in the context of a large BO service or library.

Pre-training and prior learning is directly related to meta learning, learning to learn and learning multiple tasks~\citep{UrgenSchmidhuber1995, Baxter1996, minka1997, caruana1997multitask}. Such meta learning and multi-task learning ideas can be naturally used to learn a GP prior~\citep[Chapter 5]{rasmussen2006gaussian}. We use the word pre-training to refer to supervised pre-training, which is a general approach in the deep learning community~\citep{girshick2014rich, donahue14decaf, devlin2018bert} to transfer knowledge from prior tasks to a new task. The same as pre-training deep features on a variety of tasks, \citet{wang2018regret} proposed to learn a GP prior by learning the basis functions, treating the independent function outputs as individual heads of a neural network. More recently, there has been theoretical advancement in the PAC-Bayesian framework to understand meta learning~\citep{rothfuss2021pacoh} for GPs.

Several methods, including that which \hyperbo extends, refer to their method as ``meta BO''~\citep{wang2018regret, volpp2020meta}. In this work we use the term \emph{meta BO} more generally to refer to the class of BO methods that use data from existing tasks to optimize a new task.  Since standard BO is a learning process, it is consistent to call those methods meta BO methods given that they learn how to learn. Under this viewpoint, meta BO approaches also include multi-task BO~\citep{swersky2013multi, yogatama2014efficient, poloczek2017multi} and transfer learning BO methods, e.g., based on contextual GPs~\citep{krause2011contextual, bardenet2013collaborative, poloczek2016warm},  quantiles~\citep{salinas2020quantile} or ensembles of GPs~\citep{feurer2018practical,wistuba2018scalable}. Some meta BO methods have also been studied for hyperparamter tuning tasks in machine learning~\citep{feurer2015efficient, salinas2020quantile}.

To enable meta learned models to transfer knowledge from prior tasks to a new task in BO, assumptions need to be made to capture the connections among tasks. For both multi-task and contextual BO methods, such as \cite{krause2011contextual, swersky2013multi, tighineanu2022transfer}, the connections are modeled directly through computing the similarities between tasks. These approaches typically scale cubically in both the number of tasks and observations in each task, meaning that they cannot gracefully scale across both without heavy approximations. When assuming that all inputs are equal across tasks, multi-task BO~\citep{swersky2013multi} can be sped up using a Kronecker decomposition of the kernel to a task kernel and an input kernel which can be inverted separately; a similar assumption is made by~\citet{wang2018regret}.  In comparison, \hyperbo establishes the connections among tasks by positing a shared prior which renders the tasks conditionally independent. As a result, \hyperbo scales linearly in the number of tasks (see \S\ref{ssec:complexity}), which facilitates efficient model pre-training.

Motivated by robot learning problems, \citet{kim2017learning,kim2019learning} started a different thread of meta BO literature, with a goal to transfer knowledge among robot manipulation tasks. Each task is a BO problem that optimizes a scoring function by sequentially selecting search strategies from a finite set. \citet{kim2017learning,kim2019learning} noted that the similarities among tasks are very difficult to model, since a slight change in the state can completely change the function landscape. To address this issue, they introduced a simple but elegant approach: estimating the correlations between scores of different search strategies; i.e., modeling the similarities between inputs as opposed to tasks. 

\citet{wang2018regret} provided regret bounds for \citet{kim2017learning,kim2019learning} and extended it to Bayesian linear regression with neural net basis functions. 
Similar ideas were developed by \citet{perrone2018scalable, wistuba2021few} for tuning the hyperparameters of machine learning models. \citet{wang2018regret} can be viewed as a generalization of these approaches in that \citet{perrone2018scalable} and \citet{wistuba2021few} only use zero means. 
As shown in \citet{kim2017learning,kim2019learning}, a flexible mean function is important for learning the initial datapoints to acquire. \citet{wistuba2021few} overcomes this initialization issue by using a data-driven evolutionary algorithm to warm-start the initialization. We opt to parameterize a flexible mean function using a neural network, allowing for end-to-end optimization. 

Although different terms are used, \citet{wang2018regret} and \citet{perrone2018scalable} concurrently proposed the idea of learning parameters of GP priors from multi-task datasets, while \citet{wang2018regret} was the first to clarify the assumptions of conditionally independent multi-task functions and show regret bounds for BO with an unknown GP prior.

Our proposed pre-training objectives are related to the objective functions in variational inference for approximating GPs~\citep{titsias2009variational, burt2020understanding}. Our key idea on pre-training is to minimize the KL divergence between the unknown ground truth GP and an approximate. On the other hand, variational inference in functional spaces~\citep{sun2019functional, burt2020understanding} generally considers the KL divergence between an approximate and a posterior, which aims to improve computational efficiency for posterior predictions given large scale observations. While there are significant differences in goals and methods, the objective functions all boil down to the KL divergence for functional distributions. More details on our method can be found in \S\ref{ssec:objective_overview}.

\section{Background}
\label{sec:explain}

 While there are other interpretations, we take the viewpoint of \emph{artificial intelligence} and consider \textit{Bayesian optimization} (BO) as a study of how an intelligent machine optimizes a numerical function: making sequential decisions on data acquisition by reasoning about the machine's \emph{posterior beliefs} on the function. 
The beliefs are expressed as \textit{Bayesian} probabilities, which root in logic, common sense and rational reasoning about plausibility~\citep{jaynes2003probability}. Decisions on data acquisition involve choosing the inputs to query the function and observing their corresponding outputs. %

As a mathematical tool, a popular version of BO is \textit{Gaussian process (GP) optimization}~\citep{srinivas2009gaussian, contal2014gaussian, wang2016est}, which uses \textit{GPs} as the Bayesian beliefs on the function. Popular decision making criteria on data acquisition include probability of improvement~\citep{kushner1964}, expected improvement~\citep{mockus1974}, upper confidence bound~\citep{auer2002b}, entropy search~\citep{hennig2012}, etc.

BO has been used to optimize expensive black-box functions and to solve experimental design problems. To understand how to apply BO to real-world applications, we introduce \emph{the practitioner}, e.g., a scientist or an engineer, who delegates decision making to the intelligent machine. To take over the optimization process, the machine requires the practitioner to assign a prior on information related to the function. 

The prior used by the machine should reflect the practitioner's understanding, i.e., the practitioner's posterior belief, about the function based on their past experience with related functions. %
Assigning such priors typically requires the practitioner to quantitatively encode their own belief in machine languages, which is not always clear given the stark difference between how humans think and how machines operate.

\vspace{.5em}
{\it{Writing a program to assign the prior.}}\; The practitioner's belief is based on their past experience, or more specifically, the data they observed on functions relevant to, but not necessarily the same as, the black-box function of interest. 

In this work, we seek to circumvent the hurdle of manually specifying the prior by writing a prior assignment program that works with the past observed data directly. Thus, the practitioner only needs to identify the data (partitioned to observations on different functions) that is related to the function they would like to optimize. This program takes over the task of prior assignment, which has a contract to take in user specified data as the input and output a probability distribution consistent with the ``ground truth'' practitioner's belief on the function. Note that the practitioner is typically non-Bayesian and their belief on the function does not necessarily reflect a posterior. 

\vspace{.5em}
{\it{A wrapper over the Bayesian component.}}\; We have introduced two entirely separate programs: one does BO given a prior, and the other produces a probability distribution to match the underlying prior on the function. The former is a reasoning and decision making process governed by Bayes rules. The latter serves as the prior assignment program which, in this work, is not fully Bayesian. A system composed of these two programs bypasses the need to define a specific prior in BO and instead, derives the prior from data before doing any Bayesian inference or reasoning.

\vspace{.5em}
{\it{Pre-training as the prior assignment program.}}\; The prior assignment program needs to train a probabilistic model on data observed over different functions and assign the trained model as the prior for the black-box function of interest. We use \emph{pre-training} to describe this program, as it is fundamentally the same practice as (supervised) pre-training in the deep learning literature: a model is trained on a range of tasks and the learned feature representations (i.e., basis functions) are preserved for unseen tasks for fine-tuning. 

Note that ``pre-training'' has other names in the literature. \citet{Baxter1996} explained learning a prior as ``bias learning'' from a Bayesian viewpoint. \citet{minka1997} more explicitly described it as learning bases of a probabilistic model from a dataset of tasks (each with some datapoints) and then applying the learned model as the prior for an unseen task. For ease of understanding, we use the term ``pre-training'' in this paper.

\section{Problem Formulation}
\label{sec:pf}

We follow the \emph{Gaussian process (GP) optimization} paradigm: given a real-valued function $f$ defined over a compact, hyper-rectangular space $\mathfrak X \subset \R^d$, we seek an $x \in \mathfrak X$ that maximizes $f$ with as few evaluations on $f$ as possible.

Our assumptions are similar to~\citet{minka1997}, where our training dataset is a set of \iid sets of non \iid datapoints, i.e., a dataset consisting of sets of observations on \textbf{training functions} $f_1, \cdots, f_N$.
Assumption~\ref{asp:iid} emphasizes that our training functions and test functions are all \iid samples from an unknown GP. %
Assumption~\ref{asp:noise} describes that observations are perturbed by \iid Gaussian noises with unknown variance. 

\begin{assumption}\label{asp:iid}
 There exists a non-degenerate GP $\GP(\mu^*, k^*)$ with \textbf{unknown} mean function~$\mu^*: \mathfrak X\rightarrow \R$ and \textbf{unknown} kernel~$k^*: \mathfrak X\times \mathfrak X\rightarrow \R$, such that the training functions $f_1, \cdots, f_N$ and the \textbf{test function} $f$ are all \iid samples from $\GP(\mu^*, k^*)$.
\end{assumption}

\begin{assumption}\label{asp:noise}
There exists a Gaussian distribution $\N(0, \sigma_*^2)$ with \textbf{unknown} variance $\sigma_*^2\in\R^+$, such that for any function $g\sim \GP(\mu^*, k^*)$ and any input $x \in \mathfrak X$, the observed function value $y$ is perturbed by \iid additive Gaussian noise $\N(0, \sigma_*^2)$, i.e., $y\sim \N(g(x), \sigma_*^2)$.
\end{assumption}
We use $[n]$ to denote $\{1,\cdots,n\}, \forall n\in \Z^+$. Let $M_i$ be the number of observations we have for function $f_i$ where $i\in[N]$. We use $y\sim \N(g(x), \sigma_*^2)$ as a short hand to describe the conditional distribution for $y \mid g$. Provided input $x^{(i)}_{j}\in \mathfrak X, i\in[N], j\in[M_i]$, the observed function value is $y^{(i)}_{j}\sim \N\left(f_i(x^{(i)}_{j}), \sigma_*^2 \right)$ by Assumption~\ref{asp:noise}.

\begin{figure}
\centering
\begin{tikzpicture}
\tikzstyle{main}=[circle, minimum size = 4mm, thick, draw =black!80, node distance = 6mm]
\tikzstyle{para}=[circle, minimum size = 5pt, inner sep=0pt]
\tikzstyle{connect}=[-latex, thick]
\tikzstyle{box}=[rectangle, draw=black!100]

  \node[para, fill = black!100] (mu) [label=left:$\mu^*$] {};
  \node[main] (f) [below right=0.4cm and 1.3cm of mu,label=below:$f_i$] { };
  \node[para, fill = black!100] (k) [below=of mu,label=left:$k^*$] { };
  \node[main, fill = black!10] (y) [right=1cm of f,label=below:$y^{(i)}_{j}$] { };
  \node[para, fill = black!100] (x) [above=of y,label=right:$x^{(i)}_{j}$] { };
  \node[para, fill = black!100] (sigma) [right=2.2cm of y, label=right:$\sigma^2_*$] { };

  \path 
        (sigma) edge [connect] (y)
		(mu) edge [connect] (f)
		(k) edge [connect] (f)
		(f) edge [connect] (y)
		(x) edge [connect] (y);
  \node[rectangle, inner sep=1cm, fit= (x) (y), label=below right:$M_i$, xshift=-6.5mm, yshift=-4mm] {};
  \node[rectangle, inner sep=0.9cm, draw=black!100, fit= (x) (y), xshift=2mm, yshift=-4.5mm] {};
  \node[rectangle, inner sep=0.7cm, fit= (x) (y) (f), label=below right:$N$, yshift=-5mm, xshift=5.5mm] {};
  \node[rectangle, inner sep=1.2cm, draw=black!100, fit= (x) (y) (f), yshift=-5mm, xshift=5mm] {};
\end{tikzpicture}\caption{The generating process of our training data: for each $i\in [N]$, $f_i \sim \GP(\mu^*, k^*)$, and for each $j\in[M_i]$, $y^{(i)}_{j}\sim \N\left(f_i(x^{(i)}_{j}), \sigma_*^2 \right)$, where mean function $\mu^*$, kernel function $k^*$ and noise variance $\sigma_*^2$ are unknown.}
\label{fig:gpgraph}
\end{figure}
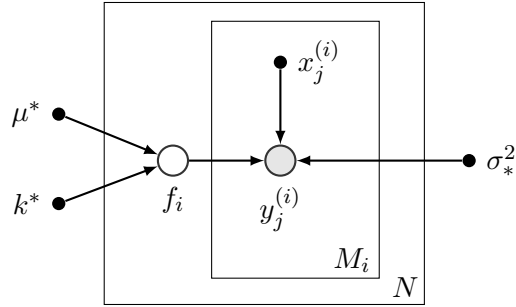

Taken together, the collection of \textbf{\emph{sub-datasets}} $D_{f_{i}} = \{(x^{(i)}_{j}, y^{(i)}_{j})\}_{j=1}^{M_i}$ constructs the \textbf{\emph{training dataset}} $D_N = \{D_{f_{i}} \}_{i=1}^N$. Figure~\ref{fig:gpgraph} shows the graphical model that illustrates the generating process of the training dataset $D_N$. Note that we explicitly assume that the \textbf{\emph{ground truth}} $\mu^*, k^*, \sigma^2_*$ exist but they are all \textbf{\emph{unknown}}.

\vspace{.5em}
{\it{Metrics.}}\;
For simplicity, we focus on sequential evaluations on the \emph{test function} $f$ where only one input is chosen for evaluation in each iteration. %

For $T$ iterations of Bayesian optimization (BO) on function $f$, we accumulate a set of observations $D_f = \{(x_t, y_t)\}_{t=1}^T$, where $y_t \sim \mathcal N\left(f(x_t), \sigma_*^2\right)$ and $T$ is a positive integer.

In hindsight, we can evaluate the quality of BO using the \emph{simple regret} metric: $R_T = \max_{x\in \mathfrak X} f(x) - f(\hat x)$, where $\hat x$ is the recommended best input. 

There are various ways of setting $\hat x$ based on the observations $D_f$ or the posterior of function $f$. In this work, we use the input that achieves the best evaluation: $\hat x = x_{T'}; T' = \argmax_{\tau\in[T]}y_\tau$.

\vspace{.5em}
{\it{Notations of models.} }\;
We use $\GP(\mu, k\circ{\sigma^2})$ to denote a GP model with mean function $\mu$ and a kernel function with perturbed diagonal terms: 

    $$k\circ{\sigma^2}(x_j, x_{j'}) = k(x_j, x_{j'}) + \mathbbm{1}_{j\equiv j'}\sigma^2,$$
 where $j$ and $j'$ are indices of the inputs. 

We also use the short-hands $k(x) = k(x, x)$ and $k\circ{\sigma^2}(x) = k(x, x) + \sigma^2$ for simplicity. 

For any collection of inputs $\vx = [x_j]_{j=1}^m \in \R^{m\times d}$ and an input $x'\in\R^d$, we denote the column vector of mean function values as $\mu(\vx) = [\mu(x_j)]_{j=1}^m\in\R^{m}$, the column vector of kernel function values between $\vx$ and $x'$ as $k(\vx, x')= [k(x_j, x')]_{j=1}^m \in\R^{m}$, and the Gram matrix as $k\circ{\sigma^2}(\vx) = [k(x_j, x_{j'})]_{j\in[m], j'\in[m]} + I\sigma^2 \in \R^{m\times m}$. 

\vspace{.5em}
\underline{\it Remarks.}\; In the language of \S\ref{sec:explain}, the practitioner's belief, a.k.a. the unknown ground truth prior of the intelligent machine, is $\GP(\mu^*, k^*\circ{\sigma_*^2})$, which describes the probability distributions of functions and observation noise. The contract of the prior assignment program is to take in as input \emph{training dataset} $D_N$ and output a \emph{pre-trained GP} $\GP(\hat\mu, \hat k\circ{\hat \sigma^2})$. Ideally, the pre-trained model should be very close to the ground truth prior. The intelligent machine can then use the pre-trained $\GP(\hat\mu, \hat k\circ{\hat \sigma^2})$ as the ``prior distribution'' in a BO program for test function $f$. 

\begin{exmp}
In machine learning hyperparameter tuning applications, the task is to find the best configuration of hyperparameters to train a specific machine learning model on a particular dataset, e.g., training a ResNet~\citep{he2016deep} on ImageNet~\citep{imagenet}. For each combination of a machine learning model and a dataset, there is an underlying function mapping from configurations of hyperparameters to an evaluation metric such as the error rate. The training functions $f_1, \cdots, f_N$ correspond to those functions that the practitioner observed in the past. The test function $f$ maps from configurations of hyperparameters to the evaluation metric of a new combination of a model and a dataset. 
\end{exmp}

\section{Pre-training Gaussian Processes}
\label{ssec:objective_overview}

In this section, we will describe our KL divergence based objective for pre-training Gaussian processes (GPs). We will introduce two approximations for the objective. First is an estimator that we call the empirical KL (EKL) loss. The second reduces to the negative log-likelihood (NLL) loss. These have different strengths depending on the property of the data. When each task has an observation at each input, EKL can naturally learn and make use of the correlations between function values. We find that this improves its empirical performance. 
NLL is more flexible, and useful for cases where the observations are made at different locations across tasks.

\subsection{Pre-training Objective}
\label{ssec:pre-train_obj}
We use the KL divergence between the ground truth $\GP(\mu^*, k^*\circ{\sigma_*^2})$ and a model $\GP(\mu, k\circ{\sigma^2})$ as the loss function. By Theorem 1 of \cite{sun2019functional}, our loss function is
\begin{align}\label{eq:loss}
\L(\mu,\, k \circ \sigma^2) &\coloneqq D_{\KL}\Bigl(\GP(\mu^*, k^*\circ{\sigma_*^2}), \GP(\mu, k\circ{\sigma^2})\Bigl)  \\
& = \sup_{\vx \subseteq \mathfrak X, \card(\vx) < \infty} D_{\KL}\Bigl(\N(\mu^*(\vx), k^*\circ{\sigma_*^2}(\vx)), \N(\mu(\vx), k\circ{\sigma^2}(\vx))\Bigl). \label{epsilonnetkl}
\end{align}
In Eq.~\ref{epsilonnetkl}, $\vx$ is a collection of inputs from $\mathfrak X \subset \R^d$ and $\card(\vx)$ is the cardinality of $\vx$. While it is often intractable to compute this loss function~\citep{burt2020understanding, sun2019functional}, it is natural to approximate the loss by truncating $\mathfrak X$ to a finite set of inputs~\citep{sun2019functional}. 
If the domain $\mathfrak X$ is finite, the loss in Eq.~\ref{eq:loss} becomes the KL divergence between two multivariate Gaussian distributions, and the supremum in Eq.~\ref{epsilonnetkl} is obtained by setting $\vx \equiv \mathfrak X$.

Our goal is to obtain the ``pre-trained'' GP model, $\GP(\hat\mu, \hat k\circ{\hat\sigma^2})$, by minimizing the loss function subject to positive definite kernel and positive noise variance; that is,
\begin{align}\label{eq:lossobj}
\hat \mu,\, \hat k \circ \hat \sigma^2\; =   \;\underset{\mu,\, k \circ \sigma^2}{\argmin}\quad &\L(\mu,\, k \circ \sigma^2)\\
 s.t.\quad & k > 0, \sigma^2 > 0 \nonumber
\end{align}
Without loss of generality, we assume there is one $\argmin$ solution\footnote{The solution depends on the space of functions to optimize the loss function over, and it is possible that a minimum does not exist. However, for practical engineering setups and choices of models, it is reasonable to assume finding one (approximate) solution is possible.} to the minimization problem in Eq.~\ref{eq:lossobj}. We slightly abuse notation by minimizing the loss over the mean function $\mu$ and perturbed kernel $k\circ \sigma^2$ without specifying their search spaces. In practice, the search spaces for these functions depend on specifications made by the practitioner. For example, we may try to minimize the loss by comparing its values on two different mean functions, e.g., $\mu(x) = |x|$ and $\mu(x) = |x|^2$, where the search space for $\mu$ contains two functions. 

More generally, minimizing the loss in Eq.~\ref{eq:loss} can involve searching over function structures and/or optimizing the parameters of functions~\citep{malkomes2016bayesian, malkomes2018automating}. For high dimensional problems, we might prefer additive kernels for interpretability, and we can learn the additive structure as well~\citep{wang2017batched, gardner2017discovering, rolland2018high}. For structured inputs, one may adopt specialized kernels, e.g., graph kernels~\citep{vishwanathan2010graph}, convolutional kernels~\citep{van2017convolutional}, etc. Note that it is not necessary to require the mean function $\mu$ or perturbed kernel $k\circ \sigma^2$ to be parametric. For example, they can be specified with memory based machine learning models~\citep{russell2010artificial,daelemans2005memory}. 

If we use parametric mean function $\mu$ and perturbed kernel $k\circ \sigma^2$ with fixed structures, we only need to optimize their parameters. \citet{wistuba2021few} proposed to use the Adam optimizer~\citep{kingma2014adam}; \citet{wang2018regret} suggested solving linear systems; and \citet{perrone2018scalable} recommended L-BFGS~\citep{liu1989limited}. The choice of optimizers may also depend on the parametric form. %
For our experiments, we defined flexible search spaces of functions using neural networks and constructed positive definite kernels on encoded representations of inputs. More details can be found in \S\ref{sec:exp}. 

Now that it is clear there exist methods to optimize over spaces of functions, we investigate a more pressing issue on our objective in Eq.~\ref{eq:lossobj}: we do not know the ground truth model $\GP(\mu^*, k^*\circ{\sigma_*^2})$ and cannot compute the loss function. 
 
In \S\ref{ssec:reg}, we introduce EKL that directly computes the KL divergence by estimating a multivariate Gaussian distribution induced by $\GP(\mu^*, k^*\circ{\sigma_*^2})$. In \S\ref{ssec:nll}, we use NLL: an expansion of the KL divergence to simulate the loss with training functions. We analyze the computational complexity of the approximated loss functions in \S\ref{ssec:complexity}. In \S\ref{ssec:nll_ekl_relation}, we explain the relations between EKL and NLL.

\subsection{Empirical KL Divergence (EKL)}
\label{ssec:reg}
The key idea of the EKL approximation is to compute the KL divergence between an empirical estimate of the ground truth and our model. Thus, it is possible to directly compute the KL divergence by manipulating the training data.

\subsubsection{Case Study: Observing Training Functions on the Same Inputs}
\label{sssec:matchingdata}
For simplicity, we first consider the case where the training dataset $D_N = \{D_{f_i}\}$ is a ``matching-input'' dataset, 
$D_{f_i} = \{(x_j, y^{(i)}_{j})\}_{j=1}^M$, where $M$ is the number of shared inputs across $N$ training functions. 

Dataset $D_N$ is composed of queries over the training functions $f_{1}, \cdots, f_{N}$ at the same set of input locations $\vx = [x_j]_{j=1}^{M} \in \R^{M\times d}$. We can re-organize the datapoints in $D_N$ as follows.

\begin{table}[h]
\centering
\large
\label{tab:ekl_data}
\begin{tabular}{c|*{5}c}
\toprule
\,\, &  $f_{1}$ & $\cdots$ & $f_{i}$ & $\cdots$ & $f_{N}$ \\
\midrule
$x_1$ &   $y^{(1)}_{1}$ & $\cdots$ & $y^{(i)}_{1}$ & $\cdots$ & $y^{(N)}_{1}$ \\ 
$\vdots$   &  $\vdots$ & $\ddots$ & $\vdots$ & $\ddots$ & $\vdots$ \\ 
$x_j$ &   $y^{(1)}_{j}$ & $\cdots$ & $y^{(i)}_{j}$ & $\cdots$ & $y^{(N)}_{j}$ \\
$\vdots$   &  $\vdots$ & $\ddots$ & $\vdots$ & $\ddots$ & $\vdots$\\ 
$x_M$ &   $y^{(1)}_{M}$ & $\cdots$ & $y^{(i)}_{M}$ & $\cdots$ & $y^{(N)}_{M}$ \\ 
\bottomrule
\end{tabular}
\quad\quad
\begin{tabular}{c}
\begin{small}
$ \vy_i = 
\begin{bmatrix} 
   \;   y^{(i)}_1 \; \\ 
      \vdots  \\
     \;    y^{(i)}_M\;
   \end{bmatrix}
$
\end{small}
\end{tabular}
\end{table}
For each training function $f_i$, we have $M$ observations which correspond to entries of a column in the illustration, i.e.,$\vy_i = [y_{j}^{(i)}]_{j=1}^{M} \in \R^{M}$. Given that each function $f_i\overset{i.i.d.}{\sim} \GP(\mu^*, k^*)$ and the observations $y_{j}^{(i)}\overset{i.i.d.}{\sim} \N(f_i(x_j), \sigma_*^2)$, we have $\vy_i\overset{i.i.d.}{\sim} \N(\vmu^*, \Sigma^*)$, where the mean vector is $\vmu^*=\mu^*(\vx)$ and the covariance matrix is $\Sigma^* = k^*\circ\sigma_*^2(\vx)$.
The distribution $\N(\vmu^*, \Sigma^*)$ captures the marginals of the ground truth $\GP(\mu^*, k^*\circ{\sigma_*^2})$.

\subsubsection{Approximation by Estimation}
\label{sssec:kl-approx}
We perform a two-step approximation of the loss function in Eq.~\ref{eq:loss}: (1) estimate the marginal ground truth distribution and (2) approximate the KL divergence. 

From the observations on all training functions, we can estimate the unknown mean vector $\vmu^*$ and covariance matrix $\Sigma^*$. By concatenating the columns $\vy_i$ together horizontally, we obtain a matrix of observations on all training functions: $Y =[\vy_i]_{i=1}^N \in \R^{M\times N}$.

In this work, we adopt maximum likelihood estimation (MLE) and get the estimated mean vector and covariance matrix as follows, 
\begin{align}
    \tilde \vmu = \frac{1}{N} Y 1_{N} \in \R^M\;\;\; \text{and}\;\;\; \tilde \Sigma = \frac{1}{N}(Y -\tilde \vmu1_N\T) (Y- \tilde \vmu1_N\T)\T \in\R^{M\times M}, \label{eq:estimate_mean_cov}
\end{align}
where $1_N$ is a column vector of size $N$ filled with $1$s. Other kinds of estimators can be used to replace MLE, such as the unbiased sample mean and covariance estimator.

The following \emph{empirical KL divergence} (EKL) approximates the loss function (Eq.~\ref{eq:loss}) with an empirical estimation of the ground truth model,
\begin{align}\label{eq:kl}
\L(\mu,\, k \circ \sigma^2) &\coloneqq D_{\KL}\Bigl(\GP(\mu^*, k^*\circ{\sigma_*^2}), \GP(\mu, k\circ{\sigma^2})\Bigl)\nonumber\\
&\approx D_{\KL}\biggl( \N(\vmu^*, \Sigma^*), \N\Bigl(\mu(\vx), k\circ{\sigma^2}(\vx) \Bigl)\biggl)\nonumber\\
&\approx D_{\KL} \biggl( \N(\tilde\vmu, \tilde\Sigma),  \N\Bigl(\vmu, \Sigma \Bigl)\biggl),
\end{align}
where $\vmu = \mu(\vx)$ and $\Sigma=k\circ{\sigma^2}(\vx)$. 
EKL in Eq.~\ref{eq:kl} measures the difference between the estimated multivariate Gaussian in Eq.~\ref{eq:estimate_mean_cov} and a model $\GP(\mu, k\circ{\sigma^2})$ evaluated on the inputs $\vx=[x_j]_{j=1}^{M}$. Pre-training by minimizing EKL means aligning the model with an intermediate estimate of the ground truth GP on finite set of points. The KL divergence is well-defined for two non-degenerate Gaussians. We can insure the non-degeneracy of the model $\GP(\mu, k\circ{\sigma^2})$ by constraining $k\circ{\sigma^2}$ to be positive definite (Eq.~\ref{eq:lossobj}); i.e., covariance matrix $\Sigma=k\circ{\sigma^2}(\vx)$ is non-singular. 

However, the estimated distribution $\mathcal N(\tilde{\vmu}, \tilde\Sigma)$ is often degenerate since obtaining more datapoints on each task can be easier than defining and obtaining more training tasks. For example, our multi-task hyperparameter tuning benchmark in \S\ref{ssec:data} has roughly 500 matching-input datapoints, but only 23 tasks (different model and dataset combinations). This is also true for other types of problems, such as robot skill learning \citep{wang2021learning}, where the tasks are robot skills like scoop, pour etc, and the datapoints are score evaluations of control parameters. 
The tasks need to be defined and implemented carefully, but obtaining datapoints only requires repetitive experimentation. Hence, it is critical for us take into account that the estimated distribution $\N(\tilde\vmu, \tilde\Sigma)$ can be degenerate.

We define $D_\KL$ in Eq.~\ref{eq:kl} as the KL divergence on the support of $\N(\tilde\vmu, \tilde\Sigma)$. This is a generic definition of KL divergence that is inclusive of both degenerate and non-degenerate distributions.

To compute the support, we construct a matrix $A$ such that $\tilde \Sigma = AA\T, A\in \R^{M\times r}$ where $r\leq M$ is the rank of $A$ and $\tilde \Sigma$. 
In practice, we use SVD to find a solution of matrix $A$ by factorizing $\tilde \Sigma$ as $\tilde \Sigma = V\Lambda V\T$ where $\Lambda\in \R^{r\times r}$ is a diagonal matrix with positive real-valued diagonal terms. We then set $A = V\sqrt{\Lambda}$ where $\sqrt{\Lambda}$ applies square root to all diagonal terms of $\Lambda$.  

The pseudoinverse of matrix $A$ is $A^+ = (A\T A)^{-1} A\T \in \R^{r\times M}$. We apply an affine transformation $\vv = A^+(\vy - \tilde\vmu)$ to map any observation vector $\vy \in \R^M$ to $\vv\in\R^r$, which allows us to obtain marginal distributions by dropping irrelevant dimensions. 

We can then compute $D_\KL$ as the KL divergence on the affine subspace of both distributions, $\N(\tilde\vmu, \tilde\Sigma)$ and $\N(\vmu, \Sigma)$, i.e.,
\begin{align} \label{eq:degenerate}
D_{\KL}\Bigl(\N(\tilde\vmu, \tilde \Sigma), \N(\vmu, \Sigma)\Bigl)
&= D_{\KL}\Bigl(\N(0, I), \N(A^+(\vmu-\tilde\vmu), A^+\Sigma(A^+)\T)\Bigl) \nonumber\\ 
&= \frac{1}{2}\left( \tr(\Sigma_p^{-1}) + (\vmu_p - \tilde \vmu_p)\T \Sigma_p^{-1}(\vmu_p - \tilde \vmu_p) + \ln|\Sigma_p| - r\right),
\end{align}
where $\vmu_p = A^{+}\vmu$, $\tilde\vmu_p = A^{+}\tilde\vmu$, $\Sigma_p = A^+\Sigma(A^+)\T$.  

Intuitively, the estimated Gaussian $\N(\tilde\vmu, \tilde \Sigma)$ is only able to reflect information on a reduced dimensional subspace of the $M$-dimensional variable $\vy\sim \N(\vmu^*, \Sigma^*)$ due to low rank. 
 What we can do is to make sure our model is aligned with the ground truth on that lower dimensional space, which can be achieved by minimizing  Eq.~\ref{eq:degenerate}. 

Note the solution for matrix $A$ is not unique. For any orthogonal matrix $B\in \R^{r\times r}$, $BB\T=I$, we have $\tilde \Sigma = ABB\T A\T = AA\T$, which means $AB$ is also a solution. However, the KL divergence is invariant under parameter transformation. More specifically, the transformation is $\vv' = (AB)^{+}(\vy-\tilde\vmu) = U\vv$ where $U = (B\T A\T A B)^{-1} B\T A\T A$. This is because $$(AB)^+ = (B\T A\T A B)^{-1} B\T A\T =  (B\T A\T A B)^{-1} B\T A\T A A^+.$$ Note that matrix $U$ is non-singular\footnote{This is because we can set $U^{-1} = (A\T A)^{-1}BB\T A\T A B$ such that $U U^{-1} = U^{-1} U= I$.}. Therefore, while matrix $A$ is not unique, the KL divergence in Eq.~\ref{eq:degenerate} remains the same.

For the non-degenerate case of the estimated distribution $\N(\tilde\vmu, \tilde\Sigma)$, we have $A^+ = A^{-1}$ and Eq.~\ref{eq:degenerate} recovers the standard definition of  KL divergence between non-degenerate multivariate Gaussians: 

\begin{align}\label{eq:nondegenerate}
D_{\KL}\Bigl(\N(\tilde\vmu, \tilde \Sigma), \N(\vmu, \Sigma)\Bigl) 
&= D_{\KL}\Bigl(\N(0, I), \N(A^{-1}(\vmu-\tilde\vmu), A^{-1}\Sigma(A^{-1})\T)\Bigl) \nonumber\\
&=\frac{1}{2}\left( \tr(\Sigma^{-1}\tilde \Sigma) + (\vmu - \tilde \vmu)\T \Sigma^{-1}(\vmu - \tilde \vmu) + \ln\frac{|\Sigma|}{|\tilde \Sigma|} - M\right). 
\end{align}
More details on the derivation for EKL can be found in \S\ref{app:equations}.


\subsubsection{Extensions to Generic Cases of Training Data}
\label{sssec:kl-extensions}
In this section, we explore the case where our dataset $D_N$ is not a ``matching-input'' dataset. That is, we have observations on different input locations for different training functions. Clearly, we cannot use the exact method for the case described in \S\ref{sssec:kl-approx}, but in fact, it is possible to heuristically transform our training data into a similar format as the ``matching-input'' dataset.

The core idea is to partition the input space into non-overlapping regions and merge the datapoints in each region to construct pseudo datapoints that align over training functions. 

More formally, we define a partition strategy, $\psi: \mathfrak X \to \mathfrak C$, as a surjective function mapping inputs into a finite number of non-overlapping regions. We represent each region with a unique point in the original domain $\mathfrak X$; i.e., the range of partition strategy $\psi$ is $\mathfrak C \subseteq \mathfrak X$, $\card(\mathfrak C) < \infty$. For each training sub-dataset $D_{f_i} = \{  (x_j^{(i)}, y_j^{(i)}) \}_{j=1}^{M_i}$, $i\in[N]$, we can apply the partition to obtain a transformed dataset $\psi(D_N) \coloneqq \{\psi(D_{f_i})\}_{i=1}^N$ where $\psi(D_{f_i}) \coloneqq \{  (\psi(x_j^{(i)}), y_j^{(i)}) \}_{j=1}^{M_i}$.  Given that $\psi(D_N)$ is a ``matching-input'' dataset, we can use the same method described in \S\ref{sssec:kl-approx} to approximate the loss function in Eq.~\ref{eq:loss}.

However, the performance of such an approach  depends heavily on the partition strategy. In the literature, there exist many ways to obtain the partitions, e.g., Mondrian trees~\citep{lakshminarayanan2016mondrian, wang2018batched}, clustering methods~\citep{omran2007overview}, etc. More recently, \citet{terenin2022numerically} showed theoretical guarantees of partitioning data using covering trees; they also suggested summarizing the datapoints in each partition to a single datapoint with their mean observed values, while adjusting the weights of each summarized datapoint for efficient computation.

We leave the partitioning strategy as a future work. In \S\ref{ssec:nll_generic}, we show an alternative approximation for generic cases of training data.

\subsubsection{Multiple ``Matching-input'' Datasets}
\label{sssec:multiple_matching}
The ``matching-input'' dataset described in \S\ref{sssec:matchingdata} requires observations on every training function. In practice, there could be multiple ``matching-input'' datasets, each of which has observations across a different set of training functions. 

Let $Q$ be the total number of these ``matching-input'' datasets, whose datapoints all exist in the training dataset $D_N$. We denote the $q$-th ``matching-input'' dataset as $\tilde D_{q} = \left\{\{(x_j, y^{(i)}_{j})\}_{j\in \mathfrak M^{(q)}}\right\}_{i\in\mathfrak N^{(q)}} $, where $\mathfrak N^{(q)} \subseteq [N]$ is a set of indices for training functions that all have observations on a collection of inputs $\vx^{(q)} = [x_j]_{j\in \mathfrak M^{(q)}}$, and $\mathfrak M^{(q)}$ is the set of indices for these inputs.

For each ``matching-input'' dataset $\tilde D_{q}$, we can use the same estimator in Eq.~\ref{eq:estimate_mean_cov} to obtain estimated mean vector $\tilde\vmu^{(q)}$ and covariance $\tilde \Sigma^{(q)}$. We then approximate the loss function in Eq.~\ref{eq:loss} as follows.
\begin{align}\label{eq:multi-kl}
\L(\mu,\, k \circ \sigma^2) &\coloneqq D_{\KL}\Bigl(\GP(\mu^*, k^*\circ{\sigma_*^2}), \GP(\mu, k\circ{\sigma^2})\Bigl)\nonumber\\
& \approx \max_{q\in[Q]} D_{\KL}\Bigl(\N(\mu^*(\vx^{(q)}), k^*\circ{\sigma_*^2}(\vx^{(q)})), \N(\mu(\vx^{(q)}), k\circ{\sigma^2}(\vx^{(q)}))\Bigl) \nonumber \\
&\approx \frac{1}{Q}\sum_{q}^Q D_{\KL} \biggl( \N\Bigl(\tilde\vmu^{(q)}, \tilde\Sigma^{(q)}\Bigl),  \N\Bigl(\vmu^{(q)}, \Sigma^{(q)} \Bigl)\biggl),
\end{align}
where $\vmu = \mu(\vx^{(q)})$ and $\Sigma=k\circ{\sigma^2}(\vx^{(q)})$. The last step of approximation with uniform weights is for computational convenience. For our experiments on the hyperparameter tuning benchmark described in \S\ref{ssec:data}, there are multiple ``matching-input'' datasets within our data, but each dataset may only have evaluations on a subset of the training functions. Hence we used Eq.~\ref{eq:multi-kl} as the objective function in all of our experiments for EKL.

\subsection{Negative Log Likelihood (NLL)}
\label{ssec:nll}
Alternative to directly computing EKL in Eq.~\ref{eq:kl}, we can expand the KL divergence in the original loss function (Eq.~\ref{epsilonnetkl}) as follows,
\begin{align}
\L(\mu,\, k \circ \sigma^2) 
& = \sup_{\vx \subseteq \mathfrak X, \card(\vx) < \infty} D_{\KL}\Bigl(\N(\mu^*(\vx), k^*\circ{\sigma_*^2}(\vx)), \N(\mu(\vx), k\circ{\sigma^2}(\vx))\Bigl) \nonumber \\
& = \sup_{\vx \subseteq \mathfrak X, \card(\vx) < \infty} \Ex_{\vy\sim \N(\mu^*(\vx), k^*\circ{\sigma_*^2}(\vx))} \log\frac{p(\vy \mid \mu^*(\vx), k^*\circ\sigma_*^2(\vx))}{p(\vy \mid \mu(\vx), k\circ\sigma^2(\vx))} \nonumber \\
& = \sup_{\vx \subseteq \mathfrak X, \card(\vx) < \infty} -\Ex_{\vy\sim \N(\mu^*(\vx), k^*\circ{\sigma_*^2}(\vx))} \log{p(\vy \mid \mu(\vx), k\circ\sigma^2(\vx))} + C^o\label{eq:pre-nll}
\end{align}
where $C^o$ is a constant that does not depend on the model $\GP(\mu, k\circ \sigma^2)$. Without loss of generality, we omit this constant by setting $C^o=0$ in the following approximations. In this section, we show that Eq.~\ref{eq:pre-nll} can be approximated by the marginal log likelihoods of the training data. Eq.~\ref{eq:pre-nll} is also closely related to cross-entropy losses typically used to pre-train deep learning models.

\vspace{.5em}
{\it Case study.}\; If the training dataset $D_N = $ is a ``matching-input'' dataset described in \S\ref{sssec:matchingdata}, we can approximate Eq.~\ref{eq:pre-nll} by truncating the domain from $\fx$ to $\vx = [x_j]_{j=1}^M$ and applying Monte Carlo estimation for the expectation; i.e., 
\begin{align}\label{eq:pre-nll-matching}
    \L(\mu,\, k \circ \sigma^2) \approx - \frac{1}{N} \sum_{i=1}^N \log p(\vy_i \mid \mu(\vx), k\circ\sigma^2(\vx)) =- \frac{1}{N} \sum_{i=1}^N \log p(D_{f_i} \mid \mu, k\circ\sigma^2) , 
\end{align}
where $\vy_i = [y_j^{(i)}]_{j=1}^M$ is the observed evaluations of training function $f_i$ on inputs $\vx$. Eq.~\ref{eq:pre-nll-matching} recovers the sum of the negative log marginal likelihoods of observations from $\iid$ training functions $f_1, \cdots, f_N$ and it is also an objective considered in multi-task learning~\citep{caruana1997multitask}. A similar result can be shown for generic cases of training datasets that are not ``matching-input''.

\subsubsection{NLL for Generic Cases of Training Data}
\label{ssec:nll_generic}
Now we investigate the general case of the training dataset $D_N = \{D_{f_{i}} \}_{i=1}^N$, where for each function, $D_{f_{i}} = \{(x^{(i)}_{j}, y^{(i)}_{j})\}_{j=1}^{M_i}$ has $M_i$ datapoints and their input locations do not have to be the same as those of other functions. We illustrate the training dataset as follows.
\begin{table}[h]
\centering
\large
\label{tab:nll_data}
\begin{tabular}{c|c|c|c|c}
\toprule
 $f_{1}$ & $\cdots$ & $f_{i}$ & $\cdots$ & $f_{N}$ \\
\midrule
$(x^{(1)}_{1}, y^{(1)}_{1})$ & $\cdots$ & $(x^{(i)}_{1}, y^{(i)}_{1})$ & $\cdots$ & $(x^{(N)}_{1}, y^{(N)}_{1})$ \\ 
 $\vdots$ & $\ddots$ & $\vdots$ & $\ddots$ & $\vdots$ \\ 
$(x^{(1)}_{j},y^{(1)}_{j})$ & $\cdots$ & $(x^{(i)}_{j},y^{(i)}_{j})$ & $\cdots$ & $(x^{(N)}_{j},y^{(N)}_{j})$ \\
 $\vdots$ & $\ddots$ & $\vdots$ & $\ddots$ & $\vdots$\\ 
  $(x^{(1)}_{M_1},y^{(1)}_{M_1})$ & $\cdots$ & $(x^{(i)}_{M_i},y^{(i)}_{M_i})$ & $\cdots$ & $(x^{(N)}_{M_N},y^{(N)}_{M_N})$ \\ 
\end{tabular}
\quad
\begin{tabular}{c}
\begin{small}
$ \vx_i = 
\begin{bmatrix} 
   \;   x^{(i)}_1 \; \\ 
      \vdots  \\
     \;    x^{(i)}_{M_i}\;
   \end{bmatrix}
$
\end{small}
\end{tabular}
\quad
\begin{tabular}{c}
\begin{small}
$ \vy_i = 
\begin{bmatrix} 
   \;   y^{(i)}_1 \; \\ 
      \vdots  \\
     \;    y^{(i)}_{M_i}\;
   \end{bmatrix}
$
\end{small}
\end{tabular}
\end{table}

Training dataset $D_N$ includes observations on a series of finite sets of inputs, $\vx_1, \cdots, \vx_N$. We further approximate the loss function in Eq.~\ref{eq:pre-nll} by restricting the supremum to these sets of inputs:
\begin{align}
 \L(\mu,\, k \circ \sigma^2) %
& \approx \max_{i\in[N]} -\Ex_{\vy\sim \N(\mu^*(\vx_i), k^*\circ{\sigma_*^2}(\vx_i))} \log{p(\vy \mid \mu(\vx_i), k\circ\sigma^2(\vx_i))}  \label{eq:pre-nll-max}\\
& \approx -\sum_{i=1}^N w_i\log\;p(D_{f_i} \mid \mu, k\circ{\sigma^2}), \label{eq:pre-nll-weight}
\end{align}
for some weight assignment $w_i$ subject to $\sum_{i=1}^N w_i = 1$. The approximation step in Eq.~\ref{eq:pre-nll-max} changes the supremum in Eq.~\ref{eq:pre-nll} to a maximum over inputs $\vx_i = [x_j^{(i)}]_{j=1}^{M_i}$ that exist in the training data. Eq.~\ref{eq:pre-nll-weight} uses one sample of $\vy = \vy_i$ to approximate the expectation and rewrites the maximum with a weight assignment. \cite{sagawa2020distributionally} introduced a group distributionally robust optimization method that can be used to optimize the loss in Eq.~\ref{eq:pre-nll-weight}. We set $w_i = \frac{1}{N}$ in this work and found in experiments that the uniform weighting is sufficient for GP pre-training.

\vspace{.5em}
{\it Summary.}\; The NLL approximation of the loss function on any training dataset $D_N = \{D_{f_i}\}$ is
\begin{align}\label{eq:nll}
   \L(\mu, k\circ\sigma^2) \approx - \frac{1}{N} \sum_{i=1}^N \log p(D_{f_i} \mid \mu, k\circ\sigma^2).
\end{align}
For each training function $f_i$, the log marginal likelihood is
\begin{align}
    \log p(D_{f_i} \mid \mu, k, \sigma^2) &=-\frac{1}{2} \left((\vy_i - \vmu_i)\T \Sigma_i^{-1}(\vy_i - \vmu_i) + \log |\Sigma_i| + M_i\log2\pi \right),\label{eq:singleml}
\end{align}
where $\vx_i = [x_j^{(i)}]_{j=1}^{M_i}$, $\vy_i = [y_j^{(i)}]_{j=1}^{M_i}$, $\vmu_i = \mu(\vx_i)$ and $\Sigma_i=k\circ\sigma^2(\vx_i)$.  %




\subsection{Computational Complexity} 
\label{ssec:complexity}
In this section, we analyze the computational complexity of computing and optimizing the approximated loss functions, EKL in Eq.~\ref{eq:kl} and NLL in Eq.~\ref{eq:nll}. We assume the loss functions are optimized over fixed dimensional real-valued parameters $\theta\in\R^{d_\theta}$ ($d_\theta \ll \min(M, N)$). The optimization methods we analyze include gradient descent (GD) and stochastic gradient descent (SGD).%

Recall that $N$ is the number of training functions and let $M=\max_{i=1}^N M_i$ be the maximum number of datapoints observed on training functions. Table \ref{tab:complexity} summarizes our analyses.
 
 \begin{table}[ht]
\begin{center}
\begin{tabular}{l|lll}
    \hline
    
    & & Time & Space \\
    \hline
    \multirow{4}{*}{EKL (Eq.~\ref{eq:kl})}& Overhead & $\mathcal O( M^2 N)$ & $\mathcal O( M^2)$\\
    &Loss function & $\mathcal O( M^3)$ & $\mathcal O( M^2)$\\
    &GD & $\mathcal O( M^3K)$ & $\mathcal O( M^2)$\\
    &SGD & $\mathcal O( B^2MK)$ & $\mathcal O( B^2)$\\
    \hline
    
    \multirow{4}{*}{NLL (Eq.~\ref{eq:nll})}& Loss function & $\mathcal O( M^3 N)$ & $\mathcal O( M^2)$\\
    &Parallel & $\mathcal O( M^3)$ & $\mathcal O( M^2N)$\\
    &GD & $\mathcal O( M^3NK)$ & $\mathcal O( M^2)$\\
    &SGD & $\mathcal O( B^2MNK)$ & $\mathcal O( B^2)$\\
    \hline
\end{tabular}
\end{center}
\caption{Time and space complexity. $K$ is the number of optimization steps (or epochs in stochastic optimization). $B$ is the mini-batch size of SGD over datapoints per training function.}
\label{tab:complexity}
\end{table}

EKL in Eq.~\ref{eq:kl} requires an overhead of estimating mean and covariance, which takes $\mathcal O( M^2 N)$ for matrix multiplication. Once the mean and covariance are estimated, EKL has a time complexity of $\mathcal O(M^3)$ to compute and a space complexity of $\mathcal O(M^2)$. %
 
NLL in Eq.~\ref{eq:nll} naturally decomposes into a sum of data likelihood terms on each sub-dataset $D_{f_i}$. The time complexity to compute Eq.~\ref{eq:nll} is $\mathcal O( M^3 N)$, %
and the space complexity is $\mathcal O( M^2)$. If we have $N$ processes computing each additive component of Eq.~\ref{eq:nll} in parallel, the time complexity can be reduced to $\mathcal O( M^3)$ while the space complexity becomes $\mathcal O( M^2N)$.

The main computational cost for EKL and NLL is solving linear systems (computing $\Sigma^{-1}$ in Eq.~\ref{eq:nondegenerate} and Eq.~\ref{eq:degenerate}; $\Sigma_i^{-1}$ in Eq.~\ref{eq:singleml}). We can optionally use approximation methods for GPs in the literature to reduce the time complexity. For example, using $V$ random features~\citep{rahimi2007random}, the time complexity becomes $\mathcal O(V^3)$ instead of $\mathcal O(M^3)$.

Our method scales (at most) linearly with the number of tasks, $N$, in contrast to the cubic $\mathcal O( M^3 N^3)$ scaling of multi-task or contextual GPs~\citep{bonilla2007multi,swersky2013multi,bardenet2013collaborative, poloczek2016warm, yogatama2014efficient}. The only cubic cost is on the number of datapoints observed on each training function.

If we optimize the loss functions with SGD, the time complexity of computing the NLL objective (Eq.~\ref{eq:nll}) can be reduced to $\mathcal O(B^2 M N)$, where $B$ is the mini-batch size of datapoints per training function. The space complexity reduces from $\mathcal O(M^2)$ in the original case to $\mathcal O(B^2)$ for the loss on mini-batches. However, SGD changes the optimization landscapes of both EKL and NLL. %

In brief, both EKL in Eq.~\ref{eq:kl} and NLL in Eq.~\ref{eq:nll} enjoy low computational costs which scale at most linearly with the number of training functions. As a result, optimizing these approximated loss functions is not very expensive.

\subsection{Relations between EKL and NLL}
\label{ssec:nll_ekl_relation}

EKL and NLL are derived in different ways, but both approximate the same loss function in Eq.~\ref{eq:loss}. In this section, we analyze the situations where EKL and NLL are equivalent (\S\ref{sssec:nll_ekl_equiv}) and explain their important differences (\S\ref{sssec:nll_ekl_diffs}) to provide deeper insights into the two objectives.

\subsubsection{Conditions for the Equivalence of EKL and NLL}
\label{sssec:nll_ekl_equiv}
\underline{\it{Claim.}}\; If the training dataset is a matching-input dataset described in \S\ref{sssec:matchingdata}, where $\vy_i$ are \iid samples from $\N(\vmu^*, \Sigma^*)$, and the estimated distribution $\N(\tilde\vmu, \tilde \Sigma)$ in Eq.~\ref{eq:estimate_mean_cov} is non-degenerate, EKL is equivalent to NLL up to a constant factor:
\begin{align*}
    D_{\KL}\Bigl(\N(\tilde\vmu, \tilde \Sigma), \N(\vmu, \Sigma)\Bigl) \equiv - \frac{1}{N} \sum_{i=1}^N \log p(\vy_i \mid \vmu, \Sigma) + \frac{1}{N} \sum_{i=1}^N \log p(\vy_i \mid \tilde\vmu, \tilde \Sigma).
\end{align*}

We prove this claim in \S\ref{app:equations}.

\subsubsection{Differences between EKL and NLL}
\label{sssec:nll_ekl_diffs}
EKL naturally captures the correlations between function values by taking advantage of the matching inputs, while NLL is more flexible and naturally adapts to observations made at different input locations across training tasks. In different scenarios, one may choose either EKL or NLL. With NLL, each sub-dataset $D_{f_i}$ may have different cardinality and arbitrary input locations. EKL works naturally with observations collected on the same inputs across training functions. While we can potentially use EKL for generic training datasets (\S\ref{sssec:kl-extensions}), it generally involves more heuristics and manipulation of data than NLL.%

\S\ref{sssec:nll_ekl_equiv} points out the interesting equivalence between EKL and NLL in special cases, but there remain important differences between EKL and NLL, including (1) quantitative evaluations, (2) implementation and data structures, (3) interpretation, and (4) extensions. 

\begin{figure}
    \centering
    \includegraphics[width=1.\textwidth]{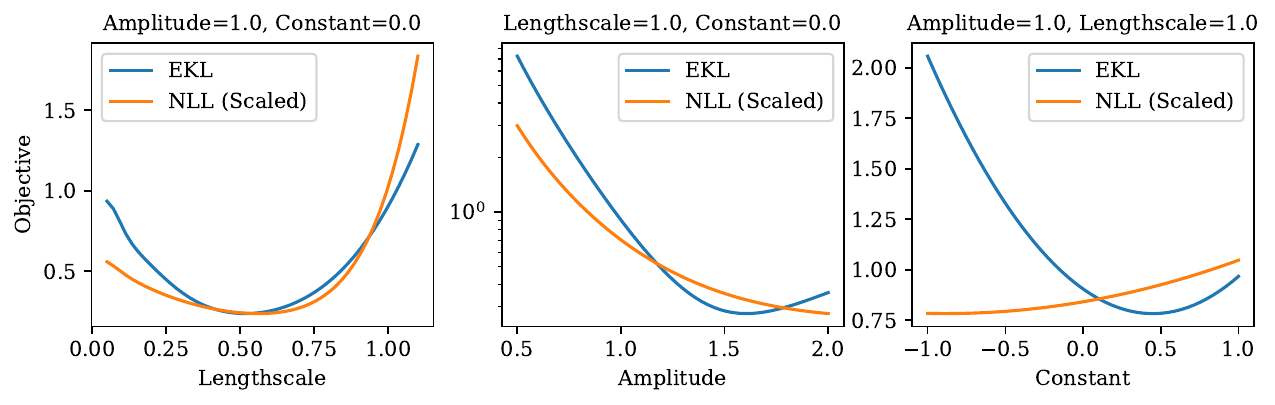}
    \caption{We define the a ground truth GP that has a zero mean and a Matern 5/2 kernel with amplitude $1.0$ and lengthscale $1.0$ on a 1-dimensional domain. We then sample 3 \iid training functions from the GP and their evaluations on 5 inputs. Those 5 inputs are sampled \iid from the standard normal distribution. The model is a GP with a constant mean function parameterized by a constant, and a squared exponential kernel parameterized by a lengthscale and an amplitude value. The figures visualize EKL and NLL (scaled by $\frac{\min EKL}{\min NLL}$ to allow consistency on scale) over each parameter with the other two fixed. In this setting, EKL and NLL have different landscapes and different $\argmin$ locations.}
    \label{fig:my_label}
\end{figure}

\vspace{.5em}
{\it Quantitative evaluations.}\; If the estimated distribution $\N(\tilde\vmu, \tilde \Sigma)$ in Eq.~\ref{eq:estimate_mean_cov} is degenerate, EKL and NLL can have different loss landscapes. 
Figure~\ref{fig:my_label} shows the visualization of the quantitative differences between EKL and NLL for a toy problem, where the loss landscapes and $\argmin$ locations of the two objectives are different.

\vspace{.5em}
{\it Implementation and data structures.}\;
Regardless of whether the estimated distribution is degenerate, the implementation procedures of EKL and NLL are different. As discussed in \S\ref{ssec:complexity}, NLL can be $N$ times more expensive to compute than EKL. Importantly, the implementation of EKL and NLL relies on different types of data structure.

It is more convenient for EKL to partition across inputs to use the [input, evaluations on all functions] structure $\{(x_j, [y_j^{(i)}]_{i=1}^N)\}_{j=1}^M$, and NLL to partition across functions to use the [input, evaluation per function] structure $\{\{(x_j, y_j^{(i)} )\}_{j=1}^M\}_{i=1}^N$. During SGD, EKL first samples inputs and then computes the loss function value on those inputs, while it is more natural for NLL to sample functions and datapoints for each function. 

In practice, there are multiple ``matching-input'' datasets (\S\ref{sssec:multiple_matching}), and NLL would have to process the [input, evaluations on all functions] structure to combine all inputs if NLL were to use EKL's data structure. And for EKL to use NLL's [input, evaluation per function] structure, EKL would have to process the data to construct or find all matching inputs. More computational costs would incur in both of these scenarios. 

\vspace{.5em}
{\it Interpretation.}\;
  Although the NLL loss function is more flexible, it can be difficult to interpret its correspondence to model quality, e.g., how high should the likelihood be for us to stop our search for a decent model? EKL in Eq.~\ref{eq:kl}, on the other hand, is a divergence that is non-negative and equals 0 if and only if the two distributions are identical. One may choose to do early stopping or model selection based on how close EKL is to $0$. 
  
  From the perspective of information theory~\citep{cover2006elements}, we know that EKL in Eq.~\ref{eq:kl} can be interpreted as the average number of extra nats (or bits if we use $2$ as the base of the logarithm instead of $e$) to encode the estimated multivariate Gaussian $\N(\tilde \vmu, \tilde K)$ with a different model, compared to the the average number of nats to describe $\N(\tilde \vmu, \tilde K)$. Comparatively, EKL in Eq.~\ref{eq:kl} is more interpretable than NLL in Eq.~\ref{eq:nll}.

\vspace{.5em}
{\it Extensions.}\;
EKL aims to measure the distance between two distributions: an estimate of the ground truth and the model. 
In this work, we use MLE as the estimate and the KL divergence as the ``distance'' for EKL, and establish the equivalence between solutions of NLL and EKL in special cases of data (Proposition~\ref{prop:gpexist}). Extensions of EKL include using different estimators for the ground truth (e.g., the unbiased sample mean sample covaraince estimator) and adopting other distance measures (e.g., the Wasserstein distance or Rényi divergence). NLL, on the other hand, does not have these straightforward extensions.

%

%

\section{\hyperbo: Bayesian Optimization with Pre-trained Gaussian Process Priors}
\label{sec:bo}
Bayesian optimization (BO) involves making decisions under uncertainty, trading off exploration and exploitation.  In the previous sections, we developed pre-training methods to improve uncertainty estimates by leveraging existing data to specify better priors. The last piece of the puzzle is to connect the pre-trained Gaussian process (GP) to \bo methods.
In this section, we present \hyperbo, a general \bo framework using the pre-trained GP as the prior.

As summarized in Algorithm~\ref{alg:hyperbo}, \hyperbo is a simple wrapper over GP pre-training and classic \bo steps for an unknown function $f$. We propose to fix the pre-trained GP in all \bo steps (lines~\ref{alg:bostart} to \ref{alg:boend}), so that we do not train the model and derive its posterior on the same set of datapoints $D_f$. %
Observations on function $f$ are used only for posterior inference $p(f\mid D_f) = \GP(\hat \mu, \hat k\circ\hat \sigma^2\mid D_f)$, but not any additional re-training steps (e.g., type II maximum likelihood) that modifies the mean and kernel functions of the GP.

 \begin{algorithm}[H]
       \caption{HyperBO for optimizing unknown function $f$.}\label{alg:hyperbo}
  \begin{algorithmic}[1]
    \Function{HyperBO\,}{$f, D_N$}
    \State $\GP(\hat \mu, \hat k\circ\hat \sigma^2) \gets \textsc{Pre-Train}(D_{N})$\label{alg:train}
    \Comment{Pre-train a GP on training dataset $D_N$ (\S\ref{ssec:objective_overview}).}
    \State $D_f \gets \emptyset$
    \For{$t = 1,\cdots, T $} \label{alg:bostart}
        \State $x_t\gets \underset{x\in\mathfrak X}{\argmax}{\,\alpha\left(x ; \GP(\hat \mu, \hat k\circ\hat \sigma^2\mid D_f) \right)}$
        \Comment{Optimize the acquisition function $\alpha(\cdot)$.}\label{alg:strategy}
        \State $y_t\gets$ \textsc{Observe}$\left(f(x_t)\right)$ \Comment{Collect noisy output of function $f$ at input $x_t$.}
        \State $ D_f \gets D_{f}\cup \{(x_t,y_t)\}$
      \EndFor \label{alg:boend}
     \State \Return $D_f$
    \EndFunction
  \end{algorithmic}
\end{algorithm}

\hyperbo is a combination of an empirical Bayes\footnote{See \S\ref{ssec:fully_bayes} for a discussion on the Bayesian interpretations of \hyperbo.} pre-training method and a fully Bayesian sequential decision making procedure. Once we obtain the estimated prior from pre-training, we treat it as the actual prior in the decision making module of \hyperbo. Thus, we can circumvent the practical issues of unknown GP priors in BO. However, to \hyperbo on an upper level, the pre-trained GP is not the ground truth GP prior. As we will see in \S\ref{ssec:finite_hyperbo} and \S\ref{ssec:bo_synthetic}, in some cases with few training functions, the difference between the ground truth and the pre-trained GP can lead to drastically different posterior predictions. This is harmful to BO. It is important to understand when this failure case can happen, so as to diagnose and avoid such scenarios. %

\subsection{Posterior Inference and Acquisition Strategies}
\label{ssec:post_acfun}
In \hyperbo, the acquisition function $\alpha(\cdot)$ (line~\ref{alg:strategy} of Algorithm~\ref{alg:hyperbo}) can be any acquisition function that is fully defined by a Gaussian process (GP) posterior. However, it is important to keep in mind that the pre-trained model \pgp, is an approximation of the ground truth \gtgp. The relations between their corresponding acquisition function values are still unclear. %

To understand the subtleties, we first compare the ground truth posterior and the pre-trained GP posterior. In the $t$-th iteration at line~\ref{alg:strategy} of Algorithm~\ref{alg:hyperbo}, we have observations $D_f = \{(x_\tau, y_\tau)\}_{\tau=1}^{t-1}$. Let $\vx = [x_\tau]_{\tau=1}^{t-1}$ and $\vy = [y_{\tau}]_{\tau=1}^{t-1}$. The ground truth posterior is \pgtgp$\vcentcolon=\GP(\mu^*_{t-1}, k^*_{t-1})$, where
\begin{align}
\mu^*_{t-1}(x) &= \mu^*(x) + k^*(x,\vx)(k^*\circ{\sigma_*^2}(\vx))^{-1}(\vy - \mu^*(\vx)), \;\;\forall x\in\mathfrak X,\label{eq:post_mu_star}\\
k^*_{t-1}(x,x') &= k^*(x,x') - k^*(x, \vx)(k^*\circ{\sigma_*^2}(\vx))^{-1} k^*(\vx, x'), \;\;\forall x, x'\in\mathfrak X. \label{eq:post_k_star}
\end{align}
The pre-trained GP posterior is \ppgp$ \vcentcolon= \GP(\hat \mu_{t-1}, \hat k_{t-1})$, where
\begin{align}
\hat \mu_{t-1}(x) &\vcentcolon= \hat \mu(x) + \hat k(x,\vx)(\hat k\circ{\hat \sigma^2}(\vx))^{-1}(\vy - \hat \mu(\vx)), \;\;\forall x\in\mathfrak X,\label{eq:post_mu_approx}\\
\hat k_{t-1}(x,x')&\vcentcolon=  \hat k(x,x') - \hat k(x, \vx)(\hat k\circ{\hat \sigma^2}(\vx))^{-1} \hat k(\vx, x'), \;\;\forall x, x'\in\mathfrak X. \label{eq:post_k_approx}
\end{align}
When optimizing the loss function (Eq.~\ref{eq:loss}) or any of its approximates, we optimize over $\hat k\circ\hat \sigma^2$ together under the constraints that the kernel $\hat k$ is positive definite and the noise variance $\hat \sigma^2$ is positive. For the same $\hat k \circ \hat\sigma^2$, there can be multiple solutions for $\hat k$ and $\hat \sigma^2$, which further complicates the analyses. Even if we achieve the minimum of the loss function (Eq.~\ref{eq:lossobj}), it is unclear if the following statement on the posterior is true:
\begin{align} \label{eq:open_question_posterior}
    \hat \mu_{t-1}, \hat k_{t-1} \; \overset{?}{=}  \;\underset{\mu, k}{\argmin} \;\,D_{\KL}\Bigl(\GP(\mu^*_{t-1}, k^*_{t-1}), \GP(\mu, k)\Bigl).
\end{align}
That is, if the pre-trained GP prior is close to the ground truth, is the pre-trained GP posterior also close to the ground truth posterior? %
We set this question aside for now and revisit in \S\ref{ssec:analyses}. 

In \hyperbo, since we have no access to the ground truth posterior, it is natural to construct acquisition strategies with the pre-trained GP posterior. Algorithmically, there is no constraint on what acquisition functions $\alpha(\cdot)$ should be used in \hyperbo.  For example, popular acquisition functions like GP-UCB~\citep{kushner1962versatile, srinivas2009gaussian}, EI~\citep{saltenis1971method, mockus1974} or PI~\citep{kushner1962versatile, kushner1964} are all directly applicable. The question remains whether an acquisition function based on the pre-trained GP reflects the strategy of the ground truth acquisition function, i.e.,
\begin{align} \label{eq:open_question_acfun}
    \underset{x\in\mathfrak X}{\argmax} \; \,
    \alpha
    \left(x ; \GP(\hat \mu, \hat k\circ\hat \sigma^2\mid D_f) \right) \; 
    \overset{?}{=}  
    \;\underset{x\in\mathfrak X}{\argmax} \;\,
    \alpha\left(x ; \GP(\mu^*,  k^*\circ\sigma_*^2\mid D_f) \right).
\end{align}

The open questions in Eq.~\ref{eq:open_question_posterior} and Eq.~\ref{eq:open_question_acfun} are the sufficient but not necessary conditions to obtain comparable regrets to the acquisition strategy with the ground truth GP prior. 

We consider the following two acquisition functions for more analyses in \S\ref{ssec:finite_hyperbo}. The acquisition functions are defined over the pre-trained GP posterior \ppgp in Eq.~\ref{eq:post_mu_approx} and Eq.~\ref{eq:post_k_approx}. %
\begin{itemize}
    \item A variant of GP-UCB~\citep{srinivas2009gaussian} with explore-exploit trade-off parameter $\beta$:
\begin{align}
\alpha^{\text{UCB}}_{t-1}(x) = \hat \mu_{t-1}(x) + \beta \sqrt{\hat k_{t-1}\circ{\hat\sigma^2}( x)}. \label{eq:acfun-ucb}
\end{align}
\item A variant of the max-value version of PI~\citep{wang2017maxvalue, kushner1964} with $\hat f^*$ as an estimate of the max-value of the test function $f$:
\begin{align}
\alpha^{\text{PI}}_{t-1}(x) = \frac{\hat \mu_{t-1}(x) - \hat f^* }{\sqrt{\hat k_{t-1}\circ{\hat\sigma^2}( x)}}.  \label{eq:acfun-pi}
\end{align} 
\end{itemize}

As shown by~\citet{wang2017maxvalue}, GP-UCB and PI are closely related to entropy search based methods~\citep{hennig2012, hernandez2014predictive}, and the max-value of test function $f$ can be estimated from the posterior on $f$.

Next, we conduct a case study on finite input domain $\mathfrak X$ and provide theoretical insights to understand asymptotic behaviours of the pre-trained GP posterior predictions in \S\ref{ssec:analyses}. As part of the verification for \hyperbo, we present its regret bounds (\S\ref{ssec:theory}) with unknown ground truth GP priors. We explain the intuitions of these theoretical analyses with synthetic examples in \S\ref{ssec:bo_synthetic}.%

\subsection{Case Study: Functions with Finite Domains}
\label{ssec:finite_hyperbo}
We study the theoretical aspects of \hyperbo where the cardinality of the input domain, $|\mathfrak X|$, is finite. We can then consider the perfect case where we observe all values of the training functions $f_1, \cdots, f_N$.  That is, we further strengthen the assumption in \S\ref{sssec:matchingdata} to be ``observing training functions on all inputs in the finite domain''. 

\begin{assumption}\label{asp:finite}
The domain $\mathfrak X = \{x_j\}_{j=1}^M$ contains a finite number of inputs. The training dataset is $D_N = \{D_{f_i}\}_{i=1}^N$ where $D_{f_i} = \{(x_j, y^{(i)}_j)\}_{j=1}^M$. Assume $N> M$ and the estimated covariance matrix in Eq.~\ref{eq:estimate_mean_cov} is full rank.
\end{assumption}

Now, we can show the pre-trained \pgp has a closed form solution.

\begin{prop}\label{prop:gpexist}
Let $\vx=[x_j]_{j=1}^M\in\R^{M\times d}, Y=[y^{(i)}_j]_{ j\in[M], i\in[N]}\in \R^{M\times N}$. Given Assumption~\ref{asp:iid},~\ref{asp:noise} and~\ref{asp:finite}, we have 
\begin{align}
    \hat\mu(\vx) = \frac{1}{N} Y 1_{N} \;\;\; \text{and}\;\;\;  \hat k\circ{\hat\sigma^2} (\vx) = \frac{1}{N}(Y -\hat\mu(\vx)1_N\T) (Y- \hat\mu(\vx)1_N\T)\T \label{eq:case_study_mean_cov}
\end{align}
such that $\hat\mu, \hat k\circ{\hat\sigma^2} = \underset{\mu, k\circ\sigma^2}{\argmin} \;\L^{\text{EKL}}(\mu, k\circ\sigma^2) = \underset{\mu, k\circ\sigma^2}{\argmin} \;\L^{\text{NLL}}(\mu, k\circ\sigma^2)$, where the loss function $\L^{\text{EKL}}$ is defined in Eq.~\ref{eq:kl} and $\L^{\text{NLL}}$ is defined in Eq.~\ref{eq:nll}.
\end{prop}
Proposition~\ref{prop:gpexist} is not difficult to show. 
The EKL loss function is a KL divergence between two multivariate Gaussian distributions. EKL is non-negative and reaches its minimum value $0$ if the two distributions are the same. Eq.~\ref{eq:case_study_mean_cov} ensures that the pre-trained model $\N(\hat\mu(\vx), \hat k\circ{\hat\sigma^2} (\vx))$ and the estimated model  $\N(\tilde \vmu, \tilde \Sigma)$ from Eq.~\ref{eq:estimate_mean_cov} are the same, so that the pre-trained model minimizes EKL. The solution to the NLL objective becomes the maximum likelihood estimate, which is the same as Eq.~\ref{eq:estimate_mean_cov}. To show that there exists functions $\hat \mu$ and $\hat k\circ \hat\sigma^2$ that satisfy Eq.~\ref{eq:case_study_mean_cov}, we can construct a simple memory based model. The model stores each element of vector $\tilde \vmu$ and matrix $\tilde \Sigma$ from Eq.~\ref{eq:estimate_mean_cov}. When making a prediction at any input or pairs of inputs, the model simply retrieves the corresponding mean or covariance values saved in the model.

Recall that one of the open questions we had in \S\ref{ssec:post_acfun} is the proximity between pre-trained GP posteriors and ground truth GP posteriors. Assumption~\ref{asp:finite} and Proposition~\ref{prop:gpexist} enable us to bound pre-trained GP posterior predictions with ground truth posterior predictions. The following analyses in \S\ref{ssec:finite_hyperbo} all assume Assumptions~\ref{asp:iid},~\ref{asp:noise} and~\ref{asp:finite}.

\subsubsection{Bounding Pre-trained GP Posterior Predictions}
\label{ssec:analyses}
As discussed in \S\ref{ssec:post_acfun}, we are interested in the relation between the pre-trained GP posterior \ppgp and the ground truth posterior \pgtgp  for each iteration of \hyperbo in Algorithm~\ref{alg:hyperbo}. Theorem~\ref{thm:posterior_bound} shows that the pre-trained GP posterior mean and variance are bounded by the ground truth posterior mean and variance.

\begin{thm}\label{thm:posterior_bound}
Assume $N > T+1$ and $T\geq t\geq 1$. At the $t$-th iteration of \hyperbo in Algorithm~\ref{alg:hyperbo}, for any input $x\in \fx$, we have 
\begin{align*}
    \Ex[\hat \mu_{t-1}(x)] = \mu_{t-1}^*(x) \;\;\; \text{and}\;\;\; \Ex[\hat k_{t-1}\circ \hat \sigma^2(x)] = \frac{N-t}{N} k^*_{t-1}\circ\sigma_*^2(x).
\end{align*}
With probability at least $1-\delta$,
\begin{align*}
   |\hat \mu_{t-1}(x) -\mu_{t-1}^*(x)|^2 < a^2 k^*_{t-1}\circ\sigma_*^2(x) \;\text{and}\; 1-2\sqrt{b } < \frac{N \hat k_{t-1}\circ \hat \sigma^2(x) }{(N-t) k^*_{t-1}\circ\sigma_*^2(x)} < 1+ 2\sqrt{b} + 2b,
\end{align*}
where $a^2 = \frac{4\left(t+2\sqrt{(t-1)\log{\frac{4}{\delta}}} + 2\log{\frac{4}{\delta}}  - 2/N \right)}{(N-t-1)\delta}, b=\frac{1}{N-t}\log\frac4{\delta}.$
\end{thm}
The proof can be found in \S\ref{sec:bound}. Theorem~\ref{thm:posterior_bound} conveys an interesting finding: as the iteration $t$ increases, both bounds become looser and the pre-trained GP posterior variance gradually becomes more biased. As a remedy, we can readjust the scale of the pre-trained GP posterior variance by $\frac{N}{N-t}$ to match the ground truth in expectation. 

Perhaps counter-intuitively, Theorem~\ref{thm:posterior_bound} means that we need to have less confidence in the pre-trained GP as we observe more data from a test function. Yet the epistemic uncertainty predictions from the pre-trained GP become smaller with more observations. We can understand this result as a rivalry between the posterior conditioned on a pre-trained GP prior and the uncertainty that comes with the pre-training approximation of the ground truth. Theorem~\ref{thm:posterior_bound} also implies that BO with pre-trained GPs is more reliable when there are more training functions and datapoints per training function than the number of BO iterations.

\subsubsection{Theoretical Analyses on Regret Bounds}
\label{ssec:theory}
The other open question in \S\ref{ssec:post_acfun} is essentially about how acquisition functions with pre-trained GP posteriors impact the performance of BO. 
We show a near-zero regret bound for \hyperbo under Assumption~\ref{asp:iid},~\ref{asp:noise} and~\ref{asp:finite}, i.e., unknown ground truth models and full observations on training functions in finite input domains. The setup of \hyperbo under Assumption~\ref{asp:finite} is equivalent to the finite-arm bandit problem where the values of the arms are distributed according to a multivariate Gaussian. 

\begin{thm} \label{thm:regret}
Let $N>T+4\log\frac{12}{\delta}$, $\delta\in(0,1)$. With probability at least $1-\delta$, the simple regret in $T$ iterations of Algorithm~\ref{alg:hyperbo} with special cases of either GP-UCB in Eq.~\ref{eq:acfun-ucb} or PI in Eq.~\ref{eq:acfun-pi} satisfies
\begin{align}
\label{eq:regret}
& R_T  < O\left(\sqrt{\frac{T}{N-T-1}} + \sqrt{\log\frac1\delta}\right) O(\sqrt{\rho_T /T} + \sigma_*),
\end{align}
where $\rho_T = \underset{A\subset\fx, |A|=T}{\max}\frac12\log|\mI+\sigma_*^{-2}k^*(A)|$.  %
\end{thm}
We describe details of the proof in Appendix~\ref{ssec:app_thm}. In Appendix~\ref{app:ssec:alternative}, we also provide a regret bound defined by the pre-trained GP instead of the ground truth. 
Theorem~\ref{thm:regret} shows that the regret bound always has a linear dependency on the observation noise $\sigma$. This is expected because in practice, we select the best observation rather than best function value (before observing a noisy version of it) to compute the simple regret. Another reason is that we learn the noise parameter $\sigma$ jointly with the kernel, as shown in Eq.~\ref{eq:loss}, and when computing acquisition functions (Eq.~\ref{eq:acfun-ucb} or Eq.~\ref{eq:acfun-pi}), the noise parameter $\sigma$ is always included in the predicted variance.

Intuitively, the more sub-datasets we have in the training dataset, the larger $N$ is, the better we are able to estimate the ground truth GP model, and the closer the regret bound is to the case where the  ground truth GP model is assumed known. Interestingly, the number of BO iterations, $T$, makes the regret smaller in the second term but larger in the first term in Eq.~\ref{eq:regret}. Usually as we get more observations, we get more information about the maximizer, and we are able to optimize the function better. However, as we get more observations on the new function, the pre-trained GP posterior predictions have more freedom to deviate from the ground truth, as shown by Theorem~\ref{thm:posterior_bound}. Hence, we get less and less confident about our predictions, which eventually leads to a looser regret bound. 

It is tempting to prove similar bounds for more general settings where inputs are not the same across training functions or the input domain is continuous. Though the only prerequisite is to show that the difference between the pre-trained mean/kernel and the ground truth mean/kernel is small, this prerequisite is as difficult as showing we can find a model that has bounded generalization error across the entire continuous input space of an arbitrary function. %
We leave the regret bound for general settings as an open question.

\subsection{Understanding \hyperbo with Synthetic Examples}
\label{ssec:bo_synthetic}
We here ground our theory on simple synthetic setups to more intuitively understand \hyperbo. Note that the critical component in BO is posterior inference, since all decision making relies on the posterior. Hence we focus on the posterior aspects of \hyperbo in this section.

Our analyses also rely on the notation of functions. Mathematically, function representations can involve infinite-dimensional vectors or finite parameterization with specific modeling choices. But from a computer science perspective, on a compact domain, we can represent a function with a finite number of datapoints without specifying parameterized models. This is because in a computer, numbers are represented by floats, and floats use finite bits. For example, for optimization trajectories, the points have to move at least a delta (e.g., $6.1\times 10^{-5}$ for float16) at a time. So it is sufficient for us to consider a function fully represented by finite evaluations. 

In the following, \S\ref{sssec:synthetic_few_training functions} considers small function domains with varying numbers of training functions, and provides intuitions on Theorem~\ref{thm:posterior_bound}.  \S\ref{sssec:synthetic_more_datapoints} partly invalidates Theorem~\ref{thm:posterior_bound} in settings with stationary kernels, few training functions, but more training datapoints per function.

\subsubsection{Simple Function Domains}
\label{sssec:synthetic_few_training functions}
\begin{figure}
    \centering
    \includegraphics[width=1.\textwidth]{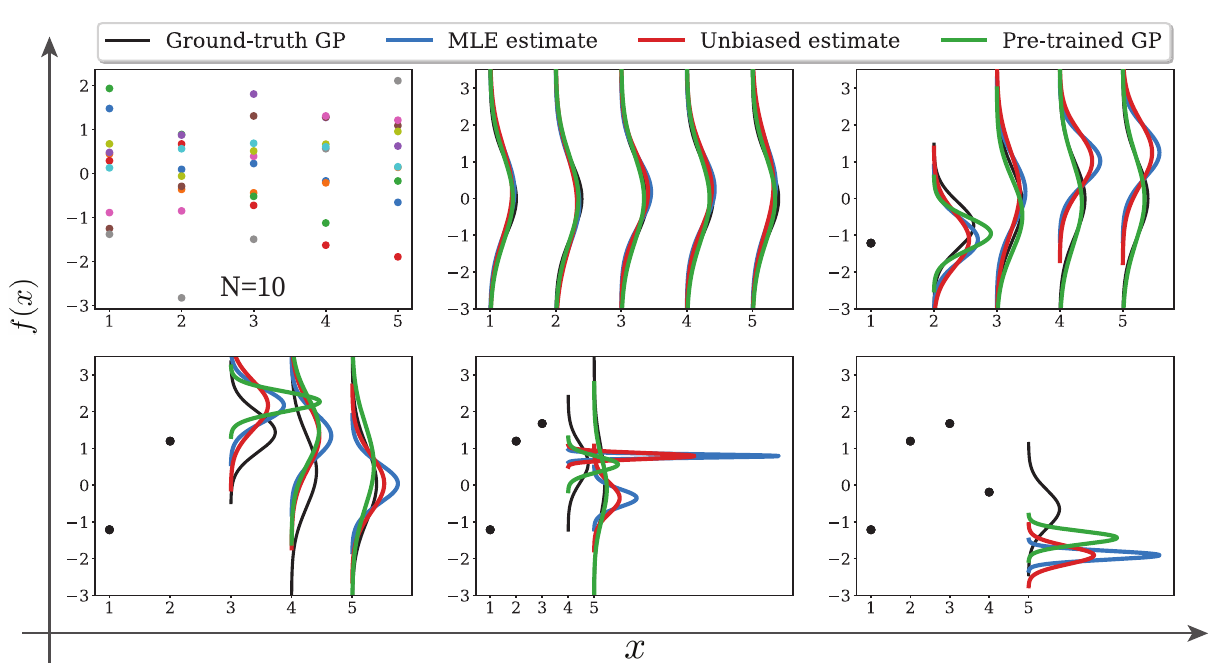}
    \caption{Top left shows 10 training functions, each with one color, sampled from a ground truth GP (a multivariate Gaussian) on a finite domain $\mathfrak X = \{1,2,3,4,5\}$. The top middle plot shows probability densities of the marginal Gaussian distribution for each function value evaluated with the ground truth, MLE estimate, unbiased estimate and pre-trained GP. The following plots show their conditional distributions (i.e., posteriors). With increased observations (black dots), estimate-based posterior predictions become less accurate despite close estimations of the prior.} %
    \label{fig:posterior}
\end{figure}

Here, we assume the ground truth model is a GP with zero noise variance, zero mean and squared exponential kernel on a 1-dimensional domain. The lengthscale parameter and signal variance of the GP are both $1$. That is,
\begin{align*}
    \mu^*(x) = 0 \;\;\; \text{and}\;\;\; k^*\circ\sigma_*^2(x, x') = e^{-(x-x')^2 / 2}, \forall x, x' \in \mathfrak X.
\end{align*}
We consider the simple setup from \S\ref{ssec:finite_hyperbo}. We set the domain to be $\mathfrak X = \{1,2,3,4,5\}$ and collect a training dataset $D_n = \{D_{f_i}\}_{i=1}^N$ by sampling from the GP, which is equivalent to a multivariate normal distribution. The top left of Figure~\ref{fig:posterior} shows the samples, where each sub-dataset $D_{f_i}$ is illustrated with the same color.

\begin{figure}
    \centering
    \includegraphics[width=1.\textwidth]{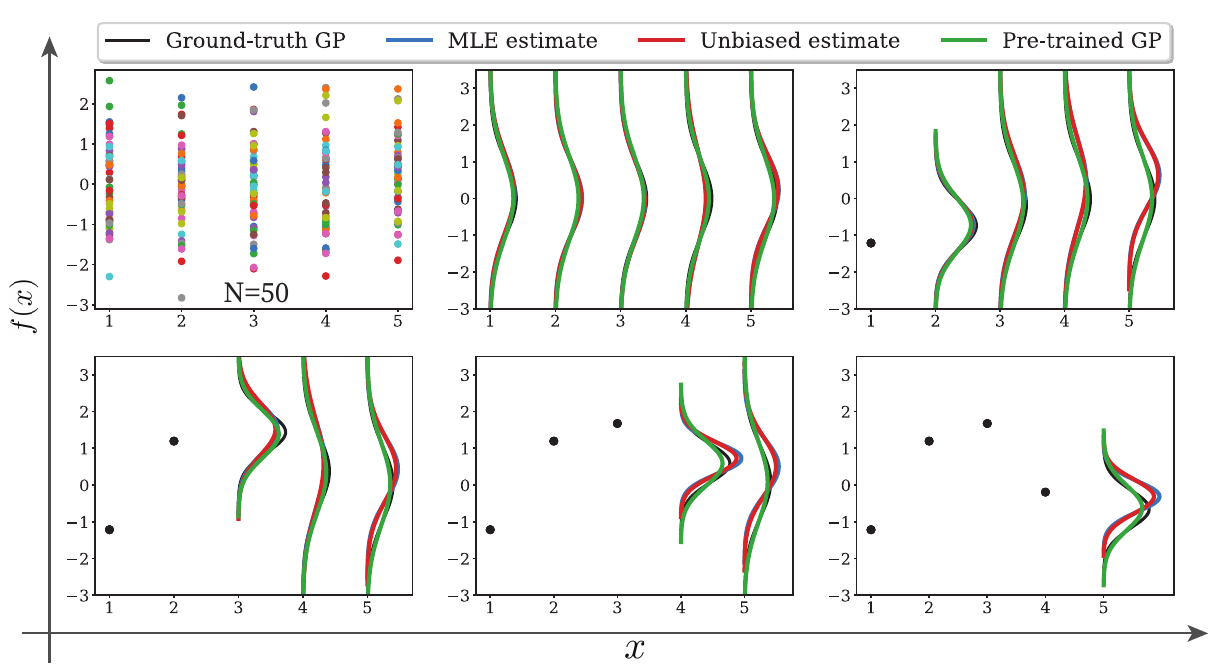}
    \caption{We used the same setup as Figure~\ref{fig:posterior}, except that the size of the training dataset is $N=50$. With more training functions, we obtain more accurate pre-trained GP posterior predictions.}
    \label{fig:posterior_50}
\end{figure}

In Figure~\ref{fig:posterior}, we illustrate the prior and posterior predictions from the ground truth and three modeling choices: the MLE estimate in blue, the unbiased estimate in red and the pre-trained GP in green. The MLE estimate uses Eq.~\ref{eq:estimate_mean_cov}. The unbiased estimator applies a rescaling factor, $\frac{N}{N-t}$, to MLE, and obtain unbiased estimates, where $t$ is the number of observations on test function $f$. If $t=1$, the unbiased estimate is the sample mean and covariance. The pre-trained GP optimizes the EKL loss function (Eq.~\ref{eq:nondegenerate}) over unknown parameters of a squared exponential kernel.

With no observations, the prior estimates (top middle plot) all look aligned with the ground truth. However, as we increase the number of observations from 1 (top right) to 4 (bottom right), posterior predictions based on the estimates deviate more and more from the ground truth posteriors. This confirms the results from Theorem~\ref{thm:posterior_bound}, which says the bounds on estimates rely on $N-t$, so the more observations, the less accurate the estimated posterior predictions become with respect to the ground truth. Note that by applying rescaling in the unbiased posterior estimate, we can avoid the overly confident posterior predictions from MLE.

In Figure~\ref{fig:posterior_50}, we increased the size of the training dataset from 10 in Figure~\ref{fig:posterior} to 50. More training functions allow us to obtain more accurate posterior predictions based on all 3 kinds of modeling choices. Rescaling also becomes less important since $t\ll N$. Interestingly, for both Figure~\ref{fig:posterior} and Figure~\ref{fig:posterior_50}, the pre-trained GP produces more accurate posterior predictions compared to either MLE or unbiased estimates. %

\subsubsection{Function Augmentation by Adding More Training Datapoints}
\label{sssec:synthetic_more_datapoints}
\begin{figure}[ht]
    \centering
    \includegraphics[width=1.\textwidth]{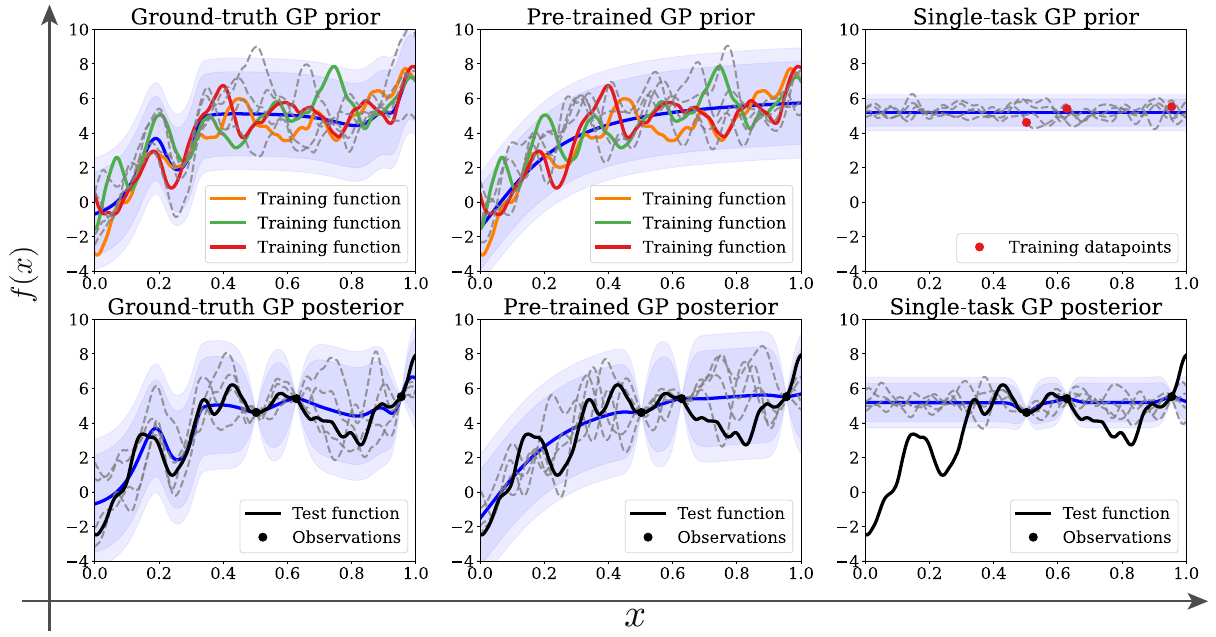}
    \caption{Comparing priors and posteriors based on the ground truth GP, pre-trained GP trained on 3 functions (colored lines), and single-task GP optimized with type-II maximum likelihood on 3 observations (black dots). GP mean functions are illustrated in blue and the shaded areas are 95\% and 99\% confidence intervals. The grey dotted lines are random samples from the GP in each figure. The confidence intervals from the single-task GP posterior failed to cover the test function (the black line), though providing a good fit for the observations. The pre-trained GP did not share the same kernel type as the ground truth. Despite having only 3 training functions, the overall trend of the pre-trained GP posterior matched the ground truth posterior and would be useful for BO.}
    \label{fig:hyperbo_posterior}
\end{figure}

Note that the above analyses for pre-training with few training functions may not hold in some practical settings. If the unknown ground truth kernel is stationary and there are enough observations per training function, as shown by \cite{bachoc2021asymptotic}, we might still be able to obtain a good pre-trained GP. For example, in Figure~\ref{fig:intro_summary}, there are only 3 training functions, but each training function has 200 datapoints covering the input domain. Figure~\ref{fig:hyperbo_posterior} shows prior and posterior predictions made by the ground truth, pre-trained and single-task GPs. 

The ground truth GP used a constant noise variance, a \matern52 kernel and a linear MLP mean function with 3 hidden layers. We used a constant noise variance, a squared exponential kernel and the same setup of a linear MLP mean function for the pre-trained GP.  The single-task GP used constant mean and squared exponential kernel. The parameters of mean and kernel functions are usually referred to as GP hyperparameters~\citep{rasmussen2006gaussian}. We performed type-II maximum likelihood to obtain GP hyperparameters of the single-task GP on 3 observations from the test function. Table~\ref{tab:hyperbo_posterior} shows 4 metrics: (1) the losses defined by NLL in Eq.~\ref{eq:nll}, (2) EKL in Eq.~\ref{eq:degenerate}, both over the training functions, (3) NLL$(f)$: the NLL on the total 200 datapoint of the test function $f$, and (4) NLL$(D_f)$: the NLL on the 3 observations on test function $f$. Single-task GP obtains the lowest NLL$(D_f)$ but all other 3 metrics are higher than other models, implying overfitting. %

\begin{table}[htbp]
\centering
\begin{tabular}{l l l l l}\toprule
 & NLL & EKL & NLL$(f)$ & NLL$(D_f)$ \\\hline
 ground truth GP & -437.1 & 3.2 & -434.5 & 3.3 \\
 Pre-trained GP & -244.4 & 3.6 & -271.5 & 2.8 \\
 Single-task GP & 181.3 & 114.1 & 246.8 & 1.6 \\
\bottomrule
\end{tabular}
\caption{The NLL (Eq.~\ref{eq:nll}) and EKL (Eq.~\ref{eq:degenerate}) loss function values on training functions, together with NLL($f$): the NLL on 200 function evaluations on test function $f$, and NLL($D_f$): the NLL on the 3 observations on test function $f$. While the single-task GP obtained the lowest NLL on the 3 observations, it is not a good model for the underlying test function as reflected by the high NLL($f$) values. On the contrary, the pre-trained GP provided a better fit for test function $f$.}
\label{tab:hyperbo_posterior}
\end{table}

Why does a pre-trained GP still work on such a small set of training functions? We can understand this through the SGD setup, where in every training step, a batch of datapoints are sampled from each training sub-dataset. Given that we have a stationary kernel, each batch can be viewed as points on a different training function. 
Thus, we can effectively augment the set of training functions, so that we won't be constrained by Theorem~\ref{thm:posterior_bound}. %
Concurrently, \cite{fan2023transfer} showed analyses with more precise theoretical and empirical analyses.

\section{Experiments}
\label{sec:exp}

In this section, we evaluate the performance of \hyperbo on a variety of real-world hyperparameter tuning tasks. In total, we performed experiments on 17 search spaces. Each search space has multiple tasks, and the tasks correspond to black-box functions (e.g., for evaluating validation error rates) over the same set of hyperparameters of a system. For every search space, the tasks are divided into training tasks and test tasks. The datapoints from the training tasks compose a training dataset, which we can use to pre-train a Gaussian process (GP). The performance of Bayesian optimization (BO) is evaluated on the test tasks.

Among those search spaces, the first one corresponds to the problem of hyperparameters tuning for the optimizer of modern deep learning models, and the goal is to obtain a tuning strategy that can generalize over different combinations of model architectures \citep[e.g., ResNet50 from][]{he2016deep}, datasets \citep[e.g., ImageNet from][]{imagenet} and hardware settings (which determine batch sizes). We created a new benchmark, PD1 (\S\ref{ssec:data}), for this challenging tuning problem of deep learning. PD1 is the focus of our experiments given its relevance to present-day large scale deep learning applications. The goals of our experiments on PD1 are to investigate the performance of \hyperbo and understand its properties. In \S\ref{ssec:offline_pd1}, we present the simulated offline BO and GP regression results on PD1 with detailed studies on the effects of the number of training datapoints and the number of training tasks. And in \S\ref{ssec:online}, we show the online tuning results for tasks in PD1 to demonstrate the performance of \hyperbo when it is deployed online as part of a real world hyperparameter tuning service.

The rest of the search spaces belong to HPO-B~\citep{pineda2021hpob}, a hyperparameter tuning benchmark for relatively small scale but more classic machine learning models such as decision trees and SVM. To further showcase the performance and robustness of \hyperbo, we present results for both the original HPO-B and representative variants of HPO-B with reduced training data and the ``negative transfer'' effects~\citep{rothfuss2021pacoh} in \S\ref{ssec:hpob}.

Our JAX-based~\citep{jax2018github} implementation of \hyperbo can be found at \url{https://github.com/google-research/hyperbo}, which was used for all of our experiments. To accommodate needs for more modular use cases, we also provide a Flax~\citep{flax2020github} and TensorFlow-Probability~\citep{dillon2017tensorflow} based implementation for GP pre-training at \url{https://github.com/google-research/gpax}.

In the following, we introduce PD1 in \S\ref{ssec:data}, describe compared methods in \S\ref{ssec:exp-methods} and evaluation metrics in \S\ref{ssec:exp-report}, and present the aforementioned results with analyses on PD1 and HPO-B. We give a summary of the experiments in \S\ref{ssec:exp_summary}.

\subsection{PD1: A New Hyperparameter Tuning Benchmark for Optimizing Deep Learning Models}
\label{ssec:data}

To collect our hyperparameter tuning dataset, the PD1 Neural Net Tuning Dataset,
 we defined a set of 24 neural network tuning tasks and a single, broad search space for Nesterov momentum~\citep{nesterov1983method,sutskever2013importance}. Each task is defined by a task dataset \citep[e.g., ImageNet from][]{imagenet}, a specific neural network model \citep[e.g., ResNet50 from][]{he2016deep}, and a batch size (which is determined by the hardware).
 
 To reduce ambiguity, we distinguish between datasets that individual neural networks are trained on and the dataset we collected that includes optimizer hyperparameter points with their validation errors (and other metrics). We will call the former, e.g., MNIST~\citep{lecun2010mnist} and CIFAR10~\citep{Krizhevsky09learningmultiple}, task datasets and call the latter the tuning dataset. The tuning dataset is what we described as dataset $D_N$ in \S\ref{sec:pf}.
 
 Table~\ref{tab:workloads} shows all the tasks that we consider in the tuning dataset. We used an existing code base~\citep{init2winit2021github} for neural network model training. The dataset used roughly 12,000 machine-days of computation on TPUv4i~\citep{jouppi2021ten} for approximately 50,000 hyperparameter evaluations. Depending on the task, the runtime of each hyperparameter evaluation may vary from minutes to days.

For each task, we trained the model on the task dataset repeatedly using Nesterov momentum \citep{nesterov1983method,sutskever2013importance}, with the task's minibatch size, with different hyperparameter settings drawn from the 4-dimensional search space detailed in Table~\ref{tab:pd1_search_space}. We tuned the base learning rate, $\eta$, on a log scale, the momentum, $\beta$, with $1-\beta$ on a log scale, and the polynomial learning rate decay schedule power $p$ and decay steps fraction $\lambda$. We used a polynomial decay schedule with the following form:
\begin{align}
\label{eq:lr-schedule}
\eta_\tau &= \frac{\eta}{1000} + \left(\eta - \frac{\eta}{1000}\right) \left(1 - \frac{\min(\tau, \lambda \mathcal{T} )}{\lambda \mathcal{T}} \right)^{p} ,
\end{align}
where $\tau$ is the training step and $\mathcal{T}$ is the total number of training steps for the task. 

We collected two types of data: matched and unmatched data. Matched data used the same set of uniformly-sampled hyperparameter points (i.e., ``matching-input'' in \S\ref{sssec:matchingdata}) across all tasks and unmatched data sampled new points for each task. All other training pipeline hyperparameters were fixed to hand-selected, task-specific default values. All of our tasks are classification problems, so they all used the same training loss, although occasionally task-specific regularization terms were added. For each trial (training run for a single hyperparameter point), we recorded validation error (both cross entropy error and misclassification rate). In many cases, poor optimizer hyperparameter choices can cause training to diverge. We detected divergent training when the training cost became NaN and then marked the trial but did not discard it. Please download the dataset (\url{http://storage.googleapis.com/gresearch/pint/pd1.tar.gz}) and see its descriptions for additional details about the tasks and training procedure. The different tuning tasks vary in difficulty and numbers of datapoints, but generally there are roughly 500 matched datapoints and 1500 unmatched datapoints per tuning task. Among the 500 matched datapoints, 242 datapoints correspond to training runs that did not diverge. For unmatched data only, we attempted to generate roughly similar numbers of non-divergent points across tasks, so tasks with a higher probability of sampling a hyperparameter point that causes training to diverge will tend to have more trials. 

The ImageNet ResNet50 1024 task only has 100 hyperparameter points because we abandoned it when scaling up data collection in order to save compute resources. It is used in training, but not evaluation. In total, we have 23 test tasks in PD1. For each test task, we used subsets of the other 23 tasks (including ImageNet ResNet50 1024) to compose training datasets.

\begin{table}[!htb]
\scriptsize
\begin{tabular}{cc}
\tiny
\begin{minipage}{.5\linewidth}
\centering

\begin{tabular}{llr}
\hline
Task Dataset       & Model         & Batch Sizes \\ \hline
CIFAR10       & Wide ResNet           & \{256, 2048\}  \\
CIFAR100      & Wide ResNet           & \{256, 2048\}    \\
Fashion MNIST & Max pool CNN ReLU & \{256, 2048\}    \\
Fashion MNIST & Max pool CNN tanh & \{256, 2048\}    \\
Fashion MNIST & Simple CNN    & \{256, 2048\}    \\
ImageNet      & ResNet50      & \{512, 1024, 2048\}    \\
LM1B          & Transformer   & \{2048\}        \\
MNIST         & Max pool CNN relu & \{256, 2048\}    \\
MNIST         & Max pool CNN tanh & \{256, 2048\}    \\
MNIST         & Simple CNN    & \{256, 2048\}    \\
SVHN (no extra) & Wide ResNet           & \{256, 1024\}    \\
WMT15 German-English & xformer       & \{64\}          \\
UniRef50      & Transformer   & \{128\}        
\end{tabular}

\caption{Tasks in PD1.}
\label{tab:workloads}
\end{minipage} &

\begin{minipage}{.4\linewidth}
\centering

\begin{tabular}{|c|c|c|}
\hline
Hyperparameter & Range & Warping \\ \hline
$\eta$ & $\srange{10^{-5}}{10}$ & Log \\ \hline
$p$ & $\srange{0.1}{2.0}$ & Linear \\ \hline
$1 - \beta$ & $\srange{10^{-3}}{1.0}$ & Log \\ \hline
$\lambda$ & $\srange{0.01}{0.99}$ & Linear \\ \hline
\end{tabular}

\caption{4-dimensional input search space of PD1 (see Eq.\ref{eq:lr-schedule}).}
\label{tab:pd1_search_space}
\end{minipage}
\end{tabular}

\end{table}

For our experiments on PD1, we also used output warping in addition to input warping described in Table~\ref{tab:pd1_search_space}. The validation error rate outputs are warped as $r\gets -\log (r+10^{-10})$ for all methods unless otherwise mentioned, so that we have a maximization problem for tuning.

\subsection{Description of All Compared Methods}
\label{ssec:exp-methods}
We compared two sets of methods, one was a collection of meta BO methods, including \hyperbo, and the other was for single task BO methods that did not make use of multi-task training data.

Our method \hyperbo has several variants including using different acquisition functions and different objectives. In our experiments, unless otherwise mentioned, we used a thresholded probability of improvement (PI) as the acquisition function. We set PI\footnote{The reason of using $\max_{\tau\in[t-1]} y_\tau + 0.1$ was to approximate the max-value of the function as suggested by~\cite{wang2017maxvalue}, and we found it to be effective across compared \hyperbo variants. Because the observations are (log) error rates, this acquisition function trades off exploration and exploitation - i.e., with larger error rates this seeks relatively more substantial improvements than with small error rates. We also tested other acquisition functions and the results can be found in \S\ref{app:exp-acfun}.} in line~\ref{alg:strategy} of Algorithm~\ref{alg:hyperbo} as 
$$\alpha\left(x ; \GP(\hat \mu, \hat k\circ \hat\sigma^2 \mid D_f = \{(x_\tau, y_\tau)_{\tau=1}^{t-1}\} ) \right) = \frac{\hat\mu_{t-1}(x) - (\max_{\tau\in[t-1]} y_\tau + 0.1)}{\sqrt{\hat k_{t-1} \circ \hat\sigma^2(x)} }.$$

To optimize the loss functions (\S\ref{ssec:objective_overview}), while there exist methods to search for functional structures~\citep{kemp2008discovery, malkomes2018automating}, we opted for fixed but complex and expressive structures of mean and kernel functions. 
We then optimize the objective directly via gradient based methods. Alternatively, one can perform cross-validation on a held-out validation dataset~\citep{wistuba2021few}, which we skipped for simplicity and speed.

We included the following \hyperbo variants in our experiments.
\begin{itemize}
    \item H-NLL: HyperBO with a GP pre-trained via the NLL objective (Eq.~\ref{eq:nll}). We used a 2-hidden-layer neural network of size $(32,32)$ as mean function and an anisotropic Mat\'ern52 covariance on the last feature layer of the mean function as kernel. 
    
    We used tanh activation for the neural network. The neural network gives us a mapping from inputs to embeddings. Let the mapping be $\phi:\mathfrak X\mapsto \R^
    {d_{\text{emb}}}$, a Mat\'ern covariance function be $k_{\text{Mat\'ern}}: \R^{d_{\text{emb}}} \times \R^{d_{\text{emb}}} \mapsto \R$, and a weight matrix be $W\in\R^{1\times d_{\text{emb}}}$. The mean function used by H-NLL is $\mu(x) = W\phi(x)$ for any $x\in\mathfrak X$. The kernel function used by H-NLL is $k(x, x') = k_{\text{Mat\'ern}}\left(\phi(x), \phi(x')\right)$ for any $x, x' \in \mathfrak X$. 
    
    As part of the GP model, we also used softplus warping functions for kernel and likelihood parameters.  
    The NLL objective was optimized with the Adam optimizer~\citep{kingma2014adam} implemented in Optax~\citep{deepmind2020jax} with $10^{-3}$ learning rate, 50,000 training steps and 50 batch size as recommended by~\cite{wistuba2021few}. In each iteration of Adam, we sample datapoints uniformly at random for each task, and compute the log likelihoods. Then, we sum up the log likelihoods across all tasks and use the sum as the loss function value.\footnote{In practice, we found Adam performed better than L-BFGS on NLLs of GPs with Mat\'ern52 kernel. This is partly because Mat\'ern52 kernel with a lot of datapoints can be (numerically) low-rank given that the covariance matrix is represented with finite bits in a computer. We did not investigate further in this work and decided to use Adam for NLL. This problem did not seem to occur for EKL because of much fewer matching datapoints.} 
    \item H-EKL: HyperBO with a GP pre-trained via the EKL objective, i.e., Eq.~\ref{eq:multi-kl} which sums over Eq.~\ref{eq:nondegenerate} (without the $\ln|\tilde \Sigma|$ term) on ``matching-input'' datasets. If the estimated covariance matrix is low rank, we report  Eq.~\ref{eq:degenerate} as the EKL values for the ``matching-input'' datasets. 
    We used the same GP model as H-NLL. We optimized EKL using L-BFGS~\citep{liu1989limited} for 100 iterations. %
    \item FSBO: ``Few-Shot Bayesian optimization'' proposed by~\cite{wistuba2021few}. FSBO is a special case of \hyperbo with zero mean and the NLL objective. We used the same kernel function, parameter warping and optimization method as H-NLL.
    \item ABLR: BO with ``Adaptive Bayesian Linear Regression'' proposed by~\cite{perrone2018scalable}. ABLR is a special case of \hyperbo with zero mean and linear kernel $k(x, x') = b^2 + \phi(x)^T\phi(x') / s^2$ where $s$ and $b$ are kernel parameters, and $\phi$ is the feature layer of a 2-hidden-layer neural network of size $(32,32)$. We used the same optimization method as H-NLL.
\end{itemize}
These settings of HyperBO use different mean and kernel structures in pre-training. We provide more comparisons over acquisition functions and other variants of the GP models in Appendix~\ref{app:exp}.

For other meta BO baselines, we included two scalable methods, MIMO and RFGP, that replace the GP with a regression model that can be trained using stochastic gradient descent and thus scales linearly in the number of observations. Following the multi-task setup of~\citet{springenberg2016bayesian}, we jointly trained a 5-dimensional embedding of each task, which was then added to the input of MIMO and RFGP. We also compared to MAF proposed by~\citet{volpp2020meta}.
\begin{itemize}
    \item MIMO: Multi-task BO with an ensemble of feedforward neural networks with shared subnetworks~\citep{havasi2020training, kim2021scalable} as the surrogate model. We used 1 shared dense layer of size 10 and 2 unshared layers of size 10. We used tanh activation based on Figure 2 from \citet{snoek2015scalable}. The network has one output unit with linear activation and another with $\mathrm{softplus}(10^{-4}, 1)$ activation, corresponding respectively to the mean and standard deviation parameters of a normal distribution. In each BO iteration, we trained for 1000 epochs using the Adam optimizer with learning rate $10^{-4}$ and batch size 64.
    \item RFGP:  Multi-task BO using a GP approximated by random features~\citep{snoek2015scalable, krause2011contextual}. We used the open-source implementation of random Fourier features by \citet{liu2020simple}. In each BO iteration, we trained for 1000 epochs using the Adam optimizer with learning rate $10^{-3}$ and batch size 64.
    \item MAF: The ``Meta Acquisition Function'' method from~\citet{volpp2020meta}. MAF used reinforcement learning to learn an acquisition function modeled by a neural network over a set of transfer learning tasks. All MAF results were generated using the code from \citet{volpp2020meta}. See App.~\ref{app:maf} for experimental details. As MAF takes significantly longer to run than HyperBO and other methods, we only include its results for \S\ref{sssec:hold-out-related} and Figure~\ref{fig:pd1_23}. %
\end{itemize}

Unlike these meta BO baselines, model re-training (a.k.a. fine-tuning) is optional for \hyperbo during BO iterations since pre-training is sufficient to obtain the GP prior. Unless otherwise mentioned, we did not use any model re-training for \hyperbo variants, and posterior inference (for computing acquisition functions) was done without changing the GP in the BO module of \hyperbo.

The remaining baselines are those that \emph{do not} use information from training tasks:
\begin{itemize}
    \item Rand: A random search method that samples uniformly randomly in the search space in each BO iteration. For hyperparameters with input warping, we sample from the corresponding warped input space. For example, as shown in Table~\ref{tab:pd1_search_space}, the learning rate $\eta$ has a rectangular range $[10^{-5}, 10]$ and a logarithm warping function. So, instead of uniformly randomly sampling in $[10^{-5}, 10]$, Rand samples $\log_{10} \eta$ uniformly randomly in $[-5, 1]$.  %
    \item STBO: Single task BO with the same acquisition function and GP structure as H-NLL, except that we put a Lognormal$(0, 0.1)$ prior on the warped lengthscale and signal variance parameters of the anisotropic Mat\'ern52 kernel, and a Normal$(0,0.1)$ prior on the warped noise variance. In each BO iteration, STBO optimizes the GP hyperparameters by running L-BFGS for 100 iterations on the marginal likelihood of observations from the test task. %
    \item STBOH: Single task GP-UCB (coefficient=1.8) with constant mean, anisotropic Mat\'ern52 kernel and \emph{hand-tuned} prior on GP hyperparameters and the UCB coefficient~\citep{srinivas2009gaussian,Golovin2017}. Specifically, log signal variance follows Normal(-1, 1), log lengthscale (one per input parameter) follows Normal$(0,1)$, and log observation noise variance follows Normal(-6, 3). The GP hyperparameters are post-processed by tensorflow-probability's \texttt{SoftClip} bijector to constrain the values between 1st and 99th quantiles. These prior distributions were manually tuned to obtain reasonable convergence rates on 24 analytic functions in COCO \citep{hansen2021coco}. The GP hyperparameters are optimized via maximum marginal likelihood in each BO iteration. 
\end{itemize}

\subsection{Reporting Aggregated Multi-task Results}
\label{ssec:exp-report}
For each test task, our goal is to obtain a low simple regret with as few BO iterations as possible. In many situations, we cannot compute the simple regret defined in \S\ref{sec:pf} due to the lack of information on the test function. However, in the case of running BO on the offline data of PD1 and HPO-B, the test function has a finite set of datapoints and there exists $y_{\max}$, the maximum function value in this finite set. Thus, we can empirically report the regret at the $t$-th iteration of BO as $r_t = y_{\max} - \max_{\tau\in[t]}y_\tau$, where $[y_\tau]_{\tau=1}^t$ is the sequence of observed function values accumulated in BO. %

In the experiments, we ran BO on a set of test tasks with multiple random seeds, and we obtained a collection of regrets over BO iterations: $\{\{[r^{(p,q)}_1, \cdots, r^{(p,q)}_T]\}_{p=1}^P\}_{q=1}^Q$, where $P$ is the number of test tasks and $Q$ is the number of random seeds. 
To report the performance of BO, we used the following three ways to aggregate the regrets and demonstrate how well each method performs. 

\vspace{.5em}
{\it Regret curves} show how regrets change as the number of BO iterations increases. At the $t$-th iteration, we report the median and 20/80 percentile of
    $\{\frac1P\sum_{p=1}^P r^{(p,q)}_t\}_{q=1}^Q$, where each element $\frac1P\sum_{p=1}^P r^{(p,q)}_t$ is the mean of the regrets over test tasks.

\vspace{.5em}
{\it Performance profiles}~\citep{dolan2002benchmarking} have been widely used to evaluate the performance of optimization algorithms. We define the performance profile for a BO method as the fraction of tasks that the method solves at each iteration. The notion of ``solving a task'' depends on specific criteria. We set the criteria to be achieving a regret lower than a constant $C$ at the $t$-th iteration of BO. More formally, the performance profile across BO iterations is $[\frac{1}{PQ}\sum_{q=1}^Q\sum_{p=1}^P \mathbbm{1}_{r^{(p,q)}_1 < C}, \cdots, \frac{1}{PQ}\sum_{q=1}^Q\sum_{p=1}^P \mathbbm{1}_{r^{(p,q)}_T < C}]$.

\vspace{.5em}
{\it Ranking plots} show how the rank of a method (comparing to other competing methods) changes as the number of BO iterations increases. To obtain the ranks, we take the mean of regrets over tasks (in the same way as regret curves) and rank the compared methods from low to high. If there is a tie, we assign the average of the ranks to all the tied methods (e.g., if two methods obtain the same best regret value, their ranks are both 1.5). We report the mean and 1 standard deviation of the ranks over random seeds in ranking plots.

\subsection{Results on the PD1 Offline Hyperparameter Tuning Tasks}
\label{ssec:offline_pd1}
Many tasks in \S\ref{ssec:data} can use a lot of compute resources, which makes it infeasible to perform a wide variety of experiments to analyze the characteristics of BO methods. Hence we adopted an offline approximation, which sets the search space of a test task to be the finite set of points that the corresponding tuning sub-dataset contains. We also filtered the diverged training runs in PD1 to ensure the regrets can be computed. To simulate the online setting, the same datapoint can be observed multiple times, and we used zero initial data for all test tasks. Each method was repeated 5 times with different random seeds to initialize its model.  %

In the following, \S\ref{sssec:hold-out-related} demonstrates the overall performance of \hyperbo; \S\ref{sssec:pd1_nll} presents the GP regression results and their impact on BO; \S\ref{sssec:exp-pd1-datapoints} and \S\ref{ssec:num_training_tasks} show the effects of the number of training datapoints and the number of training tasks.

\subsubsection{Holding Out Relevant Tasks}
\label{sssec:hold-out-related}
\begin{figure}[t]
    \centering
    \includegraphics[width=1.\textwidth]{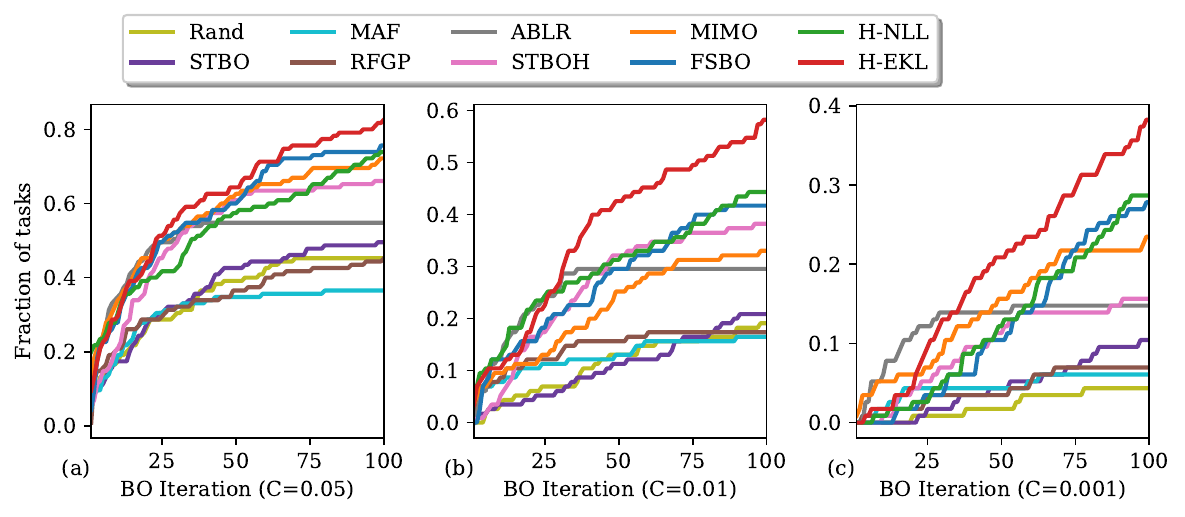}
    \caption{Performance profiles that show the fraction of tasks each method can solve, i.e., reach at most C amount of regret, in each BO iteration for the PD1 problem. In (a), C=0.05, in (b), C=0.01 and in (c), C=0.001. H-EKL was able to outperform other variants of \hyperbo and other baselines.}
    \label{fig:pp}
\end{figure}
We first conducted experiments in a setting where a new task dataset is presented, and a BO method tunes the optimizer hyperparameters for a selected model on that task dataset with a specific hardware. A training dataset (see terminology in \S\ref{sec:pf}) is composed of tuning sub-datasets from at most 18 training tasks that do not involve the same task dataset as the test task. For training, H-EKL can only pre-train on the matched dataset while all other meta BO methods can access both the matched and unmatched datasets in PD1. %

Figure~\ref{fig:pp} shows \emph{performance profiles} of all compared methods described in \S\ref{ssec:exp-methods}.  The performance profiles show the fraction of all test tasks that each method is able to solve by reaching at most 0.05, 0.01 and 0.001 regrets. The larger the fraction of tasks at each BO iteration, the better the method is. From all 3 criteria, we can see that \hyperbo methods, especially H-EKL solved more tasks than other methods. As the performance criteria becomes more stringent in Figure~\ref{fig:pp}(b) and Figure~\ref{fig:pp}(c), the \hyperbo variants, H-NLL and FSBO, also outperformed the baselines.

\begin{figure}[t]
    \centering
    \includegraphics[width=1.\textwidth]{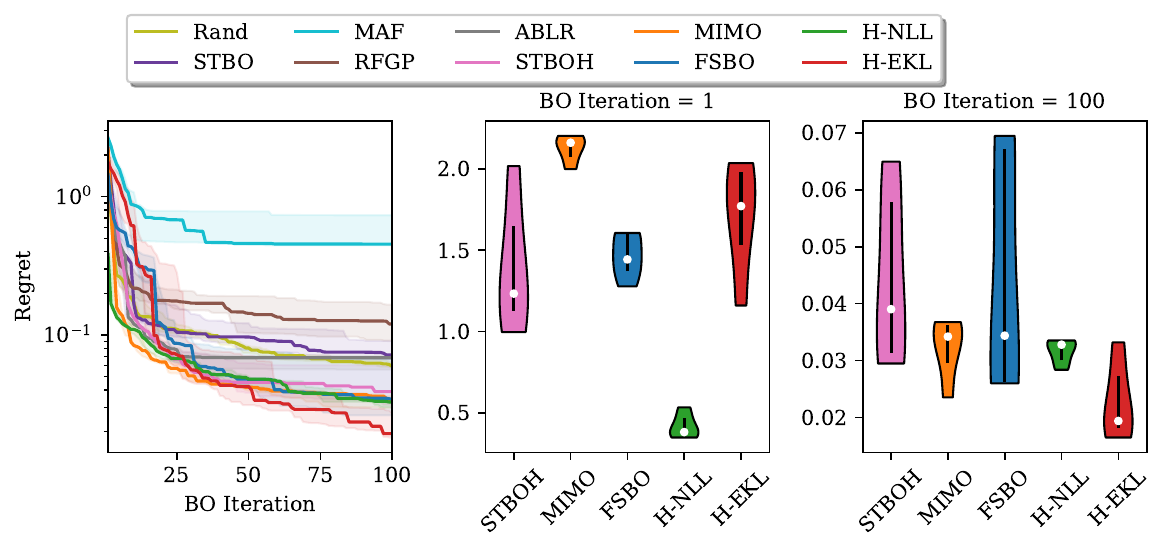}
    \caption{Regret curves for the PD1 problem. The violin plots (the middle and right plots) show the two vertical slices of the regret curves at the 1st and 100th iterations, where each white dot is the median and each black line is the 20/80 percentile. The \hyperbo variants, H-NLL, H-EKL and FSBO, performed competitively comparing to other methods. In particular, H-NLL achieved much lower regrets than other methods in the beginning BO iterations.}
    \label{fig:hold-out-related-summary}
\end{figure}
Figure~\ref{fig:hold-out-related-summary} illustrates the regret curves, together with the vertical slices at the 1st and 100th iterations. Rand fell behind most BO alternatives but outperformed MAF, RFGP and STBO in terms of average regrets. However, it is worth noticing that Figure~\ref{fig:pp}(c) shows MAF, RFGP and STBO can solve more tasks than Rand. ABLR also obtained high average regrets, but it solved more tasks than Rand for all criteria in Figure~\ref{fig:pp}. Among all transfer learning BO methods, MAF and RFGP did not seem to benefit from multi-task data, given their similar or worse performance than the single task method STBO. One problem with STBO is that it trains the GP on the data that the GP suggests to query, which is easy to overfit (as shown by Table~\ref{tab:hyperbo_posterior} in \S\ref{sssec:synthetic_more_datapoints}) if the priors on GP hyperparameters are not set carefully. %
With carefully hand-tuned priors, STBOH obtained better performance than STBO, ABLR, RFGP and MAF, showing the benefits of good priors designed with strong expert knowledge. MIMO surpassed STBOH, showing the potential of transfer learning in absence of expert knowledge. H-EKL did not locate good points in the beginning iterations, but found better points than other methods in later iterations; this was likely because our search spaces for mean and kernel functions encouraged the EKL objective (Eq.~\ref{eq:kl}) to overfit the mean function while matching the covariance matrix, which helps the BO performance once enough datapoints are collected.  

\begin{wrapfigure}{r}{0.45\textwidth}
\vspace{-1.5em}
\centering
    \includegraphics[width=.4\textwidth]{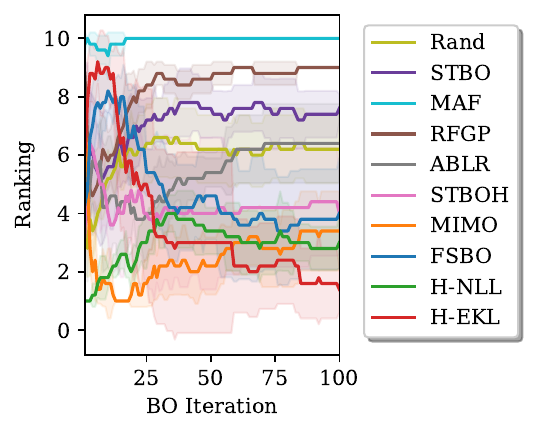}
    \vspace{-.5em}
\caption{The ranking plot for PD1. }
\label{fig:pd1_ranking}
\vspace{-1em}
\end{wrapfigure}
Figure~\ref{fig:pd1_ranking} presents the ranking plot. Among all \hyperbo variants, H-EKL was able to outperform all competing methods with BO iterations larger than 50.  ABLR used a much more constrained kernel function than other \hyperbo variants, resulting in worse performance. MIMO started off with a lower ranking, but it adapted well by re-training on new observations. The performance difference between H-NLL and FSBO shows that using a relatively complex mean function (as in H-NLL) can bring gains in performance comparing to using a zero mean (as in FSBO), especially in early stages of BO. %

\underline{\it Speedup.}\; To quantify \hyperbo's advantage in terms of sample efficiency, we computed how much faster \hyperbo can get a lower regret than the best alternative method; i.e., the median BO iterations needed for the best alternative to reach its lowest regret divided by the median BO iterations needed for \hyperbo to surpass that best alternative. 
We found that on average, H-NLL and H-EKL were at least 3 times faster than the best non-\hyperbo method and at least 7 times faster than Rand for the majority of the test tasks. %

\subsubsection{Correlations between the Test NLL, Pre-training Loss and BO Performance}
\label{sssec:pd1_nll}

Continuing the setup of \S\ref{sssec:hold-out-related} that holds out relevant tasks during pre-training, we explore how the pre-training loss correlates with the NLL on test data and BO performance. The pre-training losses, including EKL and NLL, are also measures of the GP regression performance. As a reference, we include quantitative results in Table~\ref{tab:summary} and Table~\ref{tab:nll}.  Table~\ref{tab:summary} shows the task-wise best validation error rates obtained by the top 6 methods in 100 BO iterations, and  Table~\ref{tab:nll} shows the Test NLL (Eq.~\ref{eq:singleml} on all available data of the test task in addition to observations) and EKL (Eq.~\ref{eq:multi-kl} computed by summing Eq.~\ref{eq:degenerate} on the matched data constructed from both training and test data) for each test task. Table~\ref{tab:summary} again confirms the importance of using more expressive mean and kernel functions for pre-trained GPs. In particular, H-EKL achieved the lowest error rates on over half of the test tasks.

\begin{figure}
    \centering
    \includegraphics[width=1.\textwidth]{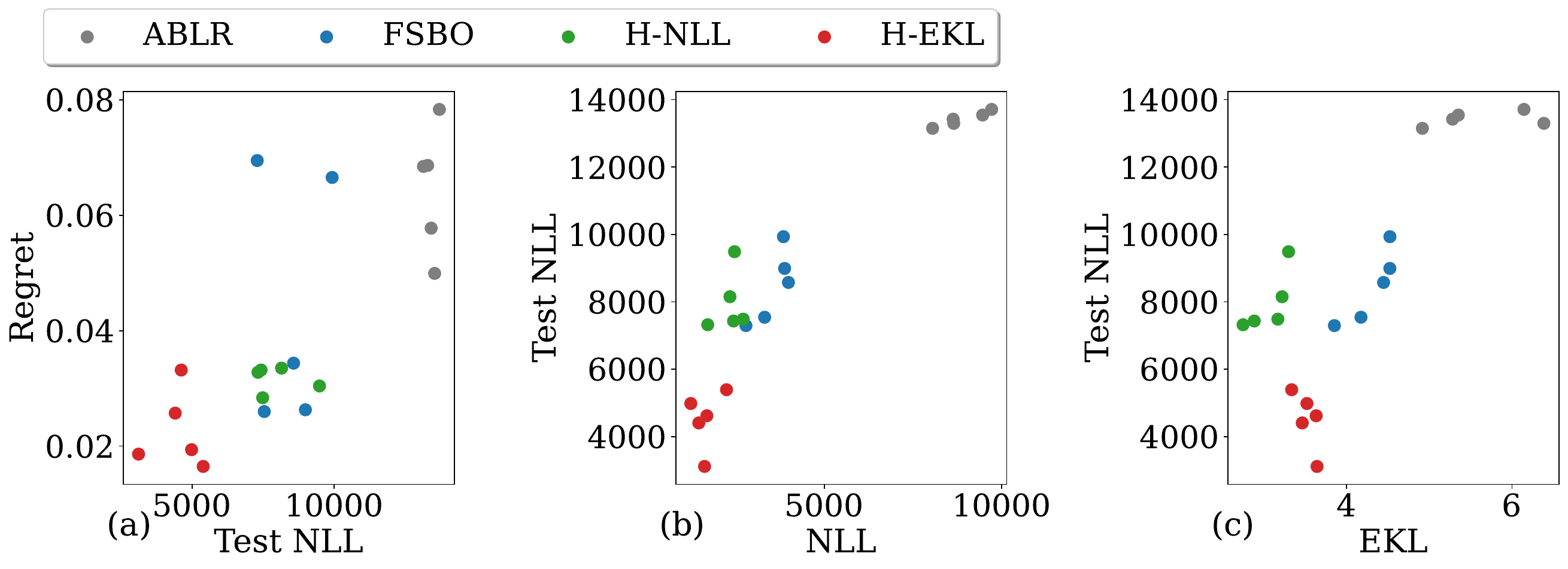}
    \caption{Correlations between (a) the simple regret at the 100th BO iteration and Test NLL, whose Pearson correlation coefficient is $0.77$; (b) Test NLL and NLL for the tasks irrelevant to the test task, whose Pearson correlation coefficient is $0.93$; and (c) Test NLL and EKL on all "matching-input" datapoints, whose Pearson correlation coefficient is $0.80$. Each point corresponds to a random repeat of the experiment, with metrics averaged over the test tasks. These plots confirm positive correlations between the simple regret, Test NLL, NLL and EKL, showing more evidence that optimizing the NLL (Eq.~\ref{eq:nll}) and EKL (Eq.~\ref{eq:multi-kl}) loss functions can bring better BO performance.}
    \label{fig:hold-out-related-correlation}
\end{figure}

Figure~\ref{fig:hold-out-related-correlation} shows the positive correlations between the simple regret, Test NLL, NLL (Eq.~\ref{eq:nll} on irrelevant tasks) and EKL. Each point in the plots of Figure~\ref{fig:hold-out-related-correlation} corresponds to an experiment, where for each test task, we pre-train a GP with random initialization, and we evaluate the metrics averaged over all test tasks. We repeated such experiment 5 times and hence there are 5 sets of regret and loss values for each method in Figure~\ref{fig:hold-out-related-correlation}. Together with Table~\ref{tab:summary} and Table~\ref{tab:nll}, we can tell that optimizing the NLL or EKL losses leads to a lower Test NLL, which correlates with a lower regret. Moreover, as shown in Table~\ref{tab:nll}, H-EKL may reach a lower Test NLL faster than H-NLL. And H-NLL may reach a lower EKL, which is evaluated on both training and test tasks, than H-EKL.

\begin{table}[t]
\centering
\tiny
\begin{tabular}{lllllll}\toprule
                               &                           ABLR &                          STBOH &                           MIMO &                           FSBO &                          H-NLL &                          H-EKL  \\ \hline
                WMT XFormer 64 &    { $\bm{33.91 \pm   0.02}$} &            $ 34.17 \pm   0.37$ &            $ 34.27 \pm   0.34$ &            $ 33.94 \pm   0.03$ &    { $\bm{ 33.91 \pm   0.02}$} &            $ 33.95 \pm   0.02$  \\ \hline
      Uniref50 Transformer 128 &            $ 79.19 \pm   0.69$ &            $ 78.92 \pm   0.27$ &            $ 79.00 \pm   0.72$ &            $ 78.81 \pm   0.22$ &            $ 78.85 \pm   0.26$ &    { $\bm{ 78.72 \pm   0.17}$}  \\ \hline
         LM1B Transformer 2048 &            $ 61.85 \pm   0.05$ &            $ 61.95 \pm   0.08$ &            $ 61.96 \pm   0.11$ &    { $\bm{ 61.81 \pm   0.02}$} &    { $\bm{ 61.81 \pm   0.03}$} &            $ 61.83 \pm   0.03$  \\ \hline
                 SVHN WRN 1024 &            $  4.07 \pm   0.01$ &            $  4.12 \pm   0.17$ &    { $\bm{  3.89 \pm   0.09}$} &            $  4.07 \pm   0.00$ &            $  4.07 \pm   0.00$ &            $  4.07 \pm   0.00$  \\ \hline
                  SVHN WRN 256 &            $  3.73 \pm   0.03$ &            $  3.72 \pm   0.06$ &    { $\bm{  3.61 \pm   0.06}$} &            $  3.74 \pm   0.04$ &            $  3.76 \pm   0.00$ &            $  3.69 \pm   0.02$  \\ \hline
         ImageNet ResNet50 256 &            $ 22.66 \pm   0.06$ &            $ 22.64 \pm   0.16$ &            $ 22.82 \pm   0.20$ &            $ 22.63 \pm   0.00$ &            $ 22.63 \pm   0.00$ &    { $\bm{ 22.53 \pm   0.00}$}  \\ \hline
         ImageNet ResNet50 512 &            $ 22.73 \pm   0.05$ &            $ 22.71 \pm   0.11$ &            $ 22.92 \pm   0.12$ &    { $\bm{ 22.63 \pm   0.00}$} &            $ 22.71 \pm   0.06$ &            $ 22.78 \pm   0.03$  \\ \hline
        MNIST CNNPoolTanh 2048 &            $  0.55 \pm   0.06$ &            $  0.53 \pm   0.02$ &            $  0.52 \pm   0.02$ &    { $\bm{  0.51 \pm   0.00}$} &    { $\bm{  0.51 \pm   0.00}$} &            $  0.53 \pm   0.01$  \\ \hline
         MNIST CNNPoolTanh 256 &            $  0.47 \pm   0.01$ &            $  0.47 \pm   0.02$ &            $  0.47 \pm   0.01$ &    { $\bm{  0.45 \pm   0.00}$} &    { $\bm{  0.45 \pm   0.00}$} &            $  0.47 \pm   0.01$  \\ \hline
        MNIST CNNPoolReLU 2048 &            $  0.74 \pm   0.02$ &            $  0.74 \pm   0.04$ &    { $\bm{  0.67 \pm   0.02}$} &            $  0.69 \pm   0.05$ &            $  0.71 \pm   0.03$ &            $  0.69 \pm   0.03$  \\ \hline
         MNIST CNNPoolReLU 256 &            $  0.51 \pm   0.01$ &            $  0.55 \pm   0.06$ &            $  0.51 \pm   0.02$ &            $  0.50 \pm   0.01$ &            $  0.50 \pm   0.00$ &    { $\bm{  0.48 \pm   0.01}$}  \\ \hline
            MNIST CNNReLU 2048 &            $  1.36 \pm   0.55$ &            $  1.19 \pm   0.18$ &            $  1.09 \pm   0.02$ &            $  1.23 \pm   0.18$ &            $  1.14 \pm   0.08$ &    { $\bm{  1.06 \pm   0.01}$}  \\ \hline
             MNIST CNNReLU 256 &            $  1.07 \pm   0.06$ &            $  1.05 \pm   0.03$ &            $  1.07 \pm   0.03$ &            $  1.40 \pm   0.41$ &    { $\bm{  1.03 \pm   0.00}$} &    { $\bm{  1.03 \pm   0.00}$}  \\ \hline
      Fashion CNNPoolTanh 2048 &            $  7.20 \pm   0.07$ &            $  7.06 \pm   0.14$ &            $  7.10 \pm   0.02$ &            $  7.04 \pm   0.07$ &            $  7.04 \pm   0.07$ &    { $\bm{  7.03 \pm   0.11}$}  \\ \hline
       Fashion CNNPoolTanh 256 &            $  6.52 \pm   0.05$ &            $  6.72 \pm   0.38$ &            $  6.42 \pm   0.17$ &            $  6.38 \pm   0.10$ &            $  6.28 \pm   0.03$ &    { $\bm{  6.26 \pm   0.00}$}  \\ \hline
      Fashion CNNPoolReLU 2048 &            $  7.62 \pm   0.05$ &    { $\bm{  7.45 \pm   0.10}$} &            $  7.55 \pm   0.02$ &            $  7.48 \pm   0.03$ &            $  7.50 \pm   0.05$ &            $  7.50 \pm   0.06$  \\ \hline
       Fashion CNNPoolReLU 256 &            $  6.77 \pm   0.09$ &            $  6.75 \pm   0.04$ &            $  7.02 \pm   0.04$ &            $  6.76 \pm   0.07$ &            $  6.79 \pm   0.11$ &    { $\bm{  6.70 \pm   0.05}$}  \\ \hline
          Fashion CNNReLU 2048 &            $  8.52 \pm   0.94$ &            $  7.56 \pm   0.02$ &            $  7.60 \pm   0.07$ &            $  7.73 \pm   0.36$ &            $  7.63 \pm   0.12$ &    { $\bm{  7.45 \pm   0.19}$}  \\ \hline
           Fashion CNNReLU 256 &            $  8.37 \pm   0.71$ &            $  7.41 \pm   0.23$ &            $  7.54 \pm   0.23$ &            $  7.58 \pm   0.51$ &    { $\bm{  7.36 \pm   0.40}$} &            $  7.48 \pm   0.54$  \\ \hline
             CIFAR100 WRN 2048 &            $ 21.79 \pm   0.32$ &            $ 20.78 \pm   0.43$ &            $ 20.90 \pm   0.65$ &            $ 20.87 \pm   0.11$ &            $ 21.22 \pm   0.37$ &    { $\bm{ 20.48 \pm   0.07}$}  \\ \hline
              CIFAR100 WRN 256 &            $ 19.20 \pm   0.23$ &            $ 19.03 \pm   0.09$ &            $ 19.36 \pm   0.20$ &    { $\bm{ 18.99 \pm   0.02}$} &            $ 19.00 \pm   0.00$ &            $ 19.03 \pm   0.07$  \\ \hline
              CIFAR10 WRN 2048 &            $  3.81 \pm   0.21$ &            $  3.42 \pm   0.16$ &            $  3.54 \pm   0.16$ &            $  3.38 \pm   0.04$ &            $  3.49 \pm   0.12$ &    { $\bm{  3.35 \pm   0.05}$}  \\ \hline
               CIFAR10 WRN 256 &            $  2.85 \pm   0.06$ &            $  2.88 \pm   0.14$ &            $  2.77 \pm   0.10$ &            $  2.83 \pm   0.03$ &            $  2.85 \pm   0.01$ &    { $\bm{  2.76 \pm   0.05}$}  \\
\bottomrule
\end{tabular}
\caption{The mean and standard deviation of best validation error rates ($\%$) for each test task in the PD1 problem. Meta BO methods including MIMO and \hyperbo variants (H-NLL, H-EKL, FSBO, ABLR) have access to training tasks that do not share the same task dataset as the test task. We show results of the top 6 methods, and we highlight the lowest error rates in bold.}
\label{tab:summary}
\end{table}

\begin{table}[htbp]
\centering
\tiny
\begin{tabular}{l llll llll}
\toprule
& \multicolumn{4}{c}{Test NLL} & \multicolumn{4}{c}{EKL} \\
\cmidrule(lr){2-5} \cmidrule(lr){6-9}

                               &                          ABLR &                          FSBO &                          H-NLL &                          H-EKL &                          ABLR &                          FSBO &                          H-NLL &                          H-EKL  \\ \hline
                WMT XFormer 64 &            $   -25 \pm      2$ &       $\bm{   -32 \pm      1}$ &            $   -32 \pm      2$ &            $   -31 \pm      1$ &            $   7.6 \pm    2.7$ &            $   5.1 \pm    1.3$ &            $   3.5 \pm    0.8$ &       $\bm{   3.5 \pm    0.2}$  \\ \hline
      Uniref50 Transformer 128 &            $   -30 \pm      2$ &       $\bm{   -36 \pm      2}$ &            $   -36 \pm      2$ &            $   -35 \pm      1$ &            $   7.6 \pm    3.1$ &            $   4.9 \pm    1.0$ &            $   3.7 \pm    1.3$ &       $\bm{   3.4 \pm    0.2}$  \\ \hline
         LM1B Transformer 2048 &            $   -27 \pm      1$ &       $\bm{   -32 \pm      2}$ &            $   -31 \pm      2$ &            $   -30 \pm      1$ &            $   6.4 \pm    1.8$ &            $   5.0 \pm    1.2$ &       $\bm{   3.4 \pm    0.9}$ &            $   3.5 \pm    0.1$  \\ \hline
                 SVHN WRN 1024 &            $   301 \pm     71$ &            $   333 \pm     70$ &            $   335 \pm     84$ &       $\bm{    44 \pm     10}$ &            $   6.7 \pm    2.7$ &            $   4.5 \pm    0.8$ &       $\bm{   3.3 \pm    0.6}$ &            $   3.5 \pm    0.2$  \\ \hline
                  SVHN WRN 256 &            $   294 \pm     55$ &            $   308 \pm     82$ &            $   311 \pm    108$ &       $\bm{    12 \pm     11}$ &            $   6.7 \pm    2.7$ &            $   4.5 \pm    0.8$ &       $\bm{   3.3 \pm    0.6}$ &            $   3.5 \pm    0.1$  \\ \hline
         ImageNet ResNet50 256 &            $     1 \pm      3$ &       $\bm{   -23 \pm      1}$ &            $   -21 \pm      1$ &            $   -16 \pm      2$ &            $   4.4 \pm    0.2$ &            $   3.9 \pm    0.1$ &       $\bm{   2.7 \pm    0.1}$ &            $   3.6 \pm    0.1$  \\ \hline
         ImageNet ResNet50 512 &            $    -2 \pm      6$ &       $\bm{   -22 \pm      1}$ &            $   -20 \pm      1$ &            $   -15 \pm      2$ &            $   4.4 \pm    0.1$ &            $   3.9 \pm    0.1$ &       $\bm{   2.7 \pm    0.1}$ &            $   3.6 \pm    0.1$  \\ \hline
        MNIST CNNPoolTanh 2048 &            $   165 \pm     17$ &            $    83 \pm     13$ &            $    68 \pm      5$ &       $\bm{    34 \pm      4}$ &            $   5.0 \pm    0.3$ &            $   4.1 \pm    0.2$ &       $\bm{   2.7 \pm    0.1}$ &            $   3.4 \pm    0.1$  \\ \hline
         MNIST CNNPoolTanh 256 &            $   199 \pm     22$ &            $    51 \pm     13$ &            $    45 \pm     11$ &       $\bm{    38 \pm      4}$ &            $   4.7 \pm    0.3$ &            $   4.0 \pm    0.1$ &       $\bm{   2.7 \pm    0.1}$ &            $   3.4 \pm    0.2$  \\ \hline
        MNIST CNNPoolReLU 2048 &            $   592 \pm     30$ &            $   537 \pm     59$ &            $   483 \pm     13$ &       $\bm{   425 \pm    111}$ &            $   9.4 \pm    1.3$ &            $   6.3 \pm    0.6$ &            $   5.4 \pm    0.6$ &       $\bm{   3.5 \pm    0.2}$  \\ \hline
         MNIST CNNPoolReLU 256 &            $   278 \pm     25$ &            $    66 \pm     33$ &            $    45 \pm      5$ &       $\bm{    28 \pm     10}$ &            $   5.0 \pm    0.7$ &            $   4.0 \pm    0.3$ &       $\bm{   2.8 \pm    0.1}$ &            $   3.5 \pm    0.1$  \\ \hline
            MNIST CNNReLU 2048 &            $   386 \pm     22$ &            $   280 \pm     37$ &       $\bm{   276 \pm     38}$ &            $   289 \pm     67$ &            $   6.9 \pm    1.2$ &            $   4.6 \pm    0.4$ &       $\bm{   3.5 \pm    0.4}$ &            $   3.6 \pm    0.1$  \\ \hline
             MNIST CNNReLU 256 &            $   275 \pm     21$ &            $   155 \pm     12$ &       $\bm{   138 \pm      9}$ &            $   245 \pm     43$ &            $   4.9 \pm    0.1$ &            $   4.4 \pm    0.3$ &       $\bm{   3.0 \pm    0.1}$ &            $   3.8 \pm    0.1$  \\ \hline
      Fashion CNNPoolTanh 2048 &            $    60 \pm      6$ &            $     0 \pm      1$ &            $     3 \pm      5$ &       $\bm{    -6 \pm      2}$ &            $   5.1 \pm    0.4$ &            $   3.9 \pm    0.2$ &       $\bm{   2.6 \pm    0.0}$ &            $   3.5 \pm    0.2$  \\ \hline
       Fashion CNNPoolTanh 256 &            $    45 \pm      7$ &            $    11 \pm      2$ &            $    13 \pm      3$ &       $\bm{     1 \pm      1}$ &            $   5.0 \pm    0.2$ &            $   3.9 \pm    0.2$ &       $\bm{   2.6 \pm    0.0}$ &            $   3.5 \pm    0.2$  \\ \hline
      Fashion CNNPoolReLU 2048 &            $    82 \pm      4$ &            $    26 \pm      4$ &            $    24 \pm      4$ &       $\bm{     9 \pm      2}$ &            $   5.0 \pm    0.2$ &            $   3.9 \pm    0.2$ &       $\bm{   2.6 \pm    0.0}$ &            $   3.5 \pm    0.2$  \\ \hline
       Fashion CNNPoolReLU 256 &            $    56 \pm      8$ &            $     9 \pm      4$ &            $     9 \pm      4$ &       $\bm{    -8 \pm      1}$ &            $   4.9 \pm    0.2$ &            $   3.9 \pm    0.2$ &       $\bm{   2.6 \pm    0.0}$ &            $   3.5 \pm    0.2$  \\ \hline
          Fashion CNNReLU 2048 &            $    75 \pm      3$ &            $    37 \pm      6$ &       $\bm{    29 \pm      3}$ &            $    43 \pm      6$ &            $   5.1 \pm    0.4$ &            $   3.9 \pm    0.2$ &       $\bm{   2.6 \pm    0.0}$ &            $   3.5 \pm    0.2$  \\ \hline
           Fashion CNNReLU 256 &            $    53 \pm      9$ &            $    15 \pm      4$ &       $\bm{    12 \pm      2}$ &            $    29 \pm      9$ &            $   5.1 \pm    0.3$ &            $   3.9 \pm    0.2$ &       $\bm{   2.6 \pm    0.0}$ &            $   3.5 \pm    0.2$  \\ \hline
             CIFAR100 WRN 2048 &            $    10 \pm      5$ &            $    -7 \pm      3$ &            $    -6 \pm      2$ &       $\bm{   -10 \pm      1}$ &            $   5.2 \pm    1.0$ &            $   4.3 \pm    0.9$ &       $\bm{   3.2 \pm    0.8}$ &            $   3.6 \pm    0.1$  \\ \hline
              CIFAR100 WRN 256 &            $     0 \pm      3$ &            $   -10 \pm      2$ &            $   -10 \pm      2$ &       $\bm{   -12 \pm      2}$ &            $   5.3 \pm    0.9$ &            $   4.3 \pm    0.9$ &       $\bm{   3.2 \pm    0.8}$ &            $   3.6 \pm    0.1$  \\ \hline
              CIFAR10 WRN 2048 &            $   187 \pm     61$ &            $   137 \pm     50$ &            $   148 \pm     69$ &       $\bm{     1 \pm      4}$ &            $   4.4 \pm    0.2$ &            $   4.0 \pm    0.2$ &       $\bm{   2.7 \pm    0.2}$ &            $   3.6 \pm    0.1$  \\ \hline
               CIFAR10 WRN 256 &            $   114 \pm     31$ &            $    63 \pm     18$ &            $    53 \pm     13$ &       $\bm{     1 \pm      3}$ &            $   4.4 \pm    0.2$ &            $   4.0 \pm    0.2$ &       $\bm{   2.7 \pm    0.2}$ &            $   3.6 \pm    0.1$  \\
               
\bottomrule
\end{tabular}
\caption{The mean and standard deviation of Test NLL ($\times 100$) and EKL for each test task in the offline optimizer hyperparameter tuning experiments. \hyperbo variants only have access to training tasks that do not share the same task dataset as the test task. We highlight the lowest Test NLL and the lowest EKL in bold.}
\label{tab:nll}
\end{table}

\subsubsection{Effects of the Number of Training Datapoints}
\label{sssec:exp-pd1-datapoints}
In this section, we answer the question: how does $M_i$ in \S\ref{sec:pf}, the number of datapoints in each training task, impact the performance of meta BO methods? %
We used data from all but the test task for meta BO methods, with the maximum number of training datapoints per task ranging from 4 to about 1600. The training datapoints were selected uniformly randomly from tuning sub-datasets. Since there are in total 242 valid ``matching-input'' datapoints across training tasks in PD1, we only show the regrets of H-EKL up to 242 datapoints per training task. We did not include MAF in the following experiments due to its high computational costs.

\begin{figure}
    \centering
    \includegraphics[width=\textwidth]{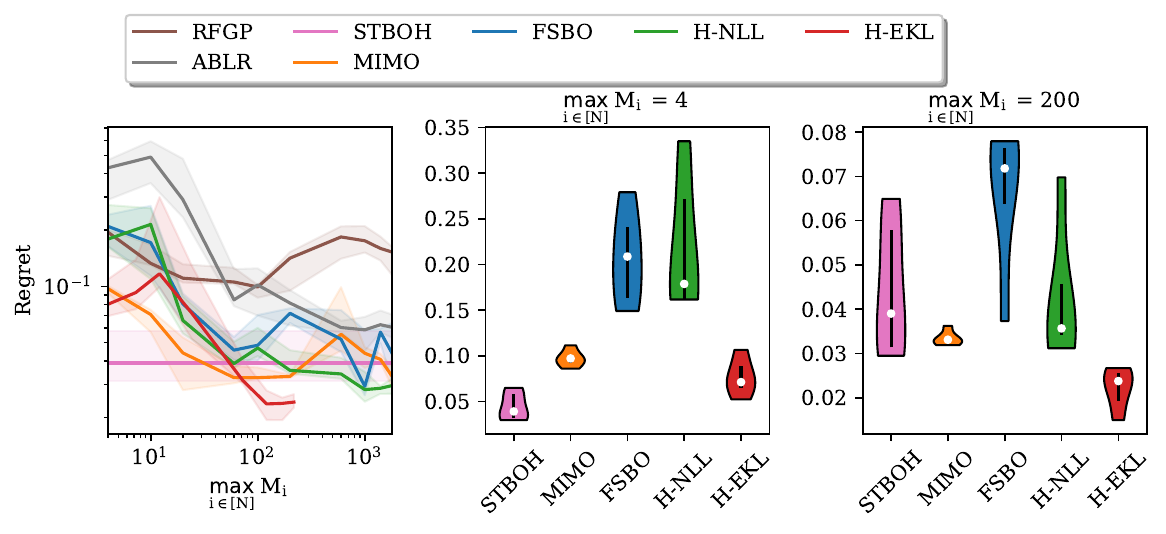}
    \caption{The median and 20/80 percentiles of simple regrets as the maximum number of datapoints per training task ($\max_{i\in[N]} M_i$) changes from 4 to about 1600 for each method. We also show the two vertical slices for $\max_{i\in[N]} M_i=4$ (middle) and $\max_{i\in[N]} M_i=200$ (right). H-EKL used significantly fewer training datapoints to obtain lower regrets than other meta BO methods.}
    \label{fig:p-remove} 
\end{figure}

Figure~\ref{fig:p-remove} shows how the simple regrets (aggregated over 23 test tasks at the 100th BO iteration) change as we increase the maximum number of datapoints per training task ($\max_{i\in[N]} M_i$). Not surprisingly, with only 4 datapoints per training task, all meta BO methods obtained worse performance than STBOH. As the number of datapoints increased to about 100, H-EKL outperformed STBOH by a large margin. And with about 200 datapoints per training task or more, both H-EKL and H-NLL surpassed STBOH. 
We observed clear trend that more training datapoints can lead to lower regrets for H-EKL, H-NLL and ABLR. FSBO and MIMO benefited from more training data but showed more variations in regrets after $\max_{i\in[N]} M_i$ reached about 100.

\subsubsection{Effects of the Number of Training Tasks}
\label{ssec:num_training_tasks}

\begin{figure}
    \centering
    \includegraphics[width=\textwidth]{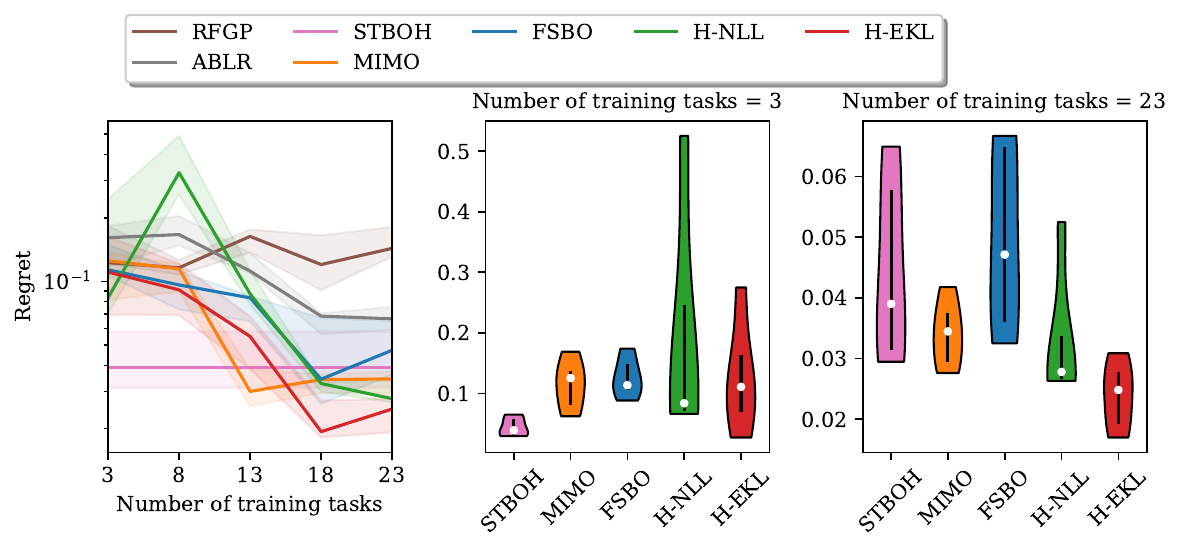}
    \caption{The median and 20/80 percentiles of simple regrets for each method with 3 to 23 training tasks. The violin plots for the top performing methods are shown for 3 and 23 training tasks, where each white dot is the median and black line the 20/80 percentiles. All meta BO methods except RFGP got lower regrets with more training tasks. H-NLL and H-EKL achieved better performance than other baselines with 23 training tasks.}
    \label{fig:hold-out-related} 
\end{figure}

We investigate the impact of the number of training tasks on the performance of meta BO methods in this section.  Figure~\ref{fig:hold-out-related} demonstrates how the regrets change as we increase the number of training tasks. We show the aggregated simple regrets of meta BO methods at the 100th BO iteration over the 23 test tasks. To reduce training tasks, we first remove the tasks that involve the same task dataset as the test task, and then remove others randomly until we reach the designated number of training tasks. This is to analyze the performance of all methods on less-related tasks. The models were never trained on the data from the test task.  %

\begin{figure}
    \centering
    \includegraphics[width=1.\textwidth]{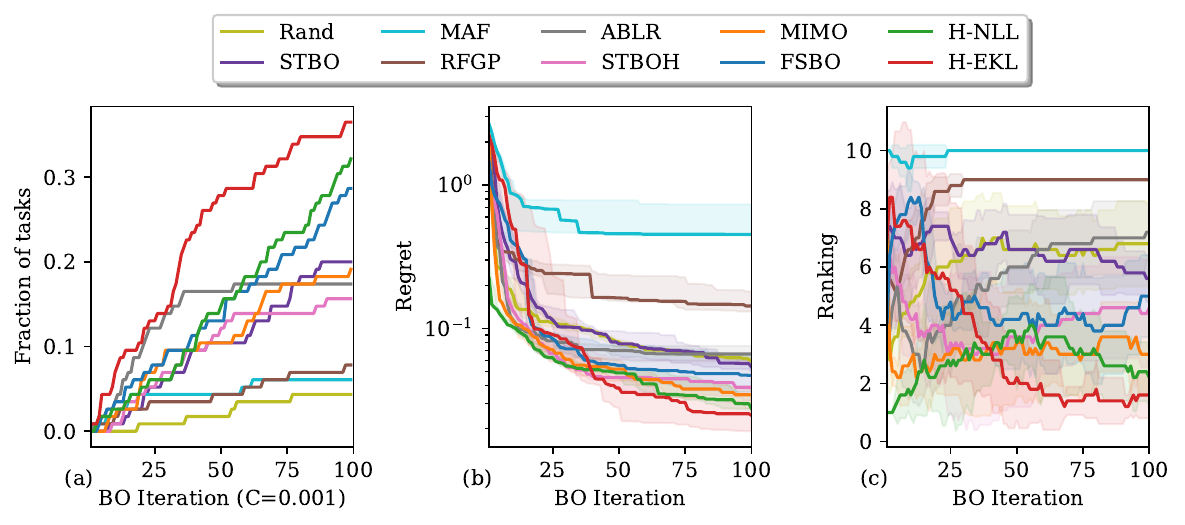}
    \caption{(a-c) are the performance profiles, regret curves and ranking plots for meta BO methods with 23 training tasks. As shown in (c), H-NLL is the best performing method in the first 30 BO iterations and H-EKL becomes the best afterwards. Both H-EKL and H-NLL outperformed baselines.}
    \label{fig:pd1_23}
\end{figure}

All meta BO methods except RFGP reduced simple regrets as more training tasks were given. In both Figure~\ref{fig:p-remove} and Figure~\ref{fig:hold-out-related} , RFGP curiously did not seem to be able to learn from multi-task data. The worse performance of RFGP was likely caused by the high confidence predictions from random feature based Bayesian linear regression models as discussed by~\cite{wang2018batched}; overly confident predictions can lead to less exploration in BO and as a result, poor BO performance. ABLR also uses Bayesian linear regression and likely suffered from the same lack of exploration issue. With 18 or more training tasks, H-EKL and H-NLL outperformed the carefully hand-tuned STBOH. Figure~\ref{fig:pd1_23} shows the superior performance of H-EKL and H-NLL comparing to other baselines in the setting where 23 training tasks are given. %

\subsection{Results on the PD1 Online Hyperparameter Tuning Tasks}
\label{ssec:online}
\begin{figure}
    \centering
    \includegraphics[width=\textwidth]{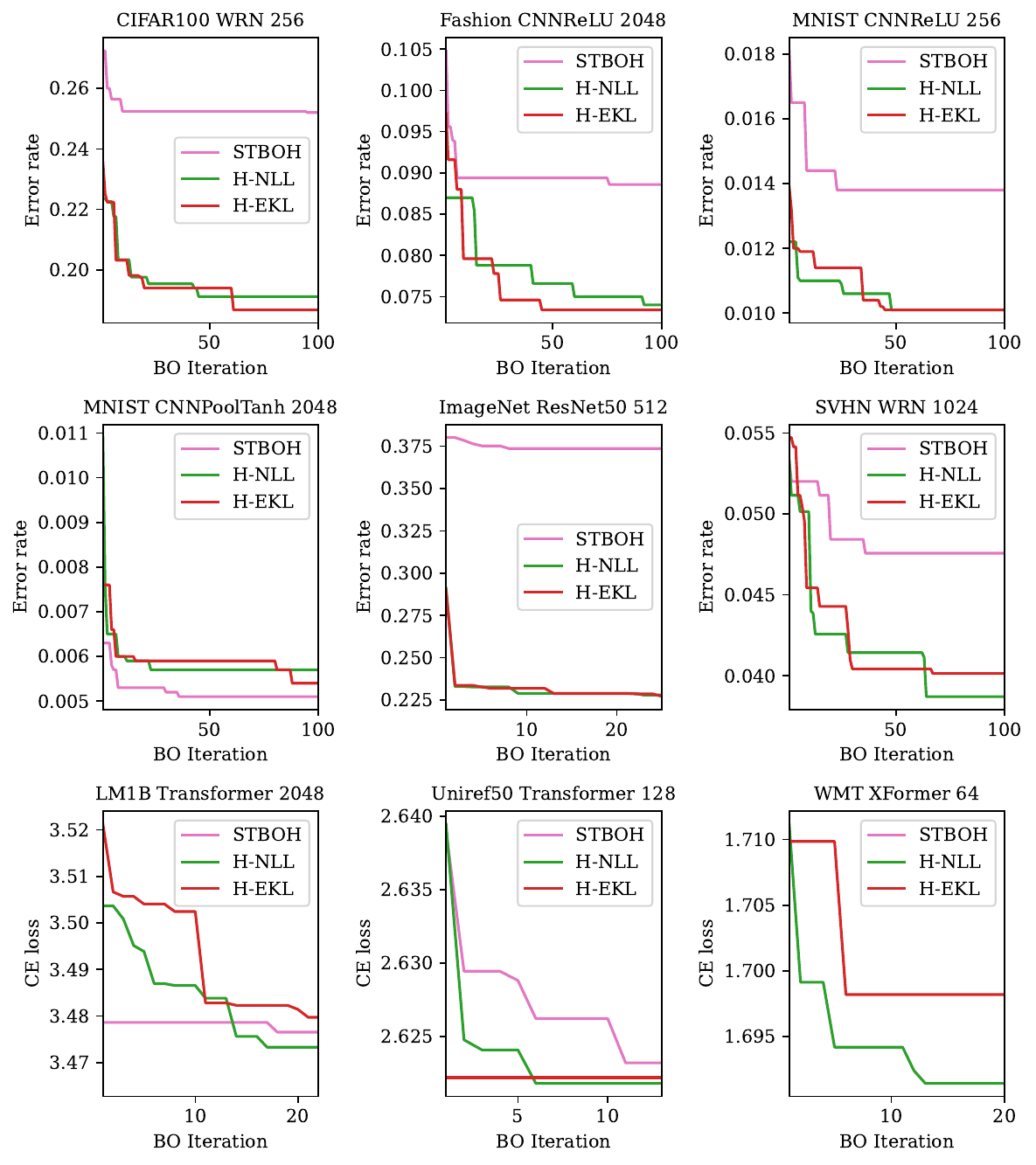}
    \caption{PD1 online hyperparameter tuning results for image based tasks with the best validation error rate as the objective, and text based tasks that use the best validation cross entropy (CE) loss as the objective. HyperBO methods achieved better performance in 8 out of 9 online tuning tasks. The CE losses of STBOH are not shown for WMT XFormer 64 because all of the datapoints acquired by STBOH on this task were infeasible.}
    \label{fig:online}
\end{figure}
In this section, we look into the online hyperparameter tuning setting where we deploy \hyperbo as a BO service and optimize over the full hyperrectangular search space detailed in Table~\ref{tab:pd1_search_space}. In the online setting, some combinations of hyperparameters may be \emph{infeasible} to evaluate. For example, an overly big learning rate may lead to divergence in gradients, in which case we do not obtain a valid model. To address this, we pre-process the function values to $[-2, 2)$ such that \emph{infeasible} evaluations map to $-2$, while bad evaluations approach asymptotically to $-2$. More precisely, for each training sub-dataset $D_{f_{i}}$, we apply for each successful $y \in \{y_j^{(i)}\}_{j=1}^{M_i}$ the following mapping:
\begin{align*}
    y \gets \frac{\mathrm{softplus}(y - \overline{y})}{\mathrm{softplus}(y_{\max} -\overline{y})} * 4 - 2
\end{align*}
where $\overline{y}$ is the median and $y_{\max}$ is the maximum of $\{y_j^{(i)}\}_{j=1}^{M_i}$. In every iteration $t$ of \hyperbo in Algorithm~\ref{alg:hyperbo}, we also apply the same type of mapping to the test sub-dataset $D_f = \{(x_\tau, y_\tau)\}_{\tau=1}^{t-1}$. For all methods that we compared, model re-training was performed on $D_f$ to allow this dynamic adjustment of the evaluations on the test function.

 Figure~\ref{fig:online} presents the online tuning results for image and text based tasks\footnote{We only experimented on the 9 tasks selected from Table~\ref{tab:workloads} due to limited compute resources.}. The goals of  image based tasks are to minimize the validation error rate, while the text based tasks (LM1B, Uniref50, WMT) aim to minimize the cross entropy (CE) loss.

We reconfigured H-NLL and H-EKL to use a one-hidden layer neural network of size 8 as mean function and Mat\'ern32 covariance function on the feature layer of the neural net as kernel. For H-EKL, we added 0.1 times the NLL objective to the EKL objective to allow re-training the model during online evaluations. For all methods, we used GP-UCB with coefficient $1.8$.

 In our experiments, we noticed that it was very difficult for STBO, MIMO and RFGP to recover from a ``bad'' datapoint, and their BO results were a lot worse than STBOH, H-NLL and H-EKL. This was partly because predictions from these models were significantly tied to the initial observations. For example, STBO may overfit to the initial bad value and believe there are bad values in the entire search space.  Figure~\ref{fig:online} presents the results of STBOH, H-NLL and H-EKL for clarity and Figure~\ref{fig_app:online} shows the results that compare more methods.
 
Figure~\ref{fig:online} shows the robust performance of H-NLL and H-EKL in the online tuning problems. In 8 out of 9 tuning tasks, \hyperbo methods outperformed STBOH.

\subsection{Results on the HPO-B Benchmark}
\label{ssec:hpob}

In this experiment, we evaluate the performance of \hyperbo on the HPO-B benchmark~\citep{pineda2021hpob}. The HPO-B benchmark is a machine learning hyperparameter tuning dataset, which includes about 6 million evaluations of hyperparameters from 16 search spaces of different models. Each search space has different sets of hyperparameters with dimensions ranging from 2 to 18. There are multiple tasks in each search spaces, which are divided to training and test tasks. In total, there are 86 test tasks. Please refer to \cite{pineda2021hpob} for more details. 

Following \cite{wistuba2021few}, we used 2-hidden layer neural nets of size (128, 128) to replace the neural nets of \hyperbo variants, and normalized all the regrets by the range of function values of each test task. The BO experiments were conducted with the original function values. The experiments were repeated 25 times with 5 different random seeds, and for each seed, 5 different test sub-dataset initialization settings specified in the HPO-B dataset. Similar to \S\ref{ssec:offline_pd1}, we adopted the offline evaluation setup on finite sets of datapoints in the test tasks of HPO-B. Due to the lack of matched datasets, we excluded H-EKL.

 In the following, \S\ref{sssec:hpob_full} shows aggregated BO results on the original HPO-B. \S\ref{ssec:hpob_reduced} aims to understand the impacts of reduced training data on \hyperbo and other transfer learning BO methods. \S\ref{sssec:negative_transfer} presents both the regret and GP regression results to investigate the ``negative transfer''~\citep{rothfuss2021pacoh} effect.

\subsubsection{Pre-training on the Full HPO-B Dataset}
\label{sssec:hpob_full}

\begin{figure}
    \centering
    \includegraphics[width=1.\textwidth]{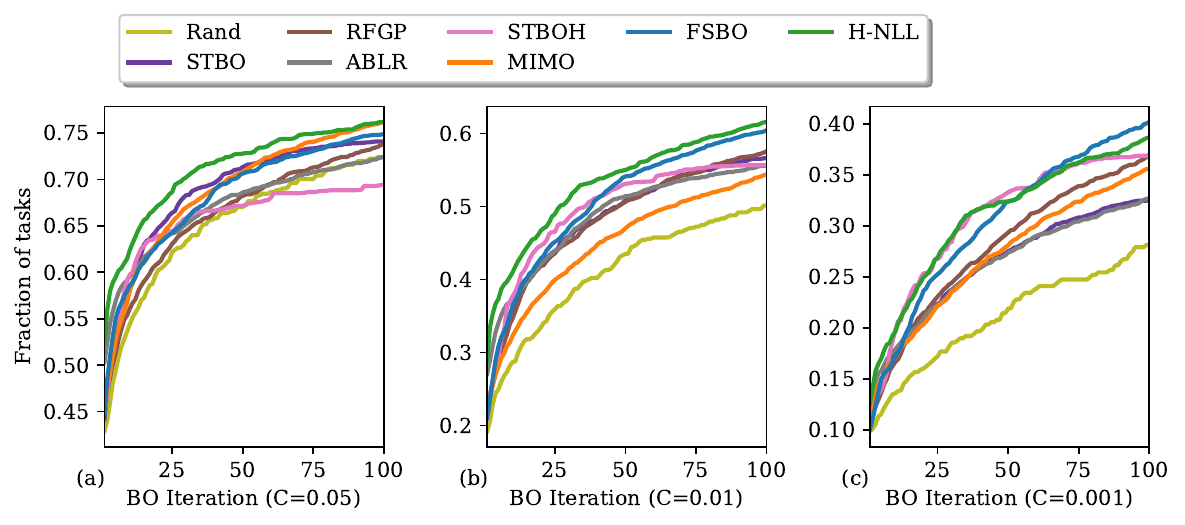}
    \caption{ Performance profiles that show the fraction of tasks each method can solve, i.e., reach at most C amount of regret, in each BO iteration for the HPO-B experiment. In (a), C=0.05, in (b), C=0.01 and in (c), C=0.001. H-NLL and FSBO performed better than or similar to the best non-HyperBO alternatives. H-NLL solved more tasks than other methods for C=0.05 and C=0.01.}
    \label{fig:hpob-pp}
\end{figure}

We first used the full HPO-B dataset from each search space for meta BO methods. Figure~\ref{fig:hpob-pp} shows the performance profiles. The two \hyperbo variants, H-NLL and FSBO achieved top performance. H-NLL was able to solve more tasks than other methods for the success criteria C=0.05 and C=0.01. For C = 0.001, FSBO solved slightly more tasks than H-NLL after about 60 BO iterations. 

\begin{figure}
    \centering
    \includegraphics[width=1.\textwidth]{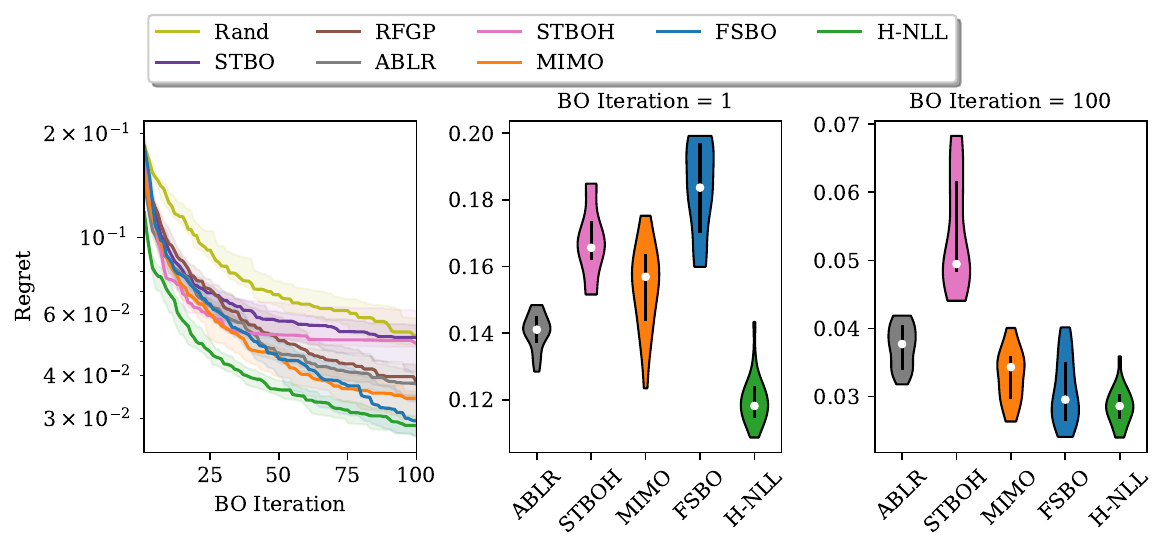}
    \caption{Regret curves for the HPO-B experiment. The violin plots (the middle and right plots) show the two vertical slices of the regret curves at the 1st and 100th iterations, where each white dot is the median and each black line is the 20/80 percentile. H-NLL achieved the lowest regrets across all BO iterations, outperforming FSBO, a \hyperbo variant with 0 mean. For this experiment, having a more flexible mean function, as in H-NLL, helped in obtaining better performance.}
    \label{fig:hpob-regret}
\end{figure}

Figure~\ref{fig:hpob-regret} presents the regret curves of all methods. All BO methods outperformed Rand, and all meta BO methods performed better than single task BO methods (STBO and STBOH) after about 50 BO iterations.  H-NLL was able to locate better hyperparameters than all other methods across BO iterations, demonstrating the sample efficiency of \hyperbo. 

\begin{wrapfigure}{r}{0.45\textwidth}
\centering
\vspace{-.5em}
    \includegraphics[width=.4\textwidth]{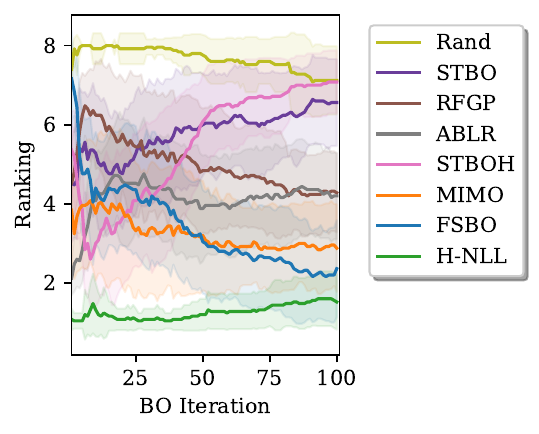}
    \vspace{-.5em}
\caption{Ranking the compared methods for the HPO-B experiment.}
 \vspace{-1.5em}
\label{fig:hpob_ranking}
\end{wrapfigure}

Figure~\ref{fig:hpob_ranking} shows the corresponding ranking plot. H-NLL consistently ranked 1st across all iterations. H-NLL and FSBO are both \hyperbo variants, and the more flexible mean function enabled H-NLL to acquire datapoints with lower regrets sooner than FSBO.

STBOH is equipped with a carefully hand-tuned prior, and it obtained lower regrets than other methods (except H-NLL) in some early iterations.  
Unlike the results in \S\ref{ssec:offline_pd1}, the performance of STBOH plateaued quickly in the HPO-B experiment. It can be difficult for hand-tuned priors to generalize across a variety of tasks, and it is impractical to require more hand-tuning for new tasks. \hyperbo provides an automatic model pre-training procedure to efficiently obtain customized priors, making this prior specification process much easier and more effective.

\vspace{.5em}
\underline{\it Speedup.}\; To compactly quantify the performance, we computed the speedup of \hyperbo methods in the same way as was done in \S\ref{sssec:hold-out-related}. We found that on average, H-NLL was at least 6 times faster than FSBO, 10 times faster than the best non-HyperBO alternative, and 16 times faster than Rand for the majority of the test tasks. %

\subsubsection{Pre-training on HPO-B with Reduced Training Data}
\label{ssec:hpob_reduced}
In this section, we investigate how the performance of pre-trained GPs changes as the number of training datapoints decreases for the tasks in HPO-B. To answer this question, we downsampled the HPO-B training datasets in each search space. For each training sub-dataset, we uniformly randomly sampled 10, 100 and 1000 training datapoints\footnote{Among the 16 search spaces in HPO-B, 5 search spaces have at most 1000 datapoints (ranging from 259 to 895) in their training tasks. In other search spaces, the maximum number of datapoints for their corresponding training tasks ranges from 1,593 to 102,306.} for the meta BO methods to perform transfer learning. %

\begin{figure}
    \centering
    \includegraphics[width=\textwidth]{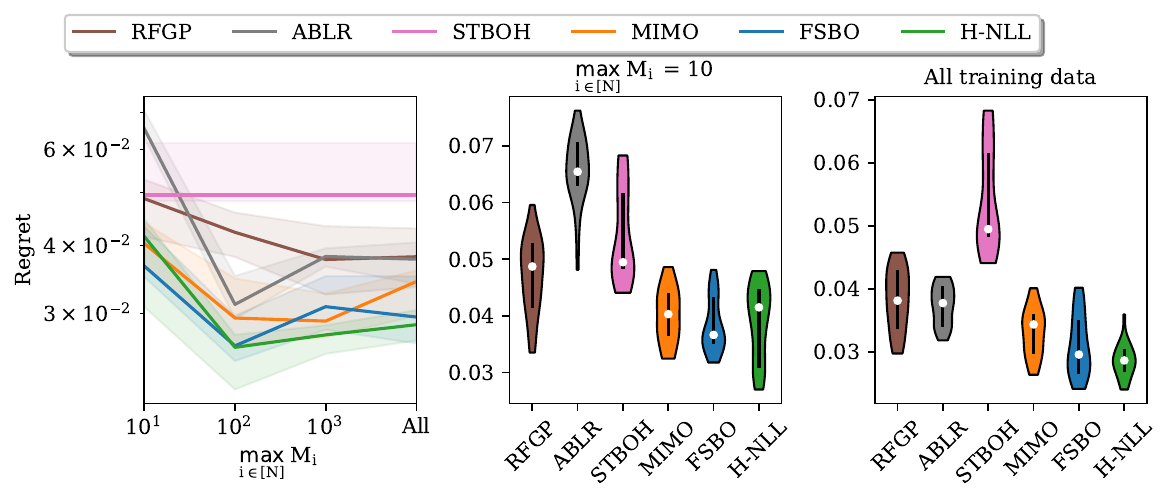}
    \caption{The median and 20/80 percentiles of simple regrets for meta BO methods as the maximum number of datapoints per training task in HPO-B ($\max_{i\in[N]} M_i$) increases. We randomly downsampled each training sub-dataset to 10, 100, 1000 datapoints or used all training data for transfer learning. H-NLL and ABLR had the most amount of improvements when increasing the number of training datapoints per task from 10 to 100. }
    \label{fig:hpob_nremain} 
\end{figure}

Figure~\ref{fig:hpob_nremain} shows the simple regrets at the 100th BO iteration of the meta BO methods that use different numbers of training datapoints per task, together with STBOH as a reference. All meta BO methods benefited from more training data as we increased the number from 10 per task to 100 per task. H-NLL and FSBO both obtained relatively better performance overall, followed by MIMO. RFGP and ABLR performed worse than other meta BO methods, similar to what we observed from a similar experiment for PD1 in Figure~\ref{fig:p-remove}. 

\begin{figure}
    \centering
    \includegraphics[width=1.\textwidth]{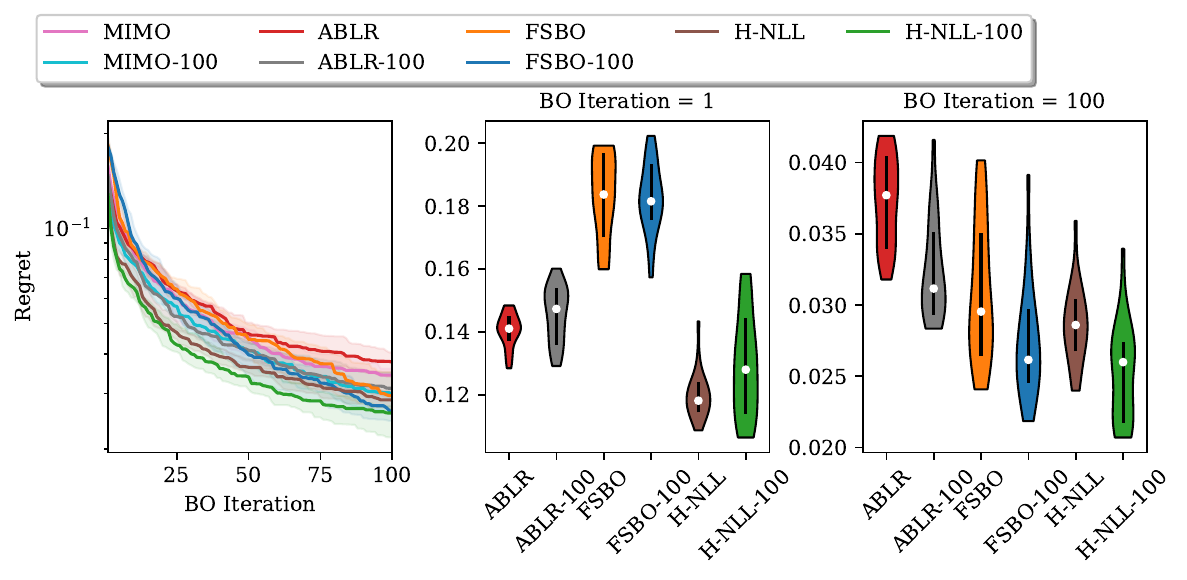}
    \caption{Regret curves to compare meta BO methods with at most 100 training datapoints per HPO-B training task (suffix ``-100'') versus all training data. We also show two vertical slices of the regret curves at the 1st and 100th iteration, where the white dot is the median and the black line is the 20/80 percentile. H-NLL-100 achieved top overall performance across BO iterations, surpassing H-NLL, which had access to the full training data.}
    \label{fig:hpob-regret-100-10000}
\end{figure}

For the cases of above 100 datapoints per task, the performance of \hyperbo variants and MIMO slightly worsened. Figure~\ref{fig:hpob-regret-100-10000} compares the regret curves of meta BO methods with max 100 datapoints per training task versus with all training data in HPO-B. Meta BO methods with max 100 training datapoints per task interestingly unanimously outperformed their counterparts with the full training dataset, indicating redundancy and noise in the HPO-B dataset. 

\subsubsection{The ``Negative Transfer'' Effect in HPO-B}
\label{sssec:negative_transfer}
``Negative transfer'' refers to the situations where transfer learning negatively impacts the performance for test tasks~\citep{rothfuss2021pacoh}. This effect can be quantitatively shown by comparing the Training NLL and the regret. A positive correlation between the Training NLL and the regret means that optimizing the pre-training objective can positively contribute to the BO performance evaluated by regrets, which is the case for PD1 as shown in Figure~\ref{fig:hold-out-related-correlation}. On the other hand, a negative correlation between the Training NLL and the regret indicates ``negative transfer'', which we observed in some search spaces of HPO-B.

\tabcolsep=0.19cm
\begin{table}[t]
\centering
\tiny
\begin{tabular}{l l ll ll ll l}
\toprule
& \multicolumn{1}{c}{MIMO} & \multicolumn{2}{c}{ABLR} & \multicolumn{2}{c}{FSBO} & \multicolumn{2}{c}{H-NLL} &\\
\cmidrule(lr){2-2} \cmidrule(lr){3-4} \cmidrule(lr){5-6} \cmidrule(lr){7-8}
                               &                         Regret &                   Training NLL &                         Regret &                   Training NLL &                         Regret &                   Training NLL &                         Regret &                        Pearson  \\ \hline
                          4796 &            $ 21.62 \pm  17.66$ &   {  $\bm{  -402 \pm      3}$} &    { $\bm{  1.73 \pm   0.25}$} &            $  -357 \pm      3$ &    { $\bm{  1.73 \pm   0.25}$} &            $  -381 \pm      4$ &            $  2.24 \pm   0.25$ &                       $  0.05$  \\ \hline
                          5527 &    { $\bm{ 26.29 \pm   6.77}$} &   {  $\bm{  -342 \pm      9}$} &            $ 27.43 \pm   1.79$ &            $  -314 \pm      7$ &            $ 31.12 \pm   1.35$ &            $  -314 \pm     10$ &            $ 33.40 \pm   1.24$ &                       $  0.71$  \\ \hline
                          5636 &            $114.56 \pm  32.41$ &   {  $\bm{  -417 \pm     26}$} &            $ 96.45 \pm   3.86$ &            $  -336 \pm     13$ &            $ 49.39 \pm   3.77$ &            $  -356 \pm      7$ &    { $\bm{ 26.50 \pm   2.09}$} &                       $ -0.78$  \\ \hline
                          5859 &            $ 54.41 \pm  11.96$ &   {  $\bm{  -416 \pm     18}$} &            $ 23.54 \pm   2.45$ &            $  -316 \pm     16$ &    { $\bm{ 20.24 \pm   2.73}$} &            $  -342 \pm     12$ &            $ 25.16 \pm   1.36$ &                       $ -0.32$  \\ \hline
                          5860 &            $ 12.29 \pm  16.43$ &            $  -221 \pm     32$ &            $  1.08 \pm   0.21$ &            $  -304 \pm     21$ &    { $\bm{  0.00 \pm   0.00}$} &   {  $\bm{  -325 \pm     13}$} &    { $\bm{  0.00 \pm   0.00}$} &                       $  0.88$  \\ \hline
                          5889 &            $ 30.97 \pm  31.84$ &   {  $\bm{  -308 \pm      1}$} &            $ 41.32 \pm  17.08$ &            $  -301 \pm      9$ &            $ 25.32 \pm   4.99$ &            $  -303 \pm     11$ &    { $\bm{ 21.32 \pm   4.99}$} &                       $ -0.33$  \\ \hline
                          5891 &            $ 22.35 \pm   1.93$ &            $  -198 \pm      9$ &            $ 30.59 \pm   2.11$ &            $  -223 \pm     22$ &    { $\bm{ 19.89 \pm   0.65}$} &   {  $\bm{  -237 \pm     24}$} &            $ 20.16 \pm   1.14$ &                       $  0.55$  \\ \hline
                          5906 &    { $\bm{ 42.14 \pm  12.98}$} &            $  -273 \pm      4$ &            $123.79 \pm  13.16$ &            $  -275 \pm      7$ &            $ 74.78 \pm   5.79$ &   {  $\bm{  -285 \pm      2}$} &            $ 82.23 \pm  10.43$ &                       $  0.38$  \\ \hline
                          5965 &    { $\bm{ 28.43 \pm   3.06}$} &   {  $\bm{  -387 \pm      4}$} &            $ 34.27 \pm   1.03$ &            $  -298 \pm     14$ &            $ 44.31 \pm   2.20$ &            $  -300 \pm     40$ &            $ 42.20 \pm   2.16$ &                       $  0.66$  \\ \hline
                          5970 &            $  3.05 \pm   1.75$ &            $  -124 \pm      3$ &            $ 17.48 \pm   8.12$ &            $  -157 \pm      7$ &            $  4.63 \pm   3.23$ &   {  $\bm{  -272 \pm     76}$} &    { $\bm{  1.17 \pm   0.31}$} &                       $  0.53$  \\ \hline
                          5971 &    { $\bm{  3.15 \pm   1.28}$} &            $  -318 \pm      3$ &            $  4.75 \pm   1.27$ &            $  -321 \pm      5$ &            $  4.34 \pm   0.51$ &   {  $\bm{  -322 \pm      4}$} &            $  3.85 \pm   0.58$ &                       $ -0.03$  \\ \hline
                          6766 &            $ 72.09 \pm   5.79$ &            $  -121 \pm      2$ &            $ 75.96 \pm   8.89$ &            $  -159 \pm     11$ &            $ 60.45 \pm   9.66$ &   {  $\bm{  -254 \pm     77}$} &    { $\bm{ 55.15 \pm   3.50}$} &                       $  0.52$  \\ \hline
                          6767 &    { $\bm{ 12.27 \pm   1.77}$} &            $  -317 \pm      2$ &            $ 26.33 \pm   3.55$ &            $  -308 \pm     13$ &            $ 32.00 \pm   5.01$ &   {  $\bm{  -326 \pm      7}$} &            $ 25.74 \pm   7.65$ &                       $  0.43$  \\ \hline
                          6794 &    { $\bm{ 31.02 \pm   4.73}$} &   {  $\bm{  -449 \pm      3}$} &            $ 33.05 \pm   1.29$ &            $  -428 \pm      9$ &            $ 33.91 \pm   1.05$ &            $  -420 \pm     29$ &            $ 36.36 \pm   1.61$ &                       $  0.33$  \\ \hline
                          7607 &    { $\bm{ 17.00 \pm   3.04}$} &   {  $\bm{  -431 \pm      4}$} &            $ 21.41 \pm   2.75$ &            $  -315 \pm     30$ &            $ 33.99 \pm   4.24$ &            $  -337 \pm     27$ &            $ 33.99 \pm   1.90$ &                       $  0.69$  \\ \hline
                          7609 &    { $\bm{  4.13 \pm   0.62}$} &   {  $\bm{  -425 \pm     10}$} &            $  5.61 \pm   0.97$ &            $  -304 \pm     30$ &            $  7.21 \pm   1.02$ &            $  -332 \pm     18$ &            $  7.77 \pm   0.70$ &                       $  0.50$  \\ 
\bottomrule
\end{tabular}
\caption{The mean and standard deviation of Training NLLs (rounded to integers) and simple regrets ($\times 10^{-3}$) for each search space in HPO-B. We highlight the best Training NLL and simple regret in bold. Lower Training NLLs does not necessarily lead to lower regrets due to ``negative transfer''.}
\label{tab:nll_regret_hpob_n_remain=100}
\end{table}
\begin{figure}[t]
    \centering
    \includegraphics[width=1.\textwidth]{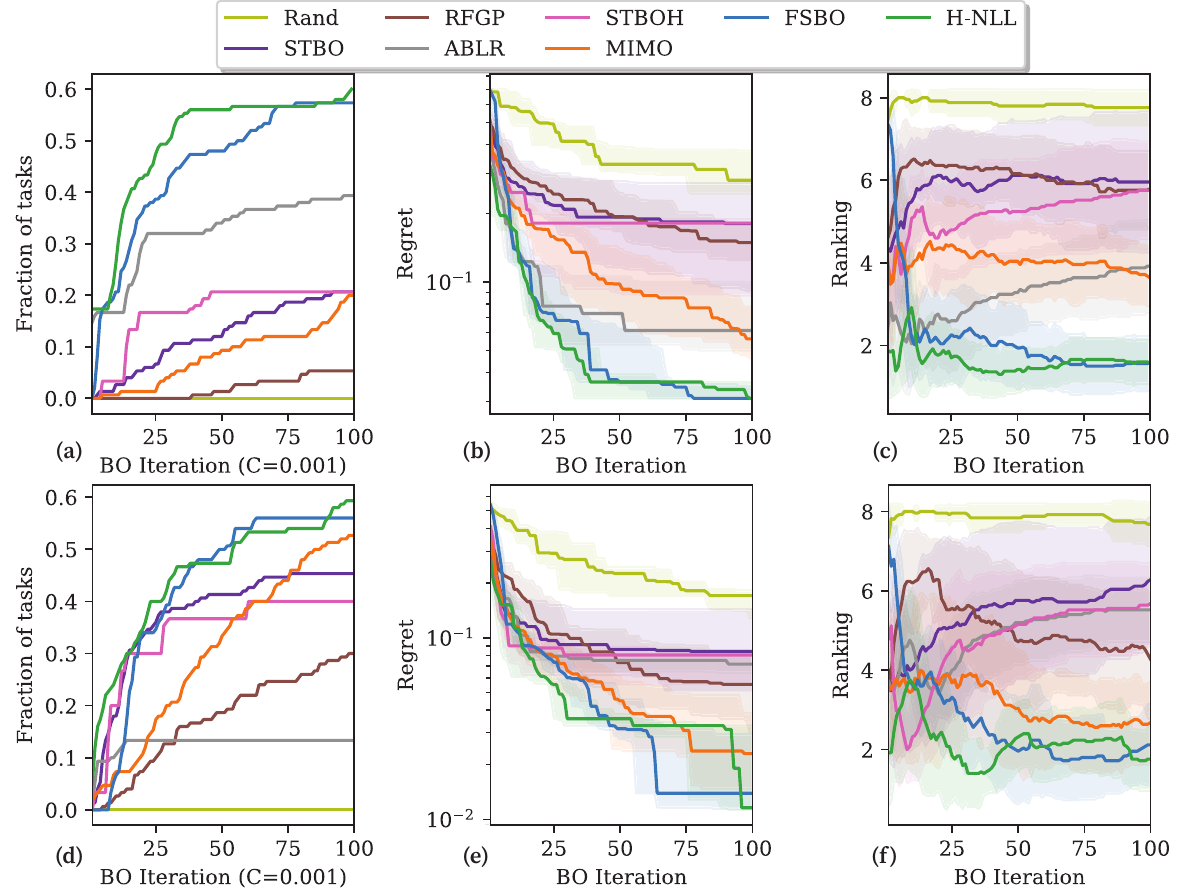}
    \caption{(a-c) are the performance profiles, regrets and ranking plots for search space 5636, and (d-f) are those for search space 5859. These search spaces contain flat training functions that encourage the ``negative transfer'' effect.}
    \label{fig:hpob-regret-5636}
\end{figure}

We adopted the setup of \S\ref{ssec:hpob_reduced} and pre-trained on at most 100 randomly sampled datapoints per training sub-dataset. Table~\ref{tab:nll_regret_hpob_n_remain=100} includes simple regrets at the $100$th BO iteration for MIMO and \hyperbo methods, as well as Training NLLs of \hyperbo methods. We did not compute Test NLLs due to the large sizes of some test sub-datasets. While most HPO-B search spaces show positive correlations between the Training NLL and the regret, some search spaces, including 5636 and 5859, have negative Pearson correlations between Training NLLs and regrets. Within search spaces 5636, there are in fact 2 training functions (out of 54 in total) that are flat; i.e., all inputs have the same function values. For search space 5859, there are 4 flat training functions (out of 56 in total). In both of these two search spaces, none of the test functions are flat, suggesting that the flat training functions may negatively transfer the knowledge to test functions.

Figure~\ref{fig:hpob-regret-5636} shows the aggregated results from search space 5636 and search space 5859. While the ``negative transfer'' effects are present in those two search spaces, H-NLL was still able to perform competitively and effectively lower the regrets for the test tasks in both search spaces.

\subsection{Summary of Experiments}
\label{ssec:exp_summary}
Our experiments on PD1 and HPO-B showed the superior performance of \hyperbo methods measured by simple regrets, fractions of succeeded tasks and rankings. H-NLL and H-EKL used the NLL and EKL loss functions defined in this work and overall performed better than competitive baselines. Our empirical analyses for PD1 showed evidence on the positive correlations between the NLL, EKL training losses, the NLL on test tasks and performance expressed by simple regrets. We studied the impact of reduced amount of training data and we also investigated the ``negative transfer'' problem in HPO-B. We found that \hyperbo methods benefit from more training tasks and datapoints per task, and they are also robust to ``negative transfer''. Our results confirmed that pre-trained GPs can better describe the test functions and as a result, obtain better BO performance.

\section{Discussion}
\label{sec:discuss}
Like most of the optimization literature, it is difficult to claim one approach is always better than the other. This phenomenon is widely recognized: “no free lunch in optimization”. In this work, however, we showed that if we pre-train a Gaussian process (GP) and use it as the prior for Bayesian optimization (BO), we can escape “no free lunch” for a distribution of optimization tasks. 

In the following, we discuss some interesting aspects of \hyperbo. In \S\ref{ssec:fully_bayes}, we present interpretations of \hyperbo in a fully Bayesian manner. In \S\ref{ssec:discuss_extensions}, we identify potential extensions of \hyperbo to enable more practical use cases. And finally in \S\ref{ssec:discuss_open_problems}, we discuss unsolved open problems.

\subsection{Fully Bayesian Interpretations of \hyperbo}\label{ssec:fully_bayes}
In \S\ref{sec:pf}, we assume the training functions $f_1, \cdots, f_N$ and the test function $f$ are independently drawn from the same GP. %
This assumption is consistent with hierarchical Bayes settings, where all functions are independent conditioned on the GP. To construct fully Bayesian interpretations of \hyperbo, one can place priors on the GP so that it is possible to perform complete Bayesian posterior updates on both the GP and the functions when we have more observed function values. There are different ways one can construct Bayesian hierarchical models for GPs. Here we focus on a generic hierarchical model where priors are placed on both the observation noise variance and the parameters of the mean and kernel functions.%

More specifically, we assume that there is a parameter $\theta \sim p(\theta; \alpha)$; a mean function $\mu$, a kernel function $k$ and a noise variance parameter $\sigma^2$ are drawn from $p(\mu, k, \sigma^2\mid \theta)$. Some functions $\{f_i\}_{i \in [N]}$ are then sampled independently from the same $\GP(\mu, k)$. The generative story of this hierarchical model is as follows:
\begin{itemize}
    \item Draw parameter $\theta$ from $p(\theta; \alpha)$.
    \item Draw mean function $\mu$, kernel function $k$ and noise variance $\sigma^2$ from $p(\mu, k, \sigma^2 \mid \theta)$.
    \item For each function index $i$ from $1$ to $N$,
    \begin{itemize}
        \item Draw a function $f_i$ from $\GP(\mu, k)$.
        \item For each datapoint index $j$ from $1$ to $M_i$,
        \begin{itemize}
            \item Given input $x^{(i)}_{j}$, we draw the observation $y^{(i)}_{j}\sim \N\left(f_i(x^{(i)}_{j}), \sigma^2\right)$.
        \end{itemize}
    \end{itemize}
\end{itemize}

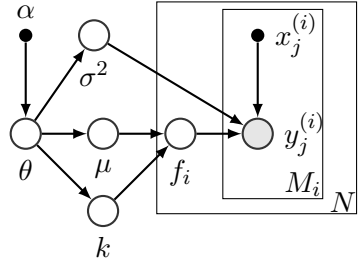
\begin{wrapfigure}{r}{0.5\textwidth}
\centering
\begin{tikzpicture}
\tikzstyle{main}=[circle, minimum size = 4mm, thick, draw =black!80, node distance = 6mm]
\tikzstyle{para}=[circle, minimum size = 5pt, inner sep=0pt]
\tikzstyle{connect}=[-latex, thick]
\tikzstyle{box}=[rectangle, draw=black!100]

  \node[main, fill = white!100] (theta) [label=below:$\theta$] { };
  \node[main] (mu) [right=of theta,label=below:$\mu$] {};
  \node[main] (f) [right=of mu,label=below:$f_i$] { };
  \node[main] (k) [below=of mu,label=below:$k$] { };
   \node[para, fill = black!100] (alpha) [above=of theta, label=above:$\alpha$] { };
   \node[main, fill = white!100] (sigma) [right=of alpha, label=below:$\sigma^2$] { };
  \node[main, fill = black!10] (y) [right=of f,label=right:$y^{(i)}_{j}$] { };
  \node[para, fill = black!100] (x) [above=of y,label=right:$x^{(i)}_{j}$] { };

  \path (alpha) edge [connect] (theta)
        (theta) edge [connect] (sigma)
        (sigma) edge [connect] (y)
        (theta) edge [connect] (mu)
        (theta) edge [connect] (k)
		(mu) edge [connect] (f)
		(k) edge [connect] (f)
		(f) edge [connect] (y)
		(x) edge [connect] (y);
  \node[rectangle, inner sep=0mm, fit= (x) (y), label=below right:$M_i$, xshift=0mm, yshift=-7.5mm] {};
  \node[rectangle, inner sep=4.5mm,draw=black!100, fit= (x) (y), xshift=2mm, yshift=-2mm] {};
  \node[rectangle, inner sep=0mm, fit= (x) (y) (f), label=below right:$N$, xshift=6mm, yshift=-5mm] {};
  \node[rectangle, inner sep=6mm,draw=black!100, fit= (x) (y) (f), yshift=-2.5mm, xshift=5mm] {};
\end{tikzpicture}
\caption{Graphical model for a hierarchical GP.}
\label{fig:fully_bayes}
\end{wrapfigure}

Figure~\ref{fig:fully_bayes} illustrates the graphical model for the hierarchical GP that we described above.
We can simplify this hierarchical setting by restricting $p(\mu, k, \sigma^2 \mid \theta)$ to be Dirac delta functions: mean function $\mu$, kernel function $k$  and noise variance $\sigma^2$ are all fully determined given the parameter $\theta$. This also means that the only free parameter in our model is $\theta$. Now, given the training dataset $D_N = \{D_{f_i}\}_{i=1}^N$ and a sub-dataset $D_f$ (the set of observations on the test function $f$), we can infer the posterior of the parameter $\theta$: $p(\theta \mid D_N, D_f;\alpha)$. Then, we can also get the posterior for the test function $f$ by marginalizing out the parameters
\begin{align}
\label{eq:f_posterior_full_bayesian}
    p(f\mid D_N, D_f; \alpha) &= \int_{\theta} p(f\mid \theta)p(\theta \mid D_N, D_f;\alpha).
\end{align}
In other words, the posterior of function $f$ depends on observations from all other conditionally \textit{i.i.d.} function samples $f_1, \cdots, f_N$. So at each iteration of BO, we need to solve Eq.~\ref{eq:f_posterior_full_bayesian}, which typically involves approximation and can have much worse time complexity (e.g., we may have to sample from the posterior over $\theta$ whenever we have new observations on $f$). %

The solution we had in \S\ref{ssec:objective_overview} can be interpreted as a cheap empirical Bayes approximation of Eq.~\ref{eq:f_posterior_full_bayesian}: obtaining a point estimate for $\theta$ via MLE on training dataset $D_N$, and approximate the full posterior $p(f \mid  D_N, D_f; \alpha)$ by removing $D_N$ in the condition, i.e., $p(f \mid  D_f, \theta)$. 

Overall, it is indeed possible to interpret \hyperbo in a fully Bayesian manner, but we found it not very meaningful as it does not provide new insights on methodology and brings us back to the original problem we identified: how to remove the known prior assumption in BO.

\subsection{Important Extensions of \hyperbo to Address Real-world Challenges of BO}
\label{ssec:discuss_extensions}
In this work, we focused on the question of how to efficiently and effectively make use of multi-task training data to enable better BO via pre-training. Here we discuss extensions to our work that would enable even more flexible uses.

\vspace{.5em}
\underline{\it Batch evaluation.}\;
For simplicity of this paper, we did not consider batch evaluation but rather only focused on the prior selection dimension of the challenges in BO. However, it is straightforward to adopt any batch BO methods in conjunction with \hyperbo to support obtaining observations in parallel. For example, we can directly use batch methods from \citet{snoek2012practical, kathuria2016batched, wang2017batched} etc. to replace line~\ref{alg:strategy} of Algorithm~\ref{alg:hyperbo}.

\vspace{.5em}
\underline{\it High-dimensional and large scale data.}\; Our hyperparameter tuning tasks are relatively low-dimensional (2-18 dimensions) and small scale BO problems. Because the tuning objectives are typically very expensive, we set the maximum number of BO observations to be $100$. Other applications of BO might have high-dimensional inputs, and the number of observations for each BO problem might be much higher. 

Our method can potentially be combined with high-dimensional and large scale BO methods to offer more capabilities. Those methods typically adopt probabilistic models different from vanilla GPs. In line~\ref{alg:train} of Algorithm~\ref{alg:hyperbo}, we can adapt \hyperbo to optimize the objective function in Eq.~\ref{eq:lossobj} for the new model. Our meta-learning idea in this paper in fact also brings benefit to high-dimensional and large scale BO methods so that they can better identify their critical special structures, e.g., low-dimensional embeddings~\citep{wang2016bayesian}, cylindrical kernels~\citep{oh2018bock} or additive Mondrian kernels~\citep{wang2018batched}.

\vspace{.5em}
\underline{\it Different search spaces.}\;
Roughly speaking, there could be two circumstances for different search spaces. Case I is that tasks share the same search variables, but the search ranges for some variables are different. For example, we may have each function $f_i:\mathfrak X_i \rightarrow \R, i \in [N]$ and $\mathfrak X_i = \prod_{j=1}^d[l_{ij}, h_{ij}] \subset \R^d$. In this case, our solution still applies by simply setting a union search space as $\mathfrak X = \bigcup_{i=1}^N \mathfrak X_i$ for learning and use the designated search space of new tasks for optimization.

Case II is more complicated: the search space for each function $f_i$ is $\mathfrak X_i \subset \R^{d_i}$ and each dimension of $\mathfrak X_i$ may have a different meaning than another search space $\mathfrak X_j$ ($i\neq j$). This paper does not have an answer for this scenario, but concurrent works by \cite{fan2022hyperbo, fan2023transfer} provided some interesting solutions based on the \hyperbo framework. Their key ideas are to use different setups of hierarchical GPs with stationary kernels and learn domain-specific priors or a universal prior for the shared parameters (e.g., lengthscales). Another concurrent work by \cite{optformer} solves Case II by pre-training a transformer to predict the next datapoints given a sequence of observations. This approach, however, may require millions of BO trajectories for training. %

\subsection{Other Open Problems}\label{ssec:discuss_open_problems}
Along the way of developing \hyperbo in this work, our findings and understandings have gradually revealed interesting open problems that are fundamental in hindsight but seem not to be well studied in the field.  We list some of these problems below.

\vspace{.5em}
\underline{\it Bounded posterior predictions.}\; Even if the pre-trained GP is very close to the ground truth GP, the posterior conditioned on the pre-trained GP may in fact deviate more and more from the ground truth posterior as we make more observations. Theorem~\ref{thm:posterior_bound} characterized this effect and suggested rescaling posterior for unbiased predictions of uncertainty. More studies are needed to better understand the assumptions and implications of Theorem~\ref{thm:posterior_bound}. Can we use a different estimator to make the bounds tighter? Another potential question to ask is, if the loss in Eq.~\ref{eq:loss} (or the approximations in Eq.~\ref{eq:kl} or Eq.~\ref{eq:nll}) is bounded, how does that bound translate to bounds on the posterior predictions of pre-trained GPs?

\vspace{.5em}
\underline{\it The necessary conditions for bounded regrets.}\;
As mentioned in \S\ref{ssec:post_acfun}, Eq.~\ref{eq:open_question_posterior} (matching pre-trained posteriors with ground truth posteriors) and Eq.~\ref{eq:open_question_acfun} (matching acquisition strategies) are sufficient but not necessary conditions to have bounded regrets. With Assumption~\ref{asp:finite} in \S\ref{ssec:theory}, we demonstrated that we can show a near-zero regret bound as long as we can bound the differences of posterior mean and variance predictions (Theorem~\ref{thm:posterior_bound}), which is another sufficient but not necessary condition. It is interesting to consider what the necessary conditions are for bounded regrets in \hyperbo. If we know these necessary conditions, we will be able to know when \hyperbo will fail.

\vspace{.5em}
\underline{\it Regret bounds for more practical settings.}\;
 In practice, users of \hyperbo might prefer to pre-train GPs with complex kernel and mean functions on continuous domains. However, in this paper, the regret bounds in Theorem~\ref{thm:regret} requires Assumption~\ref{asp:finite}, which restricts the mean and kernels to be described by finite dimensional vectors and matrices. It remains unclear how to properly bound the regrets if Assumption~\ref{asp:finite} does not hold. \cite{wang2018regret} had a partial solution where we can relax Assumption~\ref{asp:finite} to have compact domain, but it required linear kernels and exact solutions to linear systems, which is still not practical for real world settings. To fill the gap between theory and practice, we might need more guarantees on the optimization of the pre-training objectives.

As a side note, it is important not to forget that both the regret bounds and empirical studies serve the goal of verifying BO methods and identifying obvious pitfalls, so that we know how to use and adapt these methods in practice. 

\vspace{.5em}
\underline{\it Invalid evaluations in real-world BO problems.}\; As shown by our online experiments in Figure~\ref{fig:online} in \S\ref{ssec:online}, the results can be very different from the offline ones. This is because hyperparameter evaluations can suffer from various conditions and eventually return infeasible values. For example, the training curves may diverge, hardware may have failures and certain numbers may exceed what can be represented in a finite number of bits (e.g., 64 bits for float).  There is a rich literature on Bayesian optimization with constraints~\citep{snoek-2013, gelbart-2014a, gardner-2014, gelbart-2015, lobato-2016}.  However, many of the infeasible values we have observed do not fit well the assumptions made in this line of work. Specifically, the probability of receiving an infeasible value is not necessarily conditional on the inputs, or well modeled as a function from the inputs.  Instead, phenomena such as hardware failures can occur in unstructured or structured ways that are difficult to model, such as being the result of a faulty GPU in a cluster.  As such, this is perhaps better considered as a noise process rather than constraints on the input domain.%

This scenario is not specific to machine learning hyperparameter tuning. For example, for control parameter tuning in robotics, the evaluation of a parameter depends on downstream modules (e.g., planning or control), and if any of the downstream modules fail, we cannot get a feasible evaluation. In our experiments in \S\ref{ssec:online}, we mapped the infeasible evaluations to a negative value, but this is not accurately reflecting the true evaluations if, for example, the hardware failed.  
In order to achieve high sample efficiency in BO for real-world problems, we need a more principled solution to account for such infeasible evaluations. 

\vspace{.5em}
\underline{\it Pre-training other probabilistic models for function.}\; Our pre-training method described in \S\ref{ssec:objective_overview} is for GP priors, but it is not hard to see that any functional distribution could be pre-trained on multi-task data and the interpretation of the pre-trained prior is the distribution of the tasks. We may consider other kinds of prior models~\citep{roininen2016hyperpriors, tran2022all, gu2018robust,pearce2022bayesian} that are more flexible than GPs for more complex datasets, but the challenge is in approximating the divergence between the ground truth and the model.

\vspace{.5em}
\underline{\it Alternative types of divergences as pre-training objectives.}\; We used the KL divergence in our loss function in Eq.~\ref{eq:loss} of \S\ref{ssec:pre-train_obj} to describe the ``distance'' between two GPs. However, as pointed out by \cite{burt2020understanding} and \cite{sun2019functional}, KL divergences between GPs can be ill-defined for GPs with parametric models. So if the pre-trained GP assigns measure zero to functions representable by the ground truth, the KL divergence would be infinite.  There are other types of divergence or distance, e.g., Wasserstein distance, that may behave better, and there could be approximations that give posterior predictions that are more attuned to BO and other decision making tasks.

\section{Conclusion}
\label{sec:conclu}
We proposed \hyperbo: a framework that uses pre-training methods to learn a Gaussian process (GP) prior and subsequently perform Bayesian optimization (BO) with the learned prior. We developed KL divergence based pre-training methods and analyzed in detail the implications of using pre-trained GPs in BO. With \hyperbo, we no longer have to hand-specify the exact quantitative parameters in a GP prior. Instead, we only need to identify related tasks and past observations so that they can be used for pre-training. 

Theoretically, we showed the regret bounds of \hyperbo without the assumption that the prior is known. To verify \hyperbo on real-world expensive hyperparameter tuning problems, we collected the PD1 Neural Net Tuning Dataset: the first hyperparameter tuning dataset for modern large-scale neural network model training. We demonstrated the superior empirical performance of \hyperbo on both PD1 and other transfer learning BO benchmarks. Importantly, the idea of pre-training GPs brings a new perspective to interpret epistemic uncertainty in GPs, which are especially useful for Bayesian sequential decision making tasks exemplified by BO.  %

\acks{We would like to thank Zelda Mariet and Matthias Feurer for the help and consultation on transfer learning baselines. We would also like to thank Rif A. Saurous for constructive feedback, and Rodolphe Jenatton and David Belanger for feedback on previous versions of the manuscript. In addition, we thank Sharat Chikkerur, Ben Adlam, Balaji Lakshminarayanan, Fei Sha and Eytan Bakshy for comments.}

\newpage
\appendix
\section{Author Contributions}
In the following, we list key contributions by authors.

\begin{itemize}
    \item Zi Wang proposed the problem and designed the main \hyperbo framework; set vision for the overall story of this work; conducted theoretical analyses; coordinated the design and completion of experiments, infrastructure and the new PD1 dataset; implemented the majority of \hyperbo; conducted the experiments on \hyperbo and some other baselines; consolidated discussions with co-authors and wrote the majority of the paper.
    \item George E. Dahl led the effort to create the PD1 Neural Net Tuning Dataset and co-led the init2winit project (infrastructure for PD1); managed all data collection experiments, launch scripts and data loading scripts with analyses; proposed to use performance profiles for experiment analyses; wrote \S\ref{ssec:data}; provided critical consultation on deep learning training algorithms and experiment design.
    \item Kevin Swersky refined \hyperbo by adding implementations of input warping and prior regularizers; named \hyperbo; fixed critical bugs of gradient computations in \hyperbo; refined \S\ref{sec:explain} to explain the principles behind \hyperbo; significantly contributed to animated visualizations; contributed to the final revision and re-pivoting of this work.
    \item Chansoo Lee provided substantial support on the integration of \hyperbo to internal toolboxes for online experiments; added essential internal baselines (STBOH, MIMO and RFGP) to experiments and wrote the descriptions and analyses; implemented the majority of GPax.
    \item Zachary Nado co-led the init2winit project (infrastructure for PD1) and designed new tasks for data collection; provided timely support on online experiments and integrated \hyperbo into init2winit; provided consultation on deep learning training algorithms.
    \item Justin Gilmer co-led the init2winit project (infrastructure for PD1) and designed new tasks for data collection; supported internal infrastructure; provided consultation on deep learning training algorithms.
    \item Jasper Snoek advised on project direction, modeling choices, coding framework and paper writing; refined \hyperbo by adding implementations of slice sampling, L-BFGS and warping priors; significantly contributed to code review; contributed to the final revision and re-pivoting of this work.
    \item Zoubin Ghahramani advised on project direction, philosophy, narratives and paper writing.
\end{itemize}

\section{Derivations for EKL}
\label{app:equations}

In this section, we detail the derivations of EKL in \S\ref{sssec:kl-approx} and its equivalence to NLL in special cases. To simplify the notations, we use $p$ to denote the PDF of the ground truth multivariate Gaussian distribution, $\tilde p$ to denote the PDF of the MLE estimate with mean $\tilde \vmu\in\R^{r}$ and covariance $\tilde\Sigma\in\R^{r\times r}$, and $q$ to denote the PDF of the model with mean $\vmu\in\R^{r}$ and covariance $\Sigma\in\R^{r\times r}$.

\vspace{.5em}
\underline{\it The KL divergence between two multivariate Gaussians.}\;
Assume that $\tilde p$ and $q$ are non-degenerate. By definition, the KL divergence between $\tilde p$ and $q$ is
\begin{align*}
    D_{\KL}\Bigl(\tilde p, q\Bigl) &= H(\tilde p, q) - H(\tilde p) \\
    &= \frac12 \Ex_{\vy \sim \tilde p}
    \left[
        (\vy - \vmu)^\top \Sigma^{-1} (\vy - \vmu) + \log|\Sigma| + r \log(2\pi) 
    \right] - \frac{1}{2} \left(r\log (2\pi) + \log |\tilde\Sigma| + r\right) \\
    &= \frac12 \left(\Ex_{\vy \sim \tilde p}
    \left[
        (\vy - \vmu)^\top \Sigma^{-1} (\vy - \vmu)\right] + \log \frac{|\Sigma|}{|\tilde \Sigma|} - r \right).
\end{align*}

The expectation can be derived as follows,
\begin{align*}
\Ex_{\vy \sim \tilde p}
    \left[
        (\vy - \vmu)^\top \Sigma^{-1} (\vy - \vmu)\right]
&= 
     \tr\left(\Sigma^{-1}\Ex_{\vy \sim \tilde p}\left[(\vy - \vmu) (\vy - \vmu)^\top\right]\right)
\\ &= 
    \tr\left(\Sigma^{-1}\left(\Ex_{\vy \sim \tilde p}\left[(\vy - \tilde\vmu ) (\vy -\tilde\vmu )^\top\right]
    + (\tilde\vmu - \vmu) (\tilde\vmu - \vmu)^\top\right)\right)
\\ &= 
    \tr\left(\Sigma^{-1}\tilde\Sigma\right)
    + (\tilde{\vmu} - \vmu)^\top \Sigma^{-1} (\tilde{\vmu} - \vmu).
\end{align*}
The  derivation above uses the definition of the mean and covariance of Gaussian distributions, i.e., for $\tilde p$, its mean and covariance are defined as $\tilde\vmu = \Ex_{\vy \sim \tilde p}[\vy]$, $\tilde\Sigma = \Ex_{\vy \sim \tilde p}\left[(\vy - \tilde\vmu ) (\vy - \tilde\vmu )^\top\right]$.

Hence, we have
\begin{align} \label{eq:app_kl}
    D_{\KL}\Bigl(\tilde p, q\Bigl) = \frac12 \left(\tr(\Sigma^{-1}\tilde\Sigma)
    + (\tilde{\vmu} - \vmu)^\top \Sigma^{-1} (\tilde{\vmu} - \vmu) + \log \frac{|\Sigma|}{|\tilde \Sigma|} - r \right).
\end{align}
Eq.~\ref{eq:app_kl} can be used to derive Eq.~\ref{eq:degenerate} and it is consistent with Eq.~\ref{eq:nondegenerate}.

\vspace{.5em}
\underline{\it The equivalence between EKL and NLL in special cases.}\;
Using the simplified notations, the claim in \S\ref{sssec:nll_ekl_equiv} is
\begin{align*}
    D_{\KL}\Bigl(\tilde p, q\Bigl) \equiv - \frac{1}{N} \sum_{i=1}^N \log  q(\vy_i) + \frac{1}{N} \sum_{i=1}^N \log \tilde p(\vy_i),
\end{align*}
where $\vy_i$ are \iid samples from the ground truth $p$.

\vspace{1em}
We prove the claim below.

By definition of the PDF of multivariate Gaussians, we have
\begin{align*}
    - \frac{1}{N}\sum_{i=1}^{N} \log \tilde p\left(\vy_i\right) &= \frac{1}{2N} \sum_{i=1}^{N} \left(
        (\vy_i - \tilde\vmu)^\top \tilde\Sigma^{-1} (\vy_i - \tilde\vmu) + \log|\tilde\Sigma| + r \log(2\pi) \right) \\
        &= \frac12\left(r\log (2\pi) + \log |\tilde\Sigma| \right) + \frac{1}{2N} \sum_{i=1}^{N}
        (\vy_i - \tilde\vmu)^\top \tilde\Sigma^{-1} (\vy_i - \tilde\vmu) \\
        &= \frac12\left(r\log (2\pi) + \log |\tilde\Sigma| \right) + \frac{1}{2} \tr\left(\tilde\Sigma^{-1} \tilde\Sigma \right)\\
        & = \frac{1}{2} \left(r\log (2\pi) + \log |\tilde\Sigma| + r \right).
\end{align*}
The derivation uses the construction of MLE estimate for covariance matrix: $$N\tilde\Sigma = \sum_{i=1}^{N} \left(\vy_i - \tilde{\vmu}\right)
\left(\vy_i - \tilde{\vmu}\right)^\top.$$ 
This means, $H(\tilde p) \equiv - \frac{1}{N}\sum_{i=1}^{N} \log \tilde p\left(\vy_i\right)$.

Moreover, we have
\begin{align*}
    - \frac{1}{N}\sum_{i=1}^{N} \log q\left(\vy_i\right) & =  \frac{1}{2N}\sum_{i=1}^{N} \left(
        (\vy_i - \vmu)^\top \Sigma^{-1} (\vy_i - \vmu) + \log|\Sigma| + r \log(2\pi) \right)\\
        &= \frac12\left(r\log (2\pi) + \log |\Sigma| \right) + \frac{1}{2N}\sum_{i=1}^{N} (\vy_i - \vmu)^\top \Sigma^{-1} (\vy_i - \vmu),
\end{align*}
and
\begin{align*}
  \frac{1}{N}\sum_{i=1}^{N} (\vy_i - \vmu)^\top \Sigma^{-1} (\vy_i - \vmu) & =  \tr\left(\Sigma^{-1} \frac1s \sum_{i=1}^{N} (\vy_i - \vmu) (\vy_i - \vmu)^\top \right)\\
    &= \tr\left(\Sigma^{-1}\left(\frac{1}{N} \sum_{i=1}^{N} \left(\vy_i - \tilde{\vmu}\right)
\left(\vy_i - \tilde{\vmu}\right)^\top
    + (\tilde\vmu - \vmu) (\tilde\vmu - \vmu)^\top\right)\right) \\
    & = \tr\left(\Sigma^{-1}\tilde\Sigma\right)
    + (\tilde{\vmu} - \vmu)^\top \Sigma^{-1} (\tilde{\vmu} - \vmu).
\end{align*}
The  derivation above for $RHS$ uses the construction of MLE estimates for $\tilde\vmu$ and $\tilde\Sigma$  in Eq.~\ref{eq:estimate_mean_cov}, i.e.,
\begin{align*}
&\tilde{\vmu} = \frac{1}{N}\sum_{i=1}^{N} \vy_i,
&
\frac{1}{N}\sum_{i=1}^{N} \vy_i\tilde{\vmu}^\top = \tilde{\vmu}\tilde{\vmu}^\top,
&
&\tilde\Sigma = \frac{1}{N} \sum_{i=1}^{N} \left(\vy_i - \tilde{\vmu}\right)
\left(\vy_i - \tilde{\vmu}\right)^\top.
\end{align*}
By combining the equations above, we have 
\begin{align*}
    - \frac{1}{N}\sum_{i=1}^{N} \log q\left(\vy_i\right) + \frac{1}{N}\sum_{i=1}^{N} \log \tilde p\left(\vy_i\right) = \frac12 \left(\tr(\Sigma^{-1}\tilde\Sigma)
    + (\tilde{\vmu} - \vmu)^\top \Sigma^{-1} (\tilde{\vmu} - \vmu) + \log \frac{|\Sigma|}{|\tilde \Sigma|} - r \right),
\end{align*}
which is equal to Eq.~\ref{eq:app_kl}.
Hence, the claim holds.

\section{Proofs for Theorem~\ref{thm:posterior_bound} and Theorem~\ref{thm:regret}}
\label{app:regret}

We provide proofs for Theorem~\ref{thm:posterior_bound}, which bounds the posterior predictions of pre-trained Gaussian processes (GPs) and Theorem~\ref{thm:regret}, which bounds the regrets of HyperBO. We also show an alternative regret bound (Theorem~\ref{app_thm:regret_ucb_alternative}) in this section. 
\subsection{Theorem~\ref{thm:posterior_bound}: Bounding Pre-trained GP Posterior Predictions}
\label{sec:bound}

Recall that in Eq.~\ref{eq:estimate_mean_cov}, we used MLE to estimate the mean vector and covariance matrix of the ground truth GP on finite inputs. Proposition~\ref{prop:gpexist} then showed that the pre-trained GP on finite inputs is exactly the same as Eq.~\ref{eq:estimate_mean_cov}. So if we can bound posteriors based on  Eq.~\ref{eq:estimate_mean_cov}, we can show Theorem~\ref{thm:posterior_bound}. 

Our plan for proving Theorem~\ref{thm:posterior_bound} is as follows. We first use the MLE estimates for a multivariate Gaussian distribution to derive a conditional distribution, and the parameters of that MLE based conditional distribution can be bounded by the parameters of the ground truth conditional distribution. This is done using the probability density functions for the MLE estimates of the mean vector and covariance matrix. Then, we can plug in our GP setups described in \S\ref{ssec:finite_hyperbo} and obtain Theorem~\ref{thm:posterior_bound}. 

The general proof strategy follows our prior work~\citep{wang2018regret}, but in this work, we significantly simplified the proofs, improved readability and fixed issues with coefficients.

\begin{assumption} \label{asmp:app_define_mlp}
$\vy_1, \cdots, \vy_N \in \R^{t}$ are independent and identically distributed according to $\mathcal N(\vu, V)$, $V$ is positive definite and $N > t$. Let $Y=[\vy_i]_{i=1}^N \in\R^{t\times N}$. We denote the MLE estimators of $\vu, V$ as $\hat \vu = \frac1N Y 1_N \in\R^t$ and $\hat V = \frac{1}{N} (Y - \hat \vu 1_N\T)(Y -  \hat \vu 1_N\T)\T \in \R^{t\times t}$ where $1_N$ is a column vector of size $N$ filled with 1s.
\end{assumption}

\begin{lem} \label{lem:estimate0} Given Assumption~\ref{asmp:app_define_mlp}, $\hat u$ and $\hat V$ are independent, and
\begin{align*}
&\hat \vu \sim \mathcal N(\vu, \frac{1}{N}V), \; \;\hat V \sim \mathcal W(\frac{1}{N}V, N-1).
\end{align*}
\end{lem}
Lemma~\ref{lem:estimate0} is a combination of Theorem 3.3.2 and Corollary 7.2.3 of~\cite{anderson1958introduction}. Interested readers can find the proof of  Lemma~\ref{lem:estimate0} in~\cite{anderson1958introduction}.

\begin{lem}\label{lem:wishart1}For any $\hat v\sim \mathcal W(v, n), v\in \R$ and $b>0$, we have 
\begin{align*}
&\Pr[\frac{\hat v}{vn} \geq 1+ 2\sqrt{b} + 2b] \leq e^{-bn}, \;\;\Pr[\frac{\hat v}{vn} \leq 1-2\sqrt{b } ] \leq e^{-bn}.
\end{align*}
\end{lem}
\begin{proof}
Given $\hat v\sim \mathcal W(v, n)$, $\frac{\hat v}{v}$ is distributed according to a chi-squared distribution with $n$ degrees of freedom; namely, $\frac{\hat v}{v}\sim\chi^2(n)$. By Lemma 1 in~\cite{laurent2000adaptive}, we have
\begin{align*}
&\Pr[\frac{\hat v}{v} - n \geq 2\sqrt{na } + 2a] \leq e^{-a}, \;\;\Pr[\frac{\hat v}{v} - n \leq -2\sqrt{na } ] \leq e^{-a}.
\end{align*}
Let $a = bn$, and we can reorganize the inequalities above as
\begin{align*}
&\Pr[\frac{\hat v}{vn} \geq 1+ 2\sqrt{b} + 2b] \leq e^{-bn}, \;\;\Pr[\frac{\hat v}{vn} \leq 1-2\sqrt{b } ] \leq e^{-bn}.
\end{align*}
\end{proof}

\begin{lem}\label{lem:conditional_multivariate_gaussian}
Assume Assumption~\ref{asmp:app_define_mlp} and $N > t+1$. For any $\vy \sim \N(\vu, V)$, we partition $\vy = [y_\tau]_{\tau}^t \in\R^{t}$ to $\vy_{t-1} = [y_\tau]_{\tau}^{t-1}\in\R^{t-1}$ and $y_{t}\in\R$; i.e., $\vy=[\vy_{t-1}, y_t]$. In the same way, we partition $\vu$ and $\hat \vu$. 
For matrix $V = [v_{\tau, \tau'}]_{\tau,\tau'=1}^{t}$, we partition it to $V_{t-1} = [v_{\tau, \tau'}]_{\tau,\tau'=1}^{t-1}$, $V_{t-1, t} = V_{t, t-1}\T = [v_{\tau, t}]_{\tau}^{t}$ and $v_{t}$ as a short-hand of $v_{t,t}$. We perform the same partition to matrix $\hat V$. That is,
\begin{align*}
\vy=\begin{bmatrix*}
\vy_{t-1} \\ y_t
\end{bmatrix*},\quad
\vu=\begin{bmatrix*}
\vu_{t-1} \\ u_t
\end{bmatrix*},\quad
\hat \vu=\begin{bmatrix*}
\hat\vu_{t-1} \\ \hat u_t
\end{bmatrix*},\quad
    V = 
\begin{bmatrix*}
V_{t-1} &   V_{t-1, t}\\ 
V_{t, t-1} &     v_t
\end{bmatrix*}
,\quad
       \hat V = 
\begin{bmatrix*}
\hat V_{t-1} &  \hat V_{t-1, t}\\ 
\hat V_{t,t-1} &    \hat v_t
\end{bmatrix*}.
\end{align*}
We also define the following short-hands ,
\begin{align*}
& \omega = u_t + V_{t, t-1} V_{t-1}^{-1} (\vy_{t-1} - \vu_{t-1}), \;\;
\kappa = v_t - V_{t, t-1} V_{t-1}^{-1}  V_{t-1, t}, \\
&\hat\omega = \hat u_t + \hat V_{t, t-1} \hat V_{t-1}^{-1} (\vy_{t-1} - \hat\vu_{t-1}), \;\;
\hat\kappa = \hat v_t - \hat V_{t, t-1} \hat V_{t-1}^{-1}  \hat V_{t-1, t}, 
\end{align*}
where $y_t \mid \vy_{t-1} \sim \N(\omega, \kappa)$. Then, we have
\begin{align} \label{ex_omega_kappa}
& \Ex[\hat \omega] = \omega, \;\;
\Ex[\hat \kappa] = \frac{N-t}{N} \kappa. 
\end{align}
Pick $\delta_1\in(0,1)$ and $\delta_2\in(0,1)$,
\begin{align}\label{hatomega}
   \Pr[ |\hat \omega -\omega| < a \kappa ] \geq 1-\delta_1;\; a =  \sqrt{\frac{2t+4\sqrt{(t-1)\log{\frac{2}{\delta_1}}} + 4\log{\frac{2}{\delta_1}}  - 4/N}{(N-t-1)\delta_1}}, 
\end{align}
\begin{align}\label{hatkappa}
&\Pr[1-2\sqrt{b } < \frac{N \hat \kappa }{(N-t) \kappa} < 1+ 2\sqrt{b} + 2b] > 1 - \delta_2; \;\;  b=\frac{1}{N-t}\log\frac2{\delta_2}.
\end{align}
\end{lem}

\begin{proof} 
By Lemma~\ref{lem:estimate0} and Proposition 8.7 in~\cite{Eaton07}, we have 
\begin{align*}
\hat V \sim \mathcal W(\frac{1}{N}V, N-1), \; \; \hat \kappa \sim  \mathcal W(\frac{1}{N}\kappa, N-t).
\end{align*}
Hence, $\Ex[\hat \kappa] = \frac{N-t}{N} \kappa$. By Lemma~\ref{lem:wishart1}, we obtain
\begin{align*}
&\Pr[\frac{N \hat \kappa }{(N-t) \kappa} \geq 1+ 2\sqrt{b} + 2b] \leq e^{-b(N-t)} = \frac{\delta_2}{2}, \;\;\Pr[\frac{N \hat \kappa }{(N-t) \kappa} \leq 1-2\sqrt{b } ] \leq e^{-b(N-t)} =  \frac{\delta_2}{2},
\end{align*}
where $b=\frac{1}{N-t}\log\frac2{\delta_2}$. Thus,  Eq.~\ref{hatkappa} holds. Next, we prove the claims about $\hat \omega$. 

Our goal now is to bound the difference between $\hat \omega$ and $\omega$. The plan is to first re-write random matrix $\hat V$ as a multiplication of two random matrices, then show the conditional distributions of terms involved in $\hat \omega$, and finally derive the expectation and variance of $\hat \omega$.

By Definition~8.1 of~\cite{Eaton07}, there exist $X = [\vx_i]_{i=1}^{N-1} \in \R^{t\times (N-1)}, \vx_i  \overset{\iid}{\sim} \N(0, \frac1N V)$ such that $\hat V = X X\T$. Let $X = \big[\begin{smallmatrix}
X_2 \\ X_1
\end{smallmatrix}\big]$, where $X_1\in \R^{1\times (N-1)}$ (last row of $X$) and $X_2\in\R^{(t-1)\times (N-1)}$. We then have $\hat V_{t-1} = X_2 X_2\T$, $\hat V_{t, t-1} = X_1X_2\T$ and $\hat v_t = X_1 X_1\T$. 

Conditional on $X_2$, since $\vx_i  \overset{\iid}{\sim} \N(0, \frac1N V)$, elements of $X_1$ are independent, and each element has a Gaussian distribution conditioned on each column of $X_2$; that is, 
\begin{align*}
    X_{1} \mid X_2 \sim \N(V_{t, t-1}(V_{t-1})^{-1} X_2, I_{N-1} \otimes \frac1N\kappa).
\end{align*}

Hence, conditional on $\vy_{t-1}, \hat \vu_{t-1}$ and $X_2$, $\hat \omega - \hat u_t = X_1X_2\T (X_2X_2\T)^{-1}(\vy_{t-1} - \hat \vu_{t-1})$ is a weighted sum of \iid random variables (elements of $X_1$).

Let $\bar\vy = \vy_{t-1} - \hat \vu_{t-1}$. By applying the affine transformation $X_2\T (X_2X_2\T)^{-1}\bar\vy$ to $X_1$, we have 
\begin{align*}
    \hat \omega - \hat u_t \mid  \bar\vy, X_2 \sim \N\left(V_{t,t-1}V_{t-1}^{-1}\bar\vy, \;\frac{\kappa}{N}\bar\vy\T (X_2X_2\T)^{-1}\bar\vy\right).
\end{align*}

The conditional distribution for $ \hat \omega - \hat u_t$ depends on $X_2$ only through $\hat V_{t-1} = X_2X_2\T$. So we can write it as 
\begin{align*}
    \hat \omega - \hat u_t \mid  \bar\vy, \hat V_{t-1} \sim \N\left(V_{t,t-1}V_{t-1}^{-1}\bar\vy, \;\frac{\kappa}{N}\bar\vy\T \hat V_{t-1}^{-1}\bar\vy\right).
\end{align*}

Now we need to think about the distribution for $\hat u_t$. By Lemma~\ref{lem:estimate0}, $\hat\vu$ and $\hat V$ are independent. Hence $\hat u_t \perp \hat V_{t, t-1}$ and $\hat u_t \perp \hat V_{t-1}$. As a result, $\hat u_t \perp \hat\omega - \hat u_t \mid \hat\vu_{t-1}$. Since $\hat \vu \sim \mathcal N(\vu, \frac{1}{N}V)$, we have
\begin{align*}
    \hat u_t \mid \hat \vu_{t-1} \sim   \N\left(u_t + V_{t,t-1}V_{t-1}^{-1}(\hat \vu_{t-1} - \vu_{t-1})), \; \frac{\kappa}{N} \right).
\end{align*}

We add $\hat u_t$ back to $\hat \omega - \hat u_t$, and get
\begin{align*}
    \hat \omega \mid \vy_{t-1}, \hat \vu_{t-1}, \hat V_{t-1} \sim \N\left(\omega, \; \frac{\kappa}{N}(1+\bar\vy\T \hat V_{t-1}^{-1}\bar\vy) \right).
\end{align*}

By the law of total expectation, we can complete the proof for Eq.~\ref{ex_omega_kappa} as follows,
\begin{align*}
\Ex[\hat \omega \mid \vy_{t-1} ] &= \Ex\left[ \Ex[ \hat \omega \mid \vy_{t-1}, \hat \vu_{t-1}, \hat V_{t-1}]\mid \vy_{t-1}\right]  =  \omega.
\end{align*}

By the law of total conditional variance,
\begin{align*}
\Var[\hat \omega \mid \vy_{t-1}] &= \Ex\left[ \Var[ \hat \omega \mid \vy_{t-1}, \hat \vu_{t-1}, \hat V_{t-1}  ] \mid \vy_{t-1}\right] + \Var\left[\Ex[\hat \omega \mid \vy_{t-1}, \hat \vu_{t-1}, \hat V_{t-1}  ] \mid \vy_{t-1} \right] \\
& = \Ex\left[\frac{\kappa}{N}(1+\bar\vy\T \hat V_{t-1}^{-1}\bar\vy) \mid \vy_{t-1} \right]+ \Var\left[\omega \mid \vy_{t-1} \right] 
\end{align*}
Note that $\hat V_{t-1}^{-1}$ has an inverse Wishart distribution, i.e., $\hat V_{t-1}^{-1} \sim \mathcal W^{-1}(N V_{t-1}^{-1}, N-1)$; hence $\Ex[\hat V_{t-1}^{-1}] = \frac{N}{N-t-1} V_{t-1}^{-1}$. Since $\hat \vu_{t-1} \sim \mathcal N(\vu_{t-1}, \frac{1}{N}V_{t-1})$, we also have $N\bar \vu\T V_{t-1}^{-1}\bar \vu \sim \chi^2(t-1)$ where $\bar \vu = \hat \vu_{t-1} - \vu_{t-1}$ and $\chi^2(t-1)$ is the chi-squared distribution with $t-1$ degrees of freedom. Moreover, $\omega$ is a deterministic variable given $\vy_{t-1}$. Hence, %
\begin{align*}
\Var[\hat \omega \mid \vy_{t-1}] &= \frac{1 + q  - 2/N}{N-t-1} \kappa,
\end{align*}
where $q = (\vy_{t-1} - \vu_{t-1})\T V_{t-1}^{-1}(\vy_{t-1} - \vu_{t-1})$.

Now we apply Chebyshev's inequality, and obtain
\begin{align*}
&\Pr\left[ | \hat \omega - \omega | <  \sqrt{\frac{1 + q  - 2/N}{(N-t-1)\delta_1'} \kappa}  \mid \vy_{t-1}\right] \geq 1 - \delta_1'.
\end{align*}
Notice that the randomness of $q$ is from $\vy_{t-1}$ and $\vy_{t-1}\sim \mathcal N(\vu_{t-1}, V_{t-1})$. This means $q \sim \chi^2(t-1)$. So we can further bound $q \leq t-1+2\sqrt{(t-1)\log{\frac{1}{\delta_1''}}} + 2\log{\frac{1}{\delta_1''}}$ with probability at least $1-\delta_1''$ by Lemma 1 in~\cite{laurent2000adaptive}. Hence, we can set $\delta_1' = \frac{\delta_1}{2}$ and $\delta_1'' = \frac{\delta_1}{2}$, and have 
\begin{align*}
Pr\left[ | \hat \omega - \omega | <  a \sqrt{\kappa} \right] \geq 1 - \delta_1,
\end{align*}
where $a = \sqrt{\frac{2t+4\sqrt{(t-1)\log{\frac{2}{\delta_1}}} + 4\log{\frac{2}{\delta_1}}  - 4/N}{(N-t-1)\delta_1}}$. Thus, Eq.~\ref{hatomega} holds.
\end{proof}
\vspace{.5em}
\underline{\it Remarks.}\;
When bounding $\hat \omega$, we derived its mean and variance but did not try to characterize its distribution. In fact, we can write $\hat \omega = \omega_1 + \omega_2$, where
\begin{align*}
    & \omega_1 \sim \N\left(\omega, \; \frac{\kappa}{N} \right) \\
    & \omega_2 \sim \N\left(0, \; \frac{\kappa}{N}(\vy_{t-1} - \hat \vu_{t-1})\T \hat V_{t-1}^{-1}(\vy_{t-1} - \hat \vu_{t-1}) \right).
\end{align*}
Let $\omega_2 = \sqrt{\kappa}(\vy_{t-1} - \hat \vu_{t-1})\T \omega_3$, where $\hat \vu_{t-1} \sim \mathcal N(\vu_{t-1}, \frac{1}{N}V_{t-1})$ and $\omega_3\in\R^{t-1}$. Then we have
\begin{align*}
    & \omega_3 \sim \N\left(0, \; \frac{1}{N} \hat V_{t-1}^{-1}\right); \;\; \hat V_{t-1}^{-1} \sim \mathcal W^{-1}(N V_{t-1}^{-1}, N-1),
\end{align*}
which means $(\omega_3, \hat V_{t-1}^{-1})$ is distributed according to a normal-inverse-Wishart distribution. Interested readers can derive a tighter bound for $\hat \omega$ based on this decomposition.

\paragraph{Proof of Theorem~\ref{thm:posterior_bound}.}
The following proof is for Theorem~\ref{thm:posterior_bound} in \S\ref{ssec:analyses}.

\begin{proof} (Theorem~\ref{thm:posterior_bound})
    Recall that in the $t-$th iteration at line~\ref{alg:strategy} of Algorithm~\ref{alg:hyperbo}, we have observations $D_f = \{(x_\tau, y_\tau)\}_{\tau=1}^{t-1}$. Let $\vx = [x_\tau]_{\tau=1}^{t-1}$ and $\vy' = [y_{\tau}]_{\tau=1}^{t-1}$.
    
The ground truth posterior is \pgtgp$\vcentcolon=\GP(\mu^*_{t-1}, k^*_{t-1})$, where
\begin{align*}
\mu^*_{t-1}(x) &= \mu^*(x) + k^*(x,\vx)(k^*\circ{\sigma_*^2}(\vx))^{-1}(\vy' - \mu^*(\vx)), \;\;\forall x\in\mathfrak X,\\
k^*_{t-1}(x,x') &= k^*(x,x') - k^*(x, \vx)(k^*\circ{\sigma_*^2}(\vx))^{-1} k^*(\vx, x'), \;\;\forall x, x'\in\mathfrak X.
\end{align*}

The pre-trained GP posterior is \ppgp$\vcentcolon= \GP(\hat \mu_{t-1}, \hat k_{t-1})$, where
\begin{align*}
\hat \mu_{t-1}(x) &\vcentcolon= \hat \mu(x) + \hat k(x,\vx)(\hat k\circ{\hat \sigma^2}(\vx))^{-1}(\vy' - \hat \mu(\vx)), \;\;\forall x\in\mathfrak X,\\
\hat k_{t-1}(x,x')&\vcentcolon=  \hat k(x,x') - \hat k(x, \vx)(\hat k\circ{\hat \sigma^2}(\vx))^{-1} \hat k(\vx, x'), \;\;\forall x, x'\in\mathfrak X. 
\end{align*}

We plug in the following terms to Lemma~\ref{lem:conditional_multivariate_gaussian}. 
\begin{align*}
&\vy \gets \begin{bmatrix*}
\vy' \\ f(x)
\end{bmatrix*},\quad
\vu \gets \begin{bmatrix*}
\mu^*(\vx) \\ \mu^*(x)
\end{bmatrix*},\quad
\hat \vu \gets \begin{bmatrix*}
\hat \mu(\vx) \\ \hat\mu(x)
\end{bmatrix*},\quad \\
&    V \gets
\begin{bmatrix*}
k^*\circ\sigma^2_*(\vx) &   k^*\circ\sigma^2_*(\vx, x)\\ 
k^*\circ\sigma^2_*(x,\vx) &     k^*\circ\sigma^2_*(x)
\end{bmatrix*}
,\quad
       \hat V \gets
\begin{bmatrix*}
\hat k\circ\hat\sigma^2(\vx) &   \hat k\circ\hat\sigma^2(\vx, x)\\ 
\hat k\circ\hat\sigma^2(x,\vx) &     \hat k\circ\hat\sigma^2(x)
\end{bmatrix*}.
\end{align*}

Hence,
\begin{align*}
& \omega = \mu_{t-1}^*(x), \;\;
\kappa = k^*_{t-1}(x) + \sigma_*^2 = k^*_{t-1}\circ\sigma_*^2(x), \\
&\hat\omega = \hat \mu_{t-1}(x), \;\;
\hat\kappa = \hat k_{t-1}(x) + \hat\sigma^2 = \hat k_{t-1}\circ \hat\sigma^2(x), 
\end{align*}

Furthermore, we set $\delta_1 \gets \delta/2$ and $\delta_2 \gets \delta/2$. Thus, with probability at least $1-\delta$, we have 
\begin{align*}
   |\hat \mu_{t-1}(x) -\mu_{t-1}^*(x)|^2 < a^2 k^*_{t-1}\circ\sigma_*^2(x) \;\text{and}\; 1-2\sqrt{b } < \frac{N \hat k_{t-1}\circ \hat\sigma^2(x) }{(N-t) k^*_{t-1}\circ\sigma_*^2(x)} < 1+ 2\sqrt{b} + 2b,
\end{align*}
where $a^2 = \frac{4\left(t+2\sqrt{(t-1)\log{\frac{4}{\delta}}} + 2\log{\frac{4}{\delta}}  - 2/N\right)}{(N-t-1)\delta}, b=\frac{1}{N-t}\log\frac4{\delta}.$
\end{proof}

This completes the proof for Theorem~\ref{thm:posterior_bound}.

\subsection{Theorem~\ref{thm:regret}: Bounding the Regrets of \hyperbo}
\label{ssec:app_thm}
Theorem~\ref{thm:regret} is about regret bounds of \hyperbo with acquisition strategies that are special cases of GP-UCB and PI. Recall that in \S\ref{sec:pf}, we defined simple regret as our metric: $r_T = \max_{x\in\mathfrak X}f(x) - f(\hat x)$, and we set $\hat x = x_\tau, \tau =\argmax_{t\in [T]} y_t$. We need to bound $r_T$ for running \hyperbo on a test function $f\sim$\gtgp. Note that \hyperbo does not have access to the ground truth \gtgp. Hence classic theoretical analyses tools such as~\cite{srinivas2009gaussian} do not apply to \hyperbo.

In Theorem~\ref{thm:posterior_bound}, we showed that the pre-trained GP posterior \ppgp can be close to the ground truth posterior \pgtgp. So the hope is that optimizing an acquisition function defined by \ppgp can get similar results to optimizing an acquisition function defined by \pgtgp. However, these two posteriors are still different, and we need to make sure the acquisition functions are informed of the difference.

We analyze the following two acquisition functions. 
\begin{itemize}
    \item GP-UCB~\citep{srinivas2009gaussian} with explore-exploit trade-off parameter $\beta$:
\begin{align}
\alpha^{\text{UCB}}_{t-1}(x) = \hat \mu_{t-1}(x) + \beta \sqrt{\hat k_{t-1}\circ{\hat\sigma^2}( x)},  \label{appeq:acfun-ucb}
\end{align}
where $\beta$ is defined by $a,b$ from Theorem~\ref{thm:posterior_bound}.
\item The max-value version of PI~\citep{wang2017maxvalue, kushner1964} with $\hat f^*\geq \max_{x\in\mathfrak X}f(x)$:
\begin{align}
\alpha^{\text{PI}}_{t-1}(x) = \frac{\hat \mu_{t-1}(x) - \hat f^* }{\sqrt{\hat k_{t-1}\circ{\hat\sigma^2}( x)}}.  \label{appeq:acfun-pi}
\end{align} 
\end{itemize}

Taking GP-UCB as an example, the proof strategy is to write $r_T$ with the pre-trained GP posterior mean $\hat\mu_{t-1}$ and kernel $\hat k_{t-1}(x)$. This is to ensure that we can remove the regret's dependency on $\mu_{t-1}$ by using the critical information that for any iteration $t=1,\cdots, T$, $\alpha^{\text{UCB}}_{t-1}(x_t) \geq \alpha^{\text{UCB}}_{t-1}(x^*)$, where $x_t$ is the maximizer of the acquisition function and $x^*$ is the maximizer of test function $f$. Once $r_T$ is written with the pre-trained GP posterior, we can use Theorem~\ref{thm:posterior_bound} to bound $\hat\mu_{t-1}, \hat k_{t-1}$ in $r_T$ with the ground truth $k^*_{t-1}(x_t)$. Then, we can use the max information gain $\max_{D_f}I\left(f; D_f\right)$ to bound $k^*_{t-1}(x_t)$. Since $r_T$ can be fully expressed by $k^*_{t-1}(x_t)$, we can conclude the regret bound for $r_T$ with GP-UCB. The case for PI is similar.

Next, we review some useful lemmas before proceeding to the proofs related to Theorem~\ref{thm:regret}.

\begin{cor}[Bernoulli's inequality]
\label{cor:math} For all $0\leq x\leq c$ and $a> 0$, we have $x \leq \frac{c\log(1+\frac{ax}{c})}{\log(1+a)}$.
\end{cor}
\begin{proof}
By Bernoulli's inequality, $(1+a)^{\frac{x}{c}} \leq 1+\frac{ax}{c}$. Because $\log(1+a) > 0$, by rearranging, we have $x \leq \frac{c\log(1+\frac{ax}{c})}{\log(1+a)}$.
\end{proof}

\begin{lem}
\label{lem:math} For any $0\leq x\leq c$ and $a> 0$, we have $\sqrt{x} < \sqrt{x+a} - \frac{a}{2\sqrt{c+a}}$.
\end{lem}
\begin{proof}
Numerically, for any $n\geq 1$, $\frac{1}{\sqrt{n}} < 2\sqrt{n} - 2\sqrt{n-1}$~\citep{wolfram}. Let $n = \frac{x}{a} + 1$. Then, we have
\begin{align*}
\frac{1}{\sqrt{\frac{x}{a} + 1}} &< 2\sqrt{\frac{x}{a} + 1} - 2\sqrt{\frac{x}{a}}\\
\frac{a}{\sqrt{a+c}} < \frac{a}{\sqrt{a+x}}&< 2\sqrt{x + a} - 2\sqrt{x}\\
\sqrt{x} &< \sqrt{x + a} - \frac{a}{2\sqrt{a+c}}. \qedhere
\end{align*}
\end{proof}
\begin{lem}[Lemma 5.3 of \cite{srinivas2009gaussian}] \label{lem:rho}
Let $\vx_T = [x_t]_{t=1}^T \subseteq \fx$ and $f\sim\GP(\mu, k)$ with observation noise $\sigma^2$. The mutual information between 
the function values $f(\vx_T)$ and their observations $\vy_T = [y_t]_{t=1}^T$ satisfies
$$I(f(\vx_T);\vy_T) = \frac12 \sum_{t=1}^T\log(1+\sigma^{-2} k_{t-1}(x_t)).$$
\end{lem}
\begin{cor}[\cite{srinivas2009gaussian}]
\label{lem:gauss} For any Gaussian variable $x\sim \mathcal N(\mu, \sigma^2), x\in \R,$ 
$$\Pr[x-\mu   \leq \zeta\sigma]\geq 1-\delta, \; \Pr[x-\mu   \geq -\zeta\sigma]\geq 1-\delta$$
where $\zeta = (2\log(\frac{1}{2\delta}))^\frac12$ and $\delta\in(0,1)$.
\end{cor}
\begin{proof}
Let $z=\frac{\mu - x}{\sigma}\sim\mathcal N(0,1) $. We have 
\begin{align*}
\Pr[z>\zeta] &= \int_{\zeta}^{+\infty} \frac{1}{\sqrt{2\pi}}e^{-z^2/2}\dif z \\
&=\int_{\zeta}^{+\infty} \frac{1}{\sqrt{2\pi}}e^{-(z-\zeta)^2/2-\zeta^2/2-z\zeta}\dif z\\
&\leq e^{-\zeta^2/2}\int_{\zeta}^{+\infty} \frac{1}{\sqrt{2\pi}}e^{-(z-\zeta)^2/2}\dif z\\
&=\frac12 e^{-\zeta^2/2},
\end{align*}
since $e^{-z\zeta}\leq 1$ with $z\geq \zeta$. Similarly, $\Pr[z < -\zeta] \leq \frac12 e^{-\zeta^2/2}.$ We reach the conclusion by rearranging the constants.
\end{proof}

The next lemma bounds the difference between the best sample in observations and the best observed sample.
\begin{lem} \label{lem:app_best_sample_diff}
With probability at least $1-\delta$, we have 
\begin{align*}
    f(x_{\tau'}) - f(x_\tau) \leq 2(2\log \frac{1}{\delta})^{\frac12} \sigma,
\end{align*}
where $\tau' = \argmax_{t\in [T]} f(x_t)$ and $\tau =\argmax_{t\in [T]} y_t$.
\end{lem} 
\begin{proof}
Note that $y_t \sim \N(f(x_t), \sigma^2), \forall t\in[T]$. By Corollary~\ref{lem:gauss}, with probability at least $1-\delta$, 
\begin{align*}
    f(x_{\tau}) + C \sigma \geq  y_\tau  \geq y_{\tau'} \geq f(x_{\tau'}) - C \sigma,
\end{align*}
where $C = (2\log \frac{1}{\delta})^{\frac12}$.  
Hence $f(x_{\tau'}) - f(x_{\tau}) \leq 2C\sigma$.
\end{proof}

Now we are fully equipped to prove the regret bounds. Next, we show the regret bound for GP-UCB (Eq.~\ref{appeq:acfun-ucb}) in Theorem~\ref{app_thm:regret_ucb}, and then PI (Eq.~\ref{appeq:acfun-pi}) in Theorem~\ref{app_thm:regret_pi}. Note that we do not assume the ground truth GP is known, but we use the ground truth GP to bound the regret. We provide an alternative regret bound parameterized by the pre-trained GP in \S\ref{app:ssec:alternative}. %

\begin{thm} \label{app_thm:regret_ucb}
Assume $N>T+4\log\frac{12}{\delta}$, $\delta\in(0,1)$, and there exists constant $c$ such that $c \geq k^*(x)$ for any $x\in\mathfrak X$. With probability at least $1-\delta$, GP-UCB in Eq.~\ref{appeq:acfun-ucb} obtains the simple regret
\begin{align*} 
    r^{\text{GP-UCB}}_T   \leq  (a+\zeta)(1+s) \sqrt{\frac{2c\rho_T}{T\log(1+c \sigma_*^{-2})} + \sigma_*^2} +  \zeta\sigma_*(1-\frac{\sigma_*}{\sqrt{c+\sigma_*^2}}),
\end{align*}
where $\beta = (a+\zeta)\sqrt{\frac{N}{(N-T)(1-2\sqrt{b})}}$, $\zeta = \sqrt{2\log(\frac{3}{\delta})}$, $a^2 = \frac{12\left(T+2\sqrt{(T-1)\log{\frac{12}{\delta}}} + 2\log{\frac{12}{\delta}}  - 2/N\right)}{(N-T-1)\delta}, b=\frac{1}{N-T}\log\frac{12}{\delta}, s = \sqrt{\frac{1+2\sqrt{b} + 2b}{1-2\sqrt{b}}},$ and $\rho_T = \underset{A\subseteq\fx, |A|=T}{\max}\frac12\log|\mI+\sigma^{-2} k^*(A)|.$ 

\end{thm}
\begin{proof}
Let $\tau = \argmin_{t\in[T]} k^*_{t-1}(x_t)$, $x^* = \argmax_{x\in\mathfrak X} f(x)$ and $\hat x = x_{t'}$ where $t' = \argmax_{t\in[T]} y_t$. We set $\zeta = \sqrt{2\log(\frac{1}{\delta'})}$ with $\delta' = \delta / 3$. By Lemma~\ref{lem:app_best_sample_diff} , with probability at least $1-\delta'$, 
\begin{align*}
r^{\text{GP-UCB}}_T  &= \left(f(x^*)  - \max_{t\in[T]} f(x_t)\right) +  \left(\max_{t\in[T]} f(x_t) - f(\hat x)\right)\\
& \leq r + \zeta\sigma,
\end{align*}
where $r\vcentcolon=f(x^*)  - \max_{t\in[T]} f(x_t)$. We use the following short-hands for simplicity:
\begin{align*}
    & \Delta_\mu = \mu^*_{\tau-1}(x^*) - \mu^*_{\tau-1}(x_\tau),\quad  \hat \Delta_\mu = \hat \mu_{\tau-1}(x^*) - \hat \mu_{\tau-1}(x_\tau), \\
    &  \lambda(\cdot) = \sqrt{k^*_{\tau-1}(\cdot) + \sigma_*^2},\quad   \hat \lambda(\cdot) = \sqrt{\hat k_{\tau-1}(\cdot) + \hat\sigma^2},\quad \text{and}\;\; \psi(\cdot) = \sqrt{k^*_{\tau-1}(\cdot)} 
\end{align*}

Next we  bound $r$.  By Corollary~\ref{lem:gauss}, with probability at least $1-\delta'$, 
\begin{align*}
r & \leq f(x^*)  - f(x_\tau) \\
&\leq \Delta_\mu + \zeta\psi(x^*) +\zeta\psi(x_\tau).
\end{align*}

By Theorem~\ref{thm:posterior_bound} and that $\tau \leq T$, the following inequality holds with probability at least $1-\delta'$,
\begin{align*}
\Delta_\mu < \hat \Delta_\mu + a\lambda(x^*) + a\lambda(x_\tau) \;\;\text{and}\;\; \hat\lambda(x_\tau) < h\lambda(x_\tau)  \;\;\text{and}\;\; \lambda(x^*) < h'\hat\lambda(x^*).
\end{align*}
where $a^2 = \frac{4\left(T+2\sqrt{(T-1)\log{\frac{4}{\delta'}}} + 2\log{\frac{4}{\delta'}}  - 2/N\right)}{(N-T-1)\delta'}, b=\frac{1}{N-T}\log\frac4{\delta'}, h = \sqrt{\frac{(1+2\sqrt{b}+2b)(N-T)}{N}}, h' = \sqrt{\frac{N}{(N-T)(1-2\sqrt{b})}}.$

By Lemma~\ref{lem:math} and that $c\geq\sup_{x\in\mathfrak X}k^*(x)\geq k^*_{\tau-1}(x^*)$, we have,
\begin{align*}
\psi(x^*) \leq \lambda(x^*) - \frac{\sigma_*^2}{2\sqrt{c+\sigma_*^2}}
\end{align*}

Notice that because of the acquisition strategy of GP-UCB, i.e., $\alpha^{\text{UCB}}_{\tau-1}(x) = \hat \mu_{\tau-1}(x) + \beta \hat \lambda(x)$, the following inequality holds,
\begin{align*}
    \hat \mu_{\tau-1}(x^*) + \beta \hat \lambda(x^*) &\leq \hat \mu_{\tau-1}(x_\tau) + \beta \hat \lambda(x_\tau),\;\;\; \text{i.e.,}\\
    \hat \Delta_\mu &\leq \beta \left(\hat \lambda(x_\tau) -\hat \lambda(x^*) \right).
\end{align*}

Note that $\delta' = \delta / 3$, $\beta = (a+\zeta)h'$ and $\sqrt{\sigma_1^2 + \sigma_2^2} < \sigma_1 + \sigma_2$ for $\sigma_1, \sigma_2>0$. And let $s = hh'= \sqrt{\frac{1+2\sqrt{b} + 2b}{1-2\sqrt{b}}}$. Hence, with probability at least $1-\delta$,
\begin{align*}
r & < \Delta_\mu + \zeta\lambda(x^*) + \zeta\psi(x_\tau) - \frac{\zeta\sigma_*^2}{2\sqrt{c+\sigma_*^2}} \\
& < \hat \Delta_{\mu} + (a+\zeta)\lambda(x^*) + a\lambda(x_\tau) + \zeta\psi(x_\tau) - \frac{\zeta\sigma_*^2}{2\sqrt{c+\sigma_*^2}} \\
& < \beta \left(\hat \lambda(x_\tau) -\hat \lambda(x^*) \right) + (a+\zeta)h'\hat\lambda(x^*) + a\lambda(x_\tau) + \zeta\psi(x_\tau) - \frac{\zeta\sigma_*^2}{2\sqrt{c+\sigma_*^2}} \\
& < ((a+\zeta)s+a+\zeta)\lambda(x_\tau) \ - \frac{\zeta\sigma_*^2}{\sqrt{c+\sigma_*^2}}
\end{align*}

By Corollary~\ref{cor:math} and the fact that $\tau = \argmin_{t\in[T]} k^*_{t-1}(x_t)$, we have
\begin{align*}
k^*_{\tau-1}(x_\tau) & \leq\frac{1}{T}\sum_{t=1}^T k^*_{t-1}(x_t) \\
& \leq \frac{1}{T}\sum_{t=1}^T  \frac{c\log(1+\frac{c \sigma_*^{-2}k^*_{t-1}(x_t)}{c})}{\log(1+c \sigma_*^{-2})} \\
& =  \frac{c}{T\log(1+c \sigma_*^{-2})}\sum_{t=1}^T \log(1+\sigma_*^{-2}k^*_{t-1}(x_t)).
\end{align*}
Notice that here Corollary~\ref{cor:math} applies because $0\leq k^*_{\tau-1}(x_\tau)\leq c$.

By Lemma~\ref{lem:rho}, we have $I(f(\vx_T);\vy_T) = \frac12 \sum_{t=1}^T\log(1+\sigma_*^{-2} k^*_{t-1}(x_t)) \leq \rho_T$, so 
$$k^*_{\tau-1}(x_\tau) \leq  \frac{2c\rho_T}{T\log(1+c \sigma_*^{-2})},$$
which implies 
\begin{align*}
r^{\text{GP-UCB}}_T  &  <  (a+\zeta)(1+s) \sqrt{\frac{2c\rho_T}{T\log(1+c \sigma_*^{-2})} + \sigma_*^2} +  \zeta\sigma_*(1-\frac{\sigma_*}{\sqrt{c+\sigma_*^2}}).
\end{align*}
Note that $\rho_T$ is the max information gain over $T$ observations of the ground truth GP.
\end{proof}

\begin{thm} \label{app_thm:regret_pi}
Assume $N>T+4\log\frac{12}{\delta}$, $\delta\in(0,1)$,  and there exists constant $c = \sup_{x\in\mathfrak X}k^*(x)$. 
With probability at least $1-\delta$, the simple regret in T iterations of PI in Eq.~\ref{appeq:acfun-pi} that uses $\hat f^*\geq \max_{x\in\fx}f(x)$ as its target value satisfies 
 \begin{align*}
 r^{\text{PI}}_T  & < \eta \sqrt{\frac{2c\rho_T}{T\log(1+c \sigma_*^{-2})} + \sigma_*^2} + \zeta\sigma_*(1-\frac{\sigma_*}{2\sqrt{c+\sigma_*^2}}),
 \end{align*}
 where $\eta = s(\frac{\hat f^* - \mu^*_{\tau-1}(x^*)}{\sqrt{k^*_{\tau-1}(x^*) + \sigma_*^2}}+a)+a + \zeta$, and $\zeta = \sqrt{2\log(\frac{3}{\delta})}$, $a^2 = \frac{12\left(T+2\sqrt{(T-1)\log{\frac{12}{\delta}}} + 2\log{\frac{12}{\delta}}  - 2/N\right)}{(N-T-1)\delta}, b=\frac{1}{N-T}\log\frac{12}{\delta},$ $\rho_T = \underset{A\subseteq\fx, |A|=T}{\max}\frac12\log|\mI+\sigma^{-2}\hat k(A)|$, $s = \sqrt{\frac{1+2\sqrt{b} + 2b}{1-2\sqrt{b}}}$,  $\tau = \argmin_{t\in[T]} k^*_{t-1}(x_t)$.
\end{thm}

\begin{proof}
The acquisition strategy PI in Eq.~\ref{appeq:acfun-pi} uses target value $\hat f^*$ which is an upper bound on $f$. By Corollary~\ref{lem:gauss}, with probability at least $1-\delta'$, 
\begin{align*}
\hat f^*  - \max_{t\in[T]} f(x_t)
& \leq \hat f^*  - f(x_\tau) \\
&\leq \hat f^*  - \mu^*_{\tau-1}(x_\tau) + \mu^*_{\tau-1}(x_\tau)  - f(x_\tau) \\
&\leq  \hat f^*  - \mu^*_{\tau-1}(x_\tau)  +\zeta\sqrt{k^*_{\tau-1}(x_\tau)},
\end{align*}
where $\zeta = \sqrt{2\log(\frac{1}{\delta'})}$ and $\tau = \argmin_{t\in[T]} k^*_{t-1}(x_t)$. We use the following short-hands for simplicity:
\begin{align*}
    & \lambda_\tau = \sqrt{k^*_{\tau-1}(x_\tau) + \sigma_*^2}, \quad \hat \lambda_\tau = \sqrt{\hat k_{\tau-1}(x_\tau) + \hat\sigma^2}, \\
    &  \lambda = \sqrt{k^*_{\tau-1}(x^*) + \sigma_*^2},\quad   \hat \lambda = \sqrt{\hat k_{\tau-1}(x^*) + \hat\sigma^2}.
\end{align*}

By Theorem~\ref{thm:posterior_bound} and the selection strategy of PI, with probability at least $1-\delta'$,
 \begin{align*}
  \hat f^*  - \mu^*_{\tau-1}(x_\tau) & < \hat f^* - \hat \mu_{\tau-1}(x_{\tau}) + a \lambda_\tau \\
  & \leq \frac{\hat f^* - \hat\mu_{\tau-1}(x^*)}{ \hat\lambda } \hat\lambda_\tau + a \lambda_\tau \\
  & <\frac{\hat f^* - \mu^*_{\tau-1}(x^*) + a \lambda}{\hat \lambda} \hat\lambda_\tau + a\lambda_\tau \\
  & < \frac{\hat f^* - \mu^*_{\tau-1}(x^*) + a \lambda}{ \lambda} s\lambda_\tau + a\lambda_\tau \\
  & = s\left(\frac{\hat f^* - \mu^*_{\tau-1}(x^*)}{\lambda} +a\right) \lambda_\tau + a\lambda_\tau,
 \end{align*}
 where we set $a^2 = \frac{4\left(T+2\sqrt{(T-1)\log{\frac{4}{\delta'}}} + 2\log{\frac{4}{\delta'}}  - 2/N\right)}{(N-T-1)\delta'}, b=\frac{1}{N-T}\log\frac4{\delta'}, s = \sqrt{\frac{1+2\sqrt{b} + 2b}{1-2\sqrt{b}}}$. 
 
 Let $\eta = s(\frac{\hat f^* - \mu^*_{\tau-1}(x^*)}{\sqrt{k^*_{\tau-1}(x^*) + \sigma_*^2}}+a)+a + \zeta$ and $\delta' = \delta / 3$. Then, with probability at least $1 - \delta$, the simple regret of PI satisfy
 \begin{align*}
 r^{\text{PI}}_T  & < \eta \sqrt{\frac{2c\rho_T}{T\log(1+c \sigma_*^{-2})} + \sigma_*^2} + \zeta\sigma_*(1-\frac{\sigma_*}{2\sqrt{c+\sigma_*^2}}).
 \end{align*}
\end{proof}

Theorem~\ref{thm:regret} is a summary of Theorem~\ref{app_thm:regret_ucb} and Theorem~\ref{app_thm:regret_pi}. While the regret bounds provide us with some understanding of \hyperbo in a specific setting, in practice, we need to query in a continuous input space that goes beyond the finite set of points present in the training dataset. It may or may not be possible to obtain data on a wide range of tasks to ensure $N > T + 1$. In fact, in all of our experiment, this criterion on number of tasks is not satisfied. However, we still obtained good performance.

\subsection{An Alternative Approach to Bound the Regret}
\label{app:ssec:alternative}
In this section, we provide an alternative regret bound defined by the pre-trained GP in Theorem~\ref{app_thm:regret_ucb_alternative}. This allows us to potentially examine the regret based on the observable parameters from the pre-trained GP. Here we set the GP-UCB acquisition function as 
\begin{align}\label{appeq:alternative_ucb}
    \alpha^{\text{UCB}}_{t-1}(x) = \hat \mu_{t-1}(x) + \beta \sqrt{\hat k_{t-1}( x)},
\end{align}
which is more realistic than  Eq.~\ref{appeq:acfun-ucb} in practical settings.

\begin{thm} \label{app_thm:regret_ucb_alternative}
Assume $N>T+4\log\frac{12}{\delta}$, $\delta\in(0,1)$, and there exists constant $c$ such that $c \geq k^*(x)$ and $c \geq \hat k(x)$ for any $x\in\mathfrak X$. With probability at least $1-\delta$, GP-UCB in Eq.~\ref{appeq:alternative_ucb} obtains the simple regret
\begin{align*}
    r^{\text{GP-UCB}}_T   \leq   2\beta \sqrt{\frac{2c\rho_T}{T\log(1+c \hat\sigma^{-2})}} + \beta \hat\sigma +  \zeta\sigma_*(1-\frac{\sigma_*}{\sqrt{c+\sigma_*^2}}),
\end{align*}
where $\beta = (a+\zeta)s$, $\zeta = \sqrt{2\log(\frac{3}{\delta})}$, $a^2 = \frac{12\left(T+2\sqrt{(T-1)\log{\frac{12}{\delta}}} + 2\log{\frac{12}{\delta}}  - 2/N\right)}{(N-T-1)\delta}, b=\frac{1}{N-T}\log\frac{12}{\delta}, s = \sqrt{\frac{N}{(N-T)(1-2\sqrt{b})}},$ and $\rho_T = \underset{A\subseteq\fx, |A|=T}{\max}\frac12\log|\mI+\hat\sigma^{-2}\hat k(A)|.$ 

\end{thm}
\begin{proof}
Let $\tau = \argmin_{t\in[T]} \hat k_{t-1}(x_t)$, $x^* = \argmax_{x\in\mathfrak X} f(x)$ and $\hat x = x_{t'}$ where $t' = \argmax_{t\in[T]} y_t$. We set $\zeta = \sqrt{2\log(\frac{1}{\delta'})}$ with $\delta' = \delta / 3$. By Lemma~\ref{lem:app_best_sample_diff} , with probability at least $1-\delta'$, 
\begin{align*}
r^{\text{GP-UCB}}_T  &= \left(f(x^*)  - \max_{t\in[T]} f(x_t)\right) +  \left(\max_{t\in[T]} f(x_t) - f(\hat x)\right)\\
& \leq r + \zeta\sigma_*,
\end{align*}
where $r=f(x^*)  - \max_{t\in[T]} f(x_t)$. We use the following short-hands for simplicity:
\begin{align*}
    & \Delta = \sqrt{k^*_{\tau-1}(x^*) + \sigma^2_*} + \sqrt{k^*_{\tau-1}(x_\tau) + \sigma^2_*}, \quad \hat \Delta = \sqrt{\hat k_{\tau-1}(x^*) + \hat \sigma^2} + \sqrt{\hat k_{\tau-1}(x_\tau) + \hat \sigma^2}, \\
    & \Delta_\mu = \mu^*_{\tau-1}(x^*) - \mu^*_{\tau-1}(x_\tau),\quad  \hat \Delta_\mu = \hat \mu_{\tau-1}(x^*) - \hat \mu_{\tau-1}(x_\tau).
\end{align*}

Next we  bound $r$.  By Corollary~\ref{lem:gauss}, with probability at least $1-\delta'$, 
\begin{align*}
r & \leq f(x^*)  - f(x_\tau) \\
&\leq \Delta_\mu + \zeta\sqrt{k^*_{\tau-1}(x^*)} +\zeta\sqrt{k^*_{\tau-1}(x_\tau)}.
\end{align*}

By Theorem~\ref{thm:posterior_bound} and that $\tau \leq T$, the following inequality holds with probability at least $1-\delta'$,
\begin{align*}
\Delta_\mu < \hat \Delta_\mu + a\Delta\;\; \text{and} \;\;
\Delta  < s \hat\Delta,
\end{align*}
where $a^2 = \frac{4\left(T+2\sqrt{(T-1)\log{\frac{4}{\delta'}}} + 2\log{\frac{4}{\delta'}}  - 2/N\right)}{(N-T-1)\delta'}, b=\frac{1}{N-T}\log\frac4{\delta'}, s = \sqrt{\frac{N}{(N-T)(1-2\sqrt{b})}}.$

By Lemma~\ref{lem:math} and that $c\geq\sup_{x\in\mathfrak X}k^*(x)\geq k^*_{\tau-1}(x^*)$, we have,
\begin{align*}
\sqrt{k^*_{\tau-1}(x^*)} + \sqrt{k^*_{\tau-1}(x_\tau)} \leq \Delta  - \frac{\sigma_*^2}{\sqrt{c+\sigma_*^2}}
\end{align*}

Notice that because of the acquisition strategy of GP-UCB, i.e., $\alpha^{\text{UCB}}_{\tau-1}(x) = \hat \mu_{\tau-1}(x) + \beta \sqrt{\hat k_{\tau-1}(x)}$, the following inequality holds,
\begin{align*}
    \hat \mu_{\tau-1}(x^*) + \beta \sqrt{\hat k_{\tau-1}( x^*)} &\leq \hat \mu_{\tau-1}(x_\tau) + \beta \sqrt{\hat k_{\tau-1}( x_\tau)},\;\;\; \text{i.e.,}\\
    \hat \Delta_\mu &\leq \beta \left(\sqrt{\hat k_{\tau-1}( x_\tau)} -\sqrt{\hat k_{\tau-1}( x^*)} \right).
\end{align*}

Note that $\delta' = \delta / 3$, $\beta = (a+\zeta)s$ and $\sqrt{\sigma_1^2 + \sigma_2^2} < \sigma_1 + \sigma_2$ for $\sigma_1, \sigma_2>0$. Hence, with probability at least $1-\delta$,
\begin{align*}
r & < \Delta_\mu + \zeta\Delta - \frac{\zeta\sigma_*^2}{\sqrt{c+\sigma_*^2}} \\
& < \hat \Delta_\mu + (a+\zeta)s\hat \Delta  - \frac{\zeta\sigma_*^2}{\sqrt{c+\sigma_*^2}} \\
& \leq \beta \left(\sqrt{\hat k_{\tau-1}( x_\tau)} -\sqrt{\hat k_{\tau-1}( x^*)} \right)  + (a+\zeta)s\hat \Delta  - \frac{\zeta\sigma_*^2}{\sqrt{c+\sigma_*^2}} \\
& < 2\beta(\sqrt{\hat k_{\tau-1}( x_\tau)} ) + \beta\hat\sigma - \frac{\zeta\sigma_*^2}{\sqrt{c+\sigma_*^2}}.
\end{align*}

By Corollary~\ref{cor:math} and the fact that $\tau = \argmin_{t\in[T]} \hat k_{t-1}(x_t)$, we have
\begin{align*}
\hat k_{\tau-1}(x_\tau) & \leq\frac{1}{T}\sum_{t=1}^T \hat k_{t-1}(x_t) \\
& \leq \frac{1}{T}\sum_{t=1}^T  \frac{c\log(1+\frac{c \hat \sigma^{-2}\hat k_{t-1}(x_t)}{c})}{\log(1+c \hat \sigma^{-2})} \\
& =  \frac{c}{T\log(1+c \hat \sigma^{-2})}\sum_{t=1}^T \log(1+\hat \sigma^{-2}\hat k_{t-1}(x_t)).
\end{align*}
Notice that here Corollary~\ref{cor:math} applies because $0\leq \hat k_{\tau-1}(x_\tau)\leq c$.

By Lemma~\ref{lem:rho}, if we create an imaginary function $f'\sim \GP(\hat\mu, \hat k)$ with observation noise $\hat \sigma^2$, we get $I(f'(\vx_T);\vy_T) = \frac12 \sum_{t=1}^T\log(1+\hat\sigma^{-2} \hat k_{t-1}(x_t)) \leq \rho_T$, so 
$$\hat k_{\tau-1}(x_\tau) \leq  \frac{2c\rho_T}{T\log(1+c \hat\sigma^{-2})},$$
which implies 
\begin{align*}
r^{\text{GP-UCB}}_T  &  <  2\beta \sqrt{\frac{2c\rho_T}{T\log(1+c \hat\sigma^{-2})}} + \beta \hat\sigma +  \zeta\sigma_*(1-\frac{\sigma_*}{\sqrt{c+\sigma_*^2}}).
\end{align*}
Note that $\rho_T$ is the max information gain over $T$ observations of the pre-trained GP. The last term,  $\zeta\sigma_*(1-\frac{\sigma_*}{\sqrt{c+\sigma_*^2}})$, involves the ground truth noise parameter, and it is irreducible due to the assumed observation noise.
\end{proof}

\section{Experiment Details and More Results}
\label{app:exp}
In this section, we provide more empirical results on the impact of objective functions and acquisition functions in \hyperbo. The experiment setups on PD1 are the same as \S\ref{sssec:hold-out-related}: offline and holding out related tasks. The experiment setups on HPO-B are the same as \S\ref{ssec:hpob}, which follows the original settings from \cite{pineda2021hpob}.

\subsection{More Results on the PD1 Online Tuning Tasks}
\begin{figure}
    \centering
    \includegraphics[width=\textwidth]{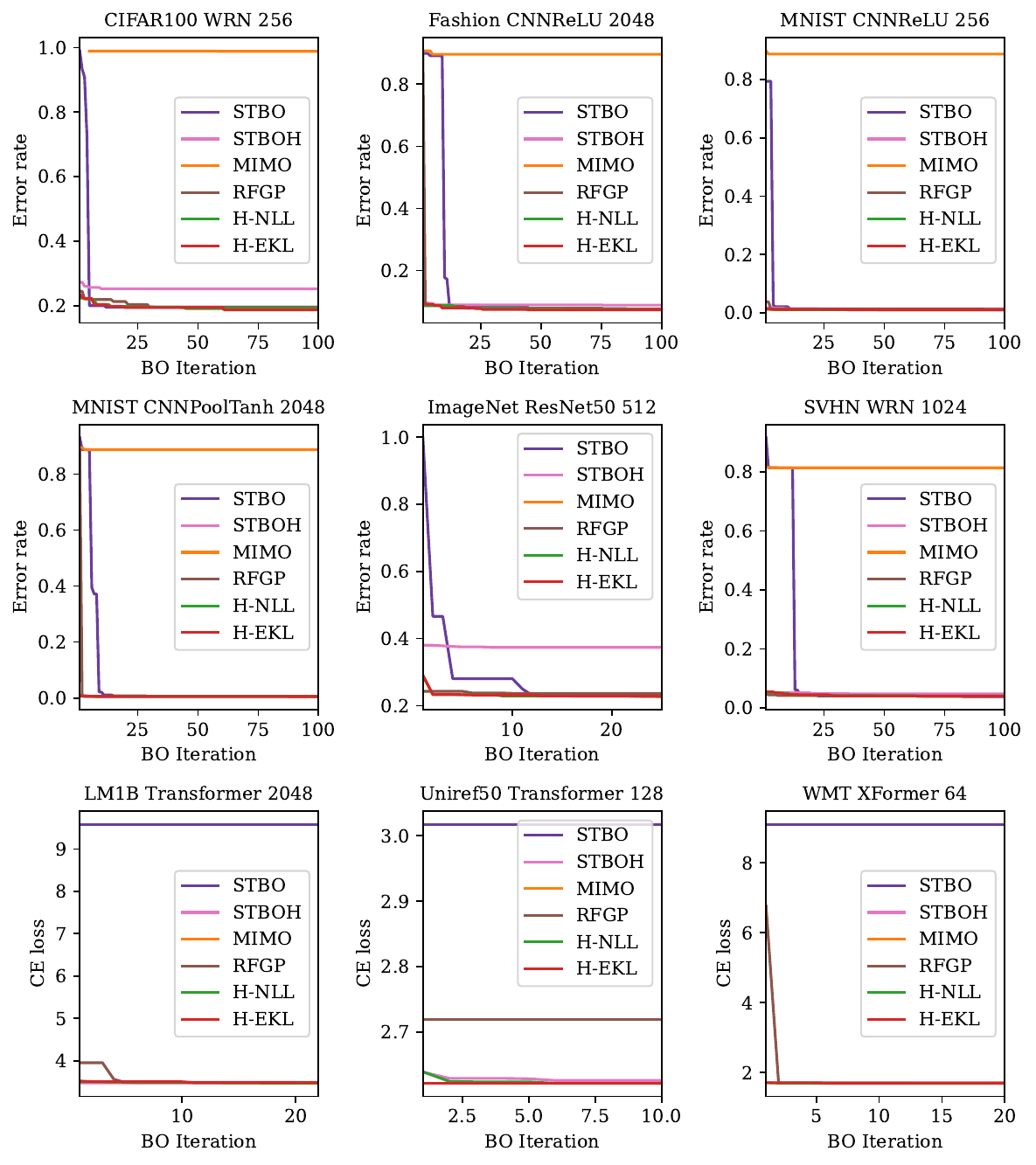}
    \caption{Extension of Figure~\ref{fig:online} to include more methods. STBO, MIMO, RFGP encountered many infeasible evaluations due to their acquisition strategies.}
    \label{fig_app:online}
\end{figure}

Figure~\ref{fig_app:online} is an extension of Figure~\ref{fig:online} and it includes more methods that are evaluated on the PD1 online tuning tasks. Due to practical constraints, we set \hyperbo variants and STBO to share exactly the same GP-UCB acquisition function as STBOH, MIMO and RFGP. We used $1.8$ as the UCB coefficient for all methods.

Overall, \hyperbo methods achieved more robust and better results than baselines on most of the tasks. It was very difficult for STBO, MIMO and RFGP to recover from points with infeasible evaluations, since they kept ``exploiting'' inputs with infeasible evaluations and as a result performed poorly. MIMO, in particular, got stuck in all of the online experiments. On ImageNet ResNet50 512, LM1B Transfoermer 2048, Uniref50 Transformer 128 and WMT XFormer 64, all of the hyperparameters selected by MIMO resulted in infeasible training runs, and so its regret curves did not exist for these tasks.

\subsection{Impact of Acquisition Functions}
\label{app:exp-acfun}
\begin{figure}
    \centering
    \includegraphics[width=0.95\textwidth]{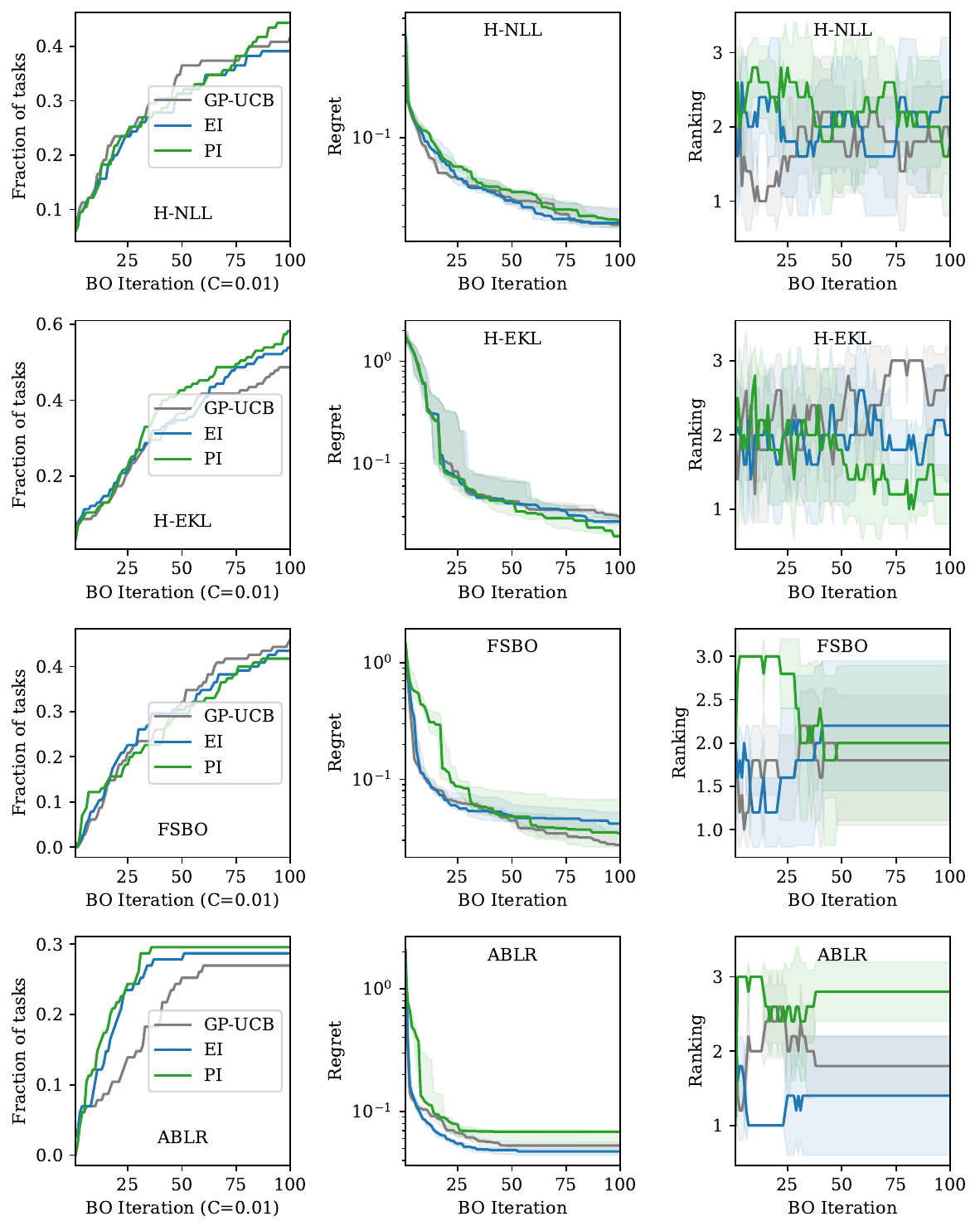}
    \caption{PD1. We compare the performance of 3 different acquisition functions under 4 variants of \hyperbo. The results of these acquisition functions are very similar and within each other's confidence intervals. The performance profiles (first column) show that PI (with threshold 0.1) has slightly better and more stable performance than EI and GP-UCB.}
    \label{fig:comp_acfun_pd1}
\end{figure}

\begin{figure}
    \centering
    \includegraphics[width=\textwidth]{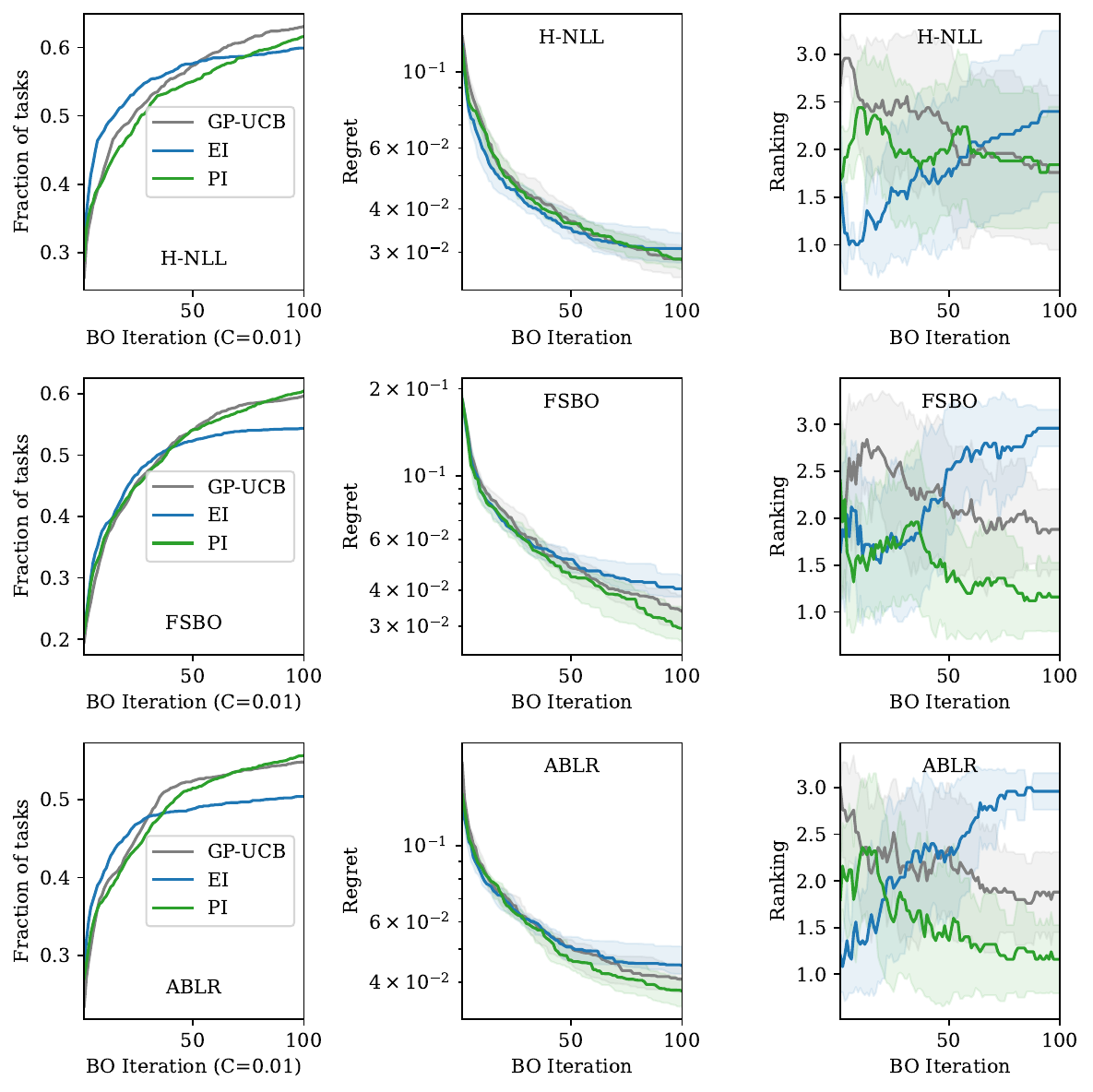}
    \caption{HPO-B. We compare the performance of 3 different acquisition functions under 3 variants of \hyperbo. There are no significant differences. PI (with threshold 0.1) and GP-UCB (with coefficient 3) achieved slightly better results than EI.}
    \label{fig:comp_acfun_hpob}
\end{figure}

We tested 3 acquisition functions in our experiments: PI from \S\ref{ssec:exp-methods}, EI~\citep{saltenis1971method, mockus1974} and GP-UCB with coefficient $\beta^{\frac12} = 3$ in Eq.~(6) of \cite{srinivas2009gaussian}. We conducted these experiments on PD1 and HPO-B for all available \hyperbo variants. The results for PD1 are shown in Figure~\ref{fig:comp_acfun_pd1} and HPO-B in Figure~\ref{fig:comp_acfun_hpob}. Both sets of results show that \hyperbo methods are not very sensitive to these 3 choices of acquisition functions, and overall PI has slightly better and more stable performance than EI and GP-UCB.

\subsection{Impact of Mean and Kernel Structures}
One open question we did not dive into in \S\ref{ssec:objective_overview} was how to determine the functional structures for mean and kernel functions to be used for pre-training. Here we tested different mean and kernel structures to evaluate their impacts on pre-trained GPs and BO results. 

\begin{figure}
    \centering
    \includegraphics[width=\textwidth]{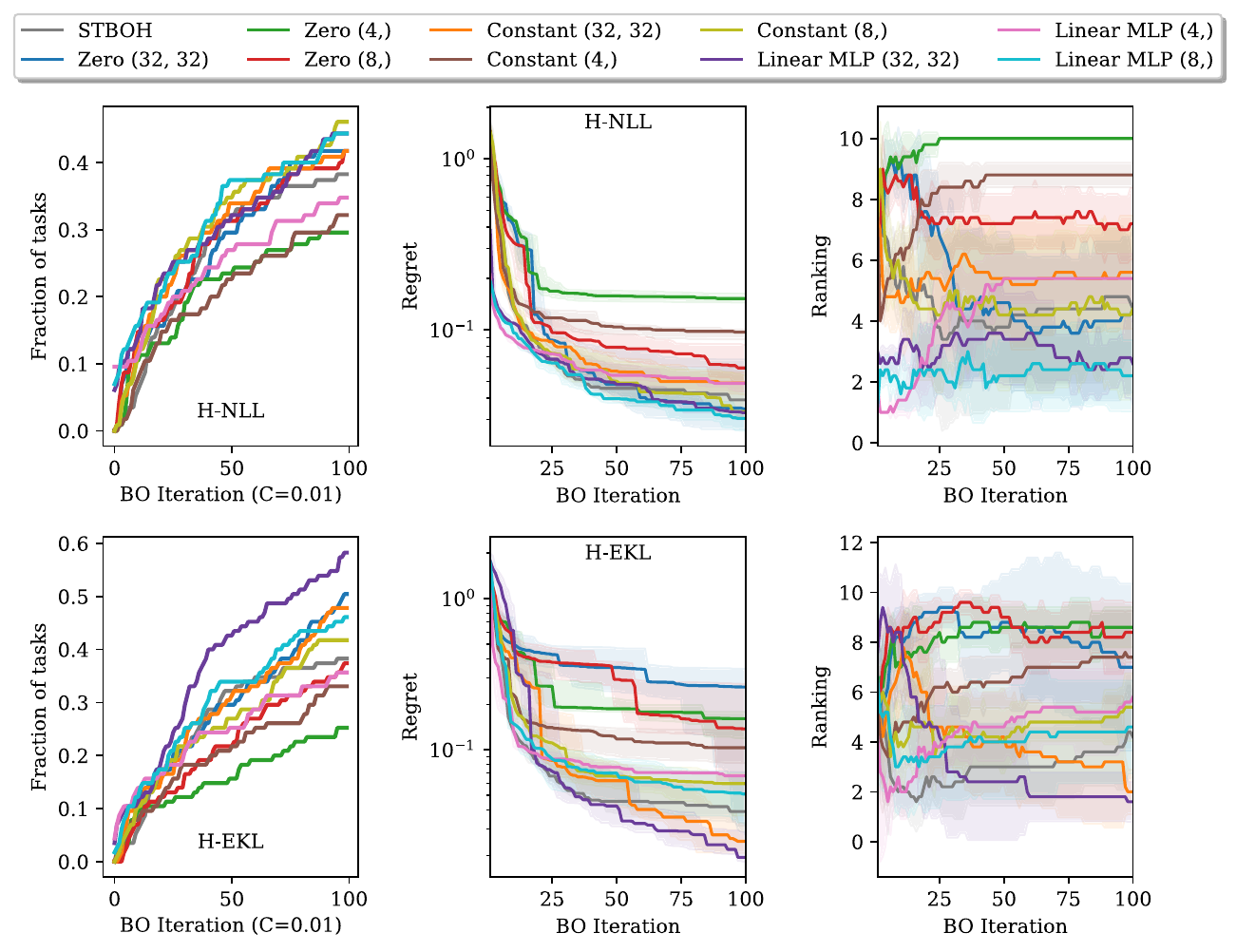}
    \caption{PD1. We compare the performance of 10 different mean and kernel structures of \hyperbo. The kernel is \matern52 with MLP features. The legend means mean function type and the numbers mean the feature size. Acquisition function is PI.}
    \label{fig:comp_structure-pd1}
\end{figure}

In Figure~\ref{fig:comp_structure-pd1} on PD1, we show the comparisons between 9 different mean and kernel combinations. The choices for the mean function include zero mean, constant mean and a linear MLP. We used \matern52 kernel on top of the same MLP as the linear MLP mean function. The MLP architectures include 1-hidden layer, either 4 or 8 neurons, or 2-hidden layer, with 32 neurons per layer. Note that the input dimension of PD1 is 4. We abbreviate these structures to ``Zero (4,)'' denoting zero mean with (4,) architecture for the \matern52 MLP kernel, ``Linear MLP (32, 32)'' denoting linear MLP mean function with (32, 32) architecture for both the mean and the \matern52 MLP kernel, etc.

For H-NLL, Zero (4,) is the simplest model among all and it performed worse than other structures. Across BO iterations, Linear MLP (8,) and Linear MLP (32, 32) achieved better performance than \hyperbo with other mean / kernel architectures. Towards 100 BO iterations, Linear MLP (8,), Linear MLP (32, 32), Zero (32, 32) and Constant (8,) achieved roughly the same performance.

For H-EKL, the performance gaps for these different structures are larger than H-NLL. Using zero mean with any of the kernel structures resulted in much worse overall BO performance than other structures. For other structures, in general, \hyperbo using a linear MLP mean function performed better than its constant mean counterparts. Linear MLP (32, 32) and Constant (32, 32) achieved better regrets towards finishing 100 BO iterations.

\begin{figure}
    \centering
    \includegraphics[width=\textwidth]{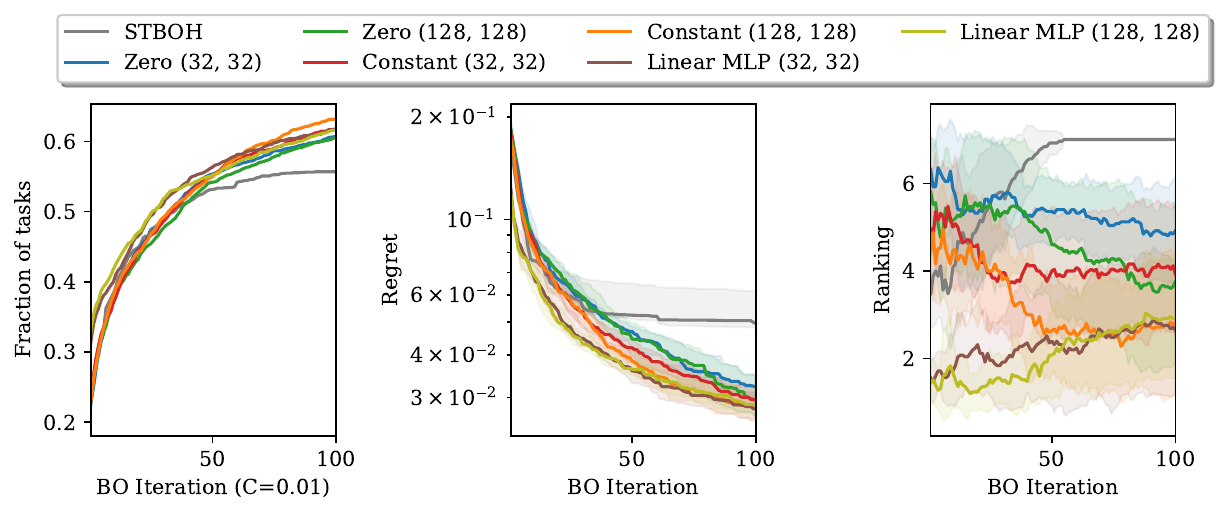}
    \caption{HPOB with H-NLL. We compare the performance of 7 different mean and kernel structures of \hyperbo. The kernel is \matern52 with MLP features. The legend denotes mean function type and the numbers are the feature sizes. Acquisition function is PI.}
    \label{fig:comp_structure-hpob}
\end{figure}

In Figure~\ref{fig:comp_structure-hpob} on HPO-B, similar to the experiment on PD1, we show the comparisons for several different mean and kernel structures, also with \matern52 kernel with MLP features. With gradually more complex structures, in general, H-NLL was able to obtain gradually better regrets. Similar to the results in PD1, \hyperbo with zero mean performed relatively worse. Across BO iterations, Linear MLP (128, 128) and Linear MLP (32, 32) achieved more stable and better results. %

Overall, Figure~\ref{fig:comp_structure-pd1} and Figure~\ref{fig:comp_structure-hpob} show that \hyperbo using a linear MLP mean function often outperforms using zero mean. While there exist performance differences among different mean and kernel structures, with relatively complex architectures, the regrets did not change much in our experiments. 

\label{app:exp-structure}

\subsection{More Results on HPO-B}
\begin{figure}
    \centering
    \includegraphics[width=\textwidth]{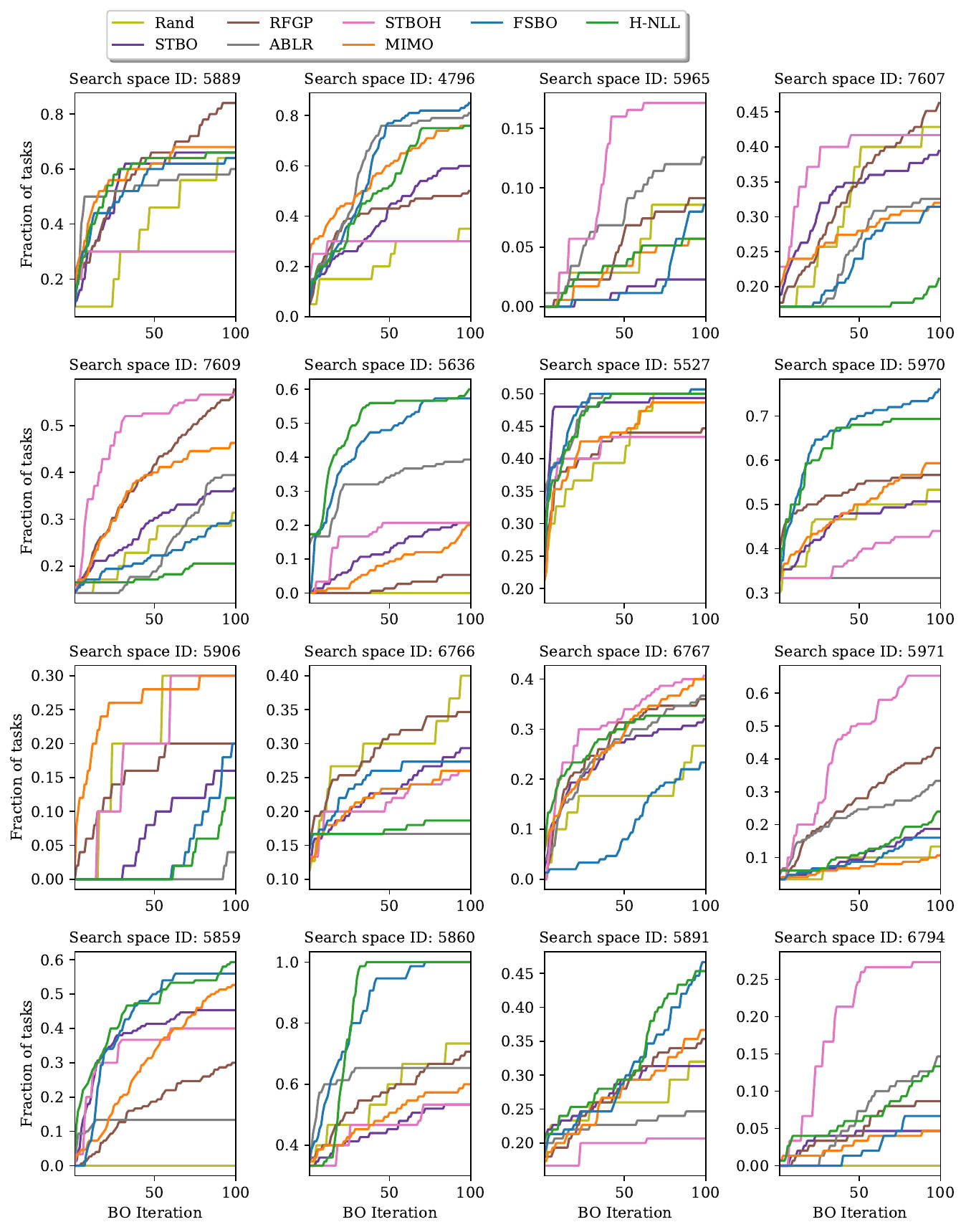}
    \caption{The performance profiles for each of the 16 search spaces in HPO-B. We use $C=0.001$ as the threshold.}
    \label{fig:reviewer5-pp}
\end{figure}

\begin{figure}
    \centering
    \includegraphics[width=\textwidth]{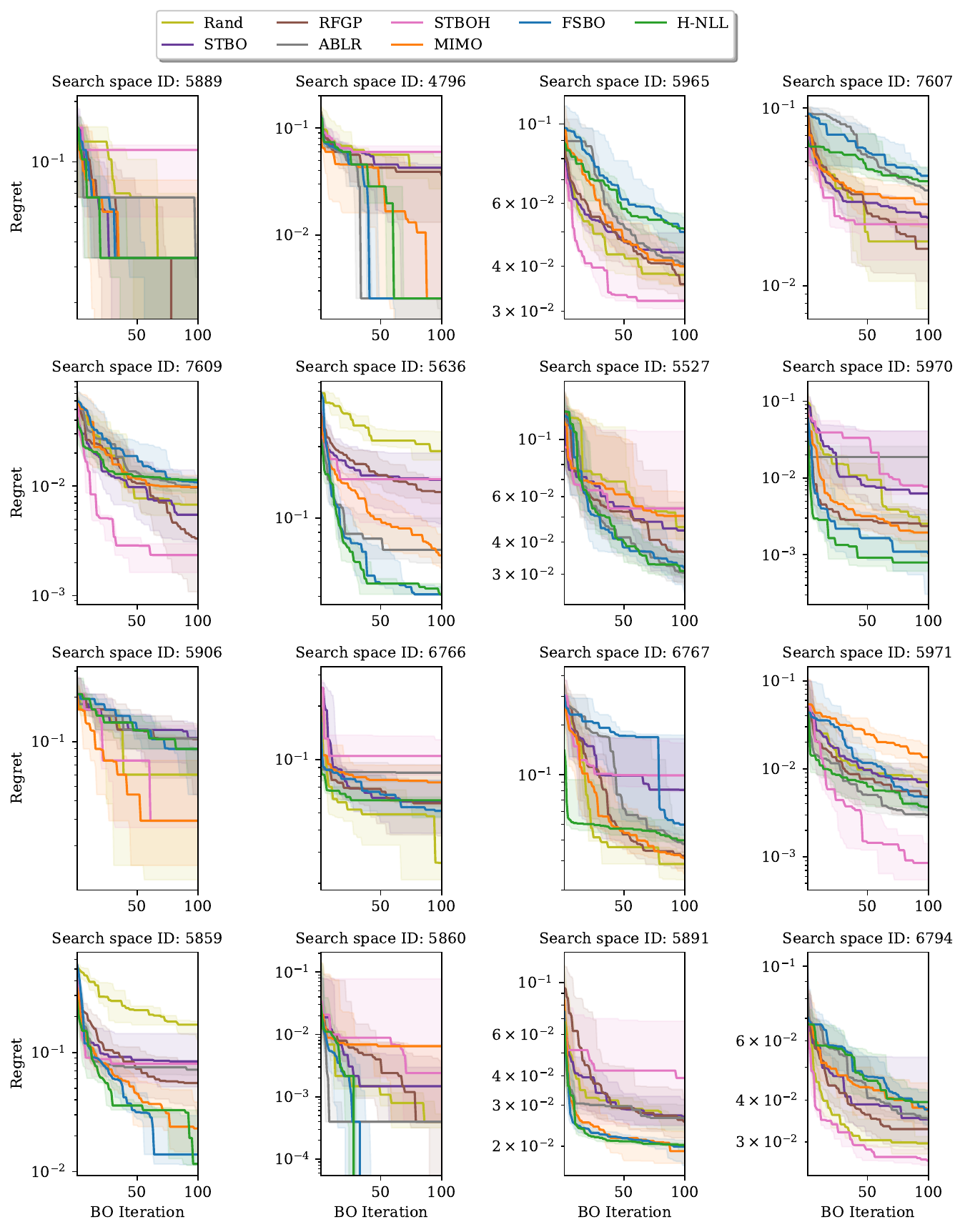}
    \caption{The regrets for each of the 16 search spaces in HPO-B.}
    \label{fig:reviewer5-regret}
\end{figure}

\begin{figure}
    \centering
    \includegraphics[width=\textwidth]{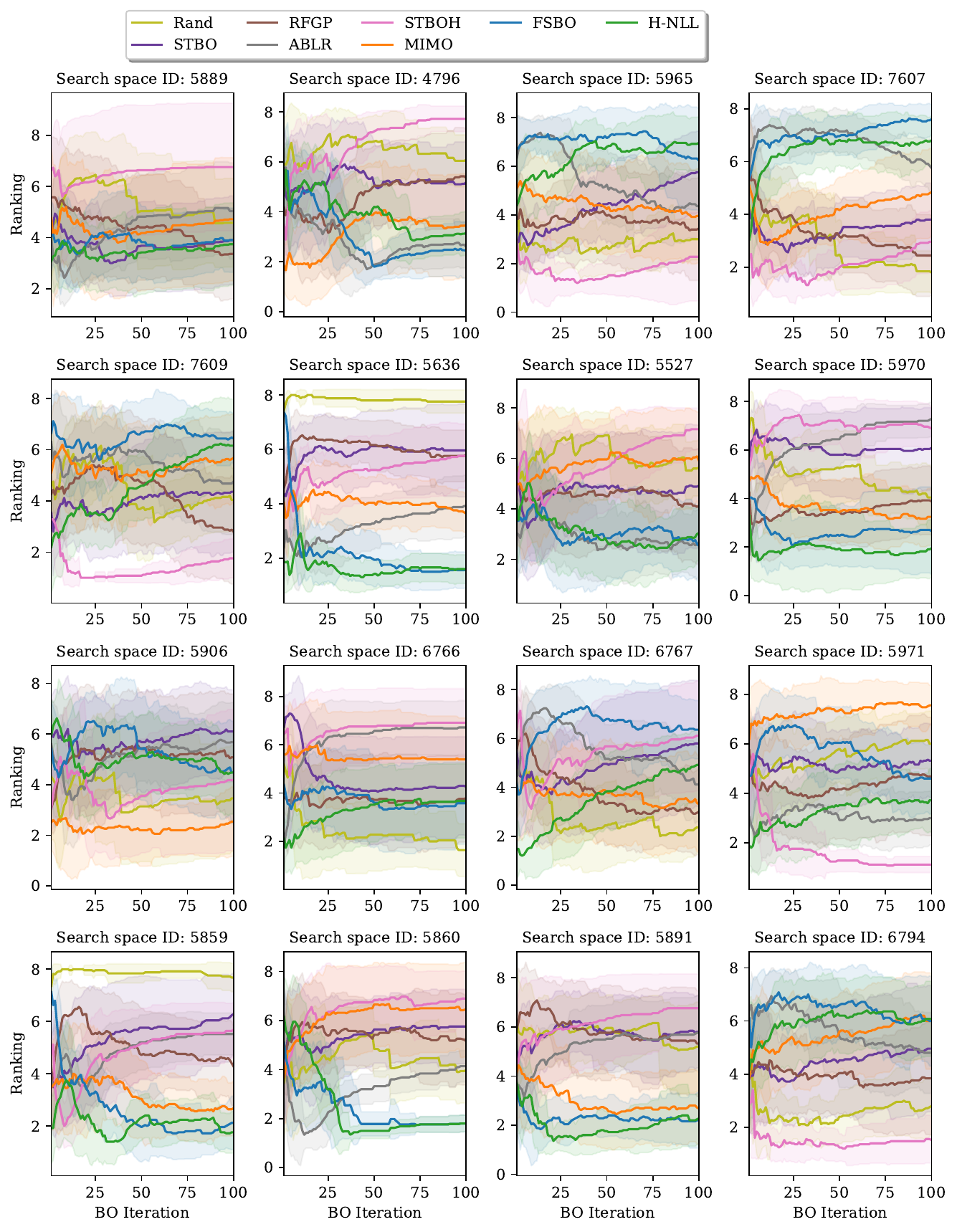}
    \caption{The ranking plots for each of the 16 search spaces in HPO-B.}
    \label{fig:reviewer5-rank}
\end{figure}

For readers who are interested in studying the details of the HPO-B experiment, we present the results separately for each of the 16 search spaces of HPO-B. The experiment setups are the same as \S\ref{sssec:hpob_full}. Figure~\ref{fig:reviewer5-pp} shows the performance profiles, Figure~\ref{fig:reviewer5-regret} shows the regret plots, and Figure~\ref{fig:reviewer5-rank} shows the ranking plots. 


\subsection{MAF Implementation Details}
\label{app:maf}
We compared to~\citep{volpp2020meta} using the code and default hyperparameters provided by the authors.\footnote{\url{https://github.com/boschresearch/MetaBO}} This method assumes the availability of the optimal set of GP hyperparameters for each task (including the task used for evaluation). Following~\citet{volpp2020meta}, these GP hyperparameters for the MAF algorithm are learned by optimizing the marginal likelihood on each training and evaluation task using the \textsc{GPy} library. Given that MAF takes significantly longer to run than HyperBO and other baselines, sub-dataset in each task was evaluated using limited random seeds.

Each neural acquisition function was trained for a total of 1000 iterations. As was done in~\citep{volpp2020meta}, we selected the optimal training iteration for the neural acquisition function by cross-validation on the transfer learning tasks; in this case, we randomly sampled 3 transfer learning task, and chose the training iteration with the lowest average simple regret.

Finally, to make use of the MAF code, we also had to ensure that (a) each task had the same number of evaluation points, and (b) that there were no duplicated tuning parameters. For this reason, we first removed all duplicate hyperparameters within each sub-dataset, then capped each sub-dataset to the first 1559 points (the size of the smallest sub-dataset) while retaining the best possible datapoint.

\newpage
\vskip 0.2in
\bibliography{23-0269}
\end{document}